%% file: main.tex
\documentclass{article}
\PassOptionsToPackage{table}{xcolor}
\PassOptionsToPackage{sort}{natbib}
\usepackage[T1]{fontenc}
\usepackage{iclr2027_conference,times}
\usepackage{url}
\usepackage{xurl}
\usepackage{booktabs}
\usepackage{amsfonts}
\usepackage{amsmath}
\usepackage{amssymb}
\usepackage{nicefrac}
\usepackage{microtype}
\usepackage[table]{xcolor}
\usepackage{colortbl}
\usepackage{graphicx}
\usepackage{tabularx}
\usepackage{longtable}
\usepackage{multirow}
\usepackage{enumitem}
\usepackage{algorithm}
\usepackage{algpseudocode}
\usepackage{listings}
\usepackage{chngcntr}
\usepackage{svg}

\usepackage{booktabs}
\usepackage{multirow}
\usepackage{makecell}
\usepackage{array}
\usepackage{graphicx}
\usepackage{tikz}

\definecolor{NumBlue}{HTML}{2F6FCC}
\definecolor{SymPurple}{HTML}{7A35B5}
\definecolor{SearchGreen}{HTML}{238636}

\newcolumntype{C}[1]{>{\centering\arraybackslash}m{#1}}
\newcolumntype{L}[1]{>{\raggedright\arraybackslash}p{#1}}

\newcommand{\scorecell}[3]{%
  \cellcolor{#1!#2}#3%
}

\newcommand{\bestscore}[3]{%
  \cellcolor{#1!#2}%
  \tikz[baseline=(X.base)]{
    \node[
      draw=#1,
      line width=1.0pt,
      rounded corners=1pt,
      inner xsep=5pt,
      inner ysep=1.5pt,
      minimum width=0.92cm
    ] (X) {\textbf{#3}};
  }%
}

\newcommand{\secondscore}[3]{%
  \cellcolor{#1!#2}%
  \tikz[baseline=(X.base)]{
    \node[
      draw=gray!70,
      line width=0.55pt,
      rounded corners=1pt,
      inner xsep=5pt,
      inner ysep=1.5pt,
      minimum width=0.92cm
    ] (X) {#3};
  }%
}

\svgsetup{inkscapelatex=false}
\usepackage{subcaption}
\usepackage{placeins}
\usepackage[normalem]{ulem}

\usepackage[hypertexnames=false]{hyperref}
\usepackage{cleveref}

\providecommand{\doi}[1]{}
\renewcommand{\doi}[1]{\mbox{\href{https://doi.org/#1}{doi:\,\nolinkurl{#1}}}}
\title{SymbolicArena: A Unified Infrastructure for Benchmark Distillation and Dynamic Evaluation in Symbolic Regression}

\author{%
  \bfseries
  \parbox{\textwidth}{\raggedright
    Ziwen Zhang\textsuperscript{1,2}$^{\dagger}$\quad
    Xiju Wu\textsuperscript{2,3}$^{\dagger}$\quad
    Yuheng Jing\textsuperscript{1,2}$^{*}$\quad
    Runxiang Wang\textsuperscript{1,2}\quad
    Boxiao Wang\textsuperscript{1,2}\quad
    Yifan Zang\textsuperscript{1,2}\quad
    Yifan Zhang\textsuperscript{2,3}\quad
    Yang Wang\textsuperscript{1,2}\quad
    Kai Li\textsuperscript{1,2}$^{*}$\quad
    Yifan Zhang\textsuperscript{2}\quad
    Huilin Xu\textsuperscript{3}\quad
    Jian Cheng\textsuperscript{2,4}\\[0.8em]
    \mdseries\footnotesize
    \textsuperscript{1}School of Artificial Intelligence, University of Chinese Academy of Sciences\\
    \textsuperscript{2}C$^{2}$DL, Institute of Automation, Chinese Academy of Sciences\\
    \textsuperscript{3}Nanjing University of Science and Technology\\
    \textsuperscript{4}School of Future Technology, University of Chinese Academy of Sciences\\[0.4em]
    Code: \textcolor{NumBlue}{\url{https://github.com/scientific-intelligent-modelling}}%
  }%
}

\newcommand{\fullset}{\mathcal{D}_{664}}       %
\newcommand{\probepanel}{\mathcal{A}_{4}}      %
\newcommand{\seedset}{\mathcal{S}}             %

\newcommand{\arena}{\textit{SymbolicArena}}
\newcommand{\core}{\textit{Core50}} 
\newcommand{\reservoir}{\textit{Full Task Set}} %
\newcommand{\calibset}{\textit{Calibration Set}}  %
\newcommand{\probe}{\textit{four probes}} %

\newcommand{\idq}{\textbf{\textit{ID}}}
\newcommand{\oodg}{\textbf{\textit{OOD}}}
\newcommand{\symf}{\textbf{\textit{SYM}}}
\newcommand{\eff}{\textbf{\textit{EFF}}}
\newcommand{\stab}{\textbf{\textit{STAB}}}
\newcommand{\minn}{\textbf{\textit{MIN}}}

\begin{document}

\addtocontents{toc}{\protect\setcounter{tocdepth}{-2}}

\iclrfinaltrue
\maketitle
{\def\thefootnote{\fnsymbol{footnote}}%
\footnotetext[2]{Equal contribution.}%
\footnotetext[1]{Corresponding authors.}}
\thispagestyle{plain}
\pagestyle{plain}

\input{sections/00_abstract}
\input{sections/01_introduction}
\input{sections/related_work}
\input{sections/02_SymbolicArena}
\input{sections/03_experiments}
\input{sections/04_limitations}
\input{sections/05_conclusion}

\input{sections/reproducibility_statement}

\input{sections/iclr2027_ai_use_statement}

\begingroup
\raggedright
\bibliography{references_iclr2027}
\endgroup

\clearpage
\appendix
\addtocontents{toc}{\protect\setcounter{tocdepth}{2}}
\begingroup
\edef\AppendixTocSavedDepth{\number\value{tocdepth}}
\small
\renewcommand{\contentsname}{Appendix Contents}
\setcounter{tocdepth}{2}
\tableofcontents
\setcounter{tocdepth}{\AppendixTocSavedDepth}
\endgroup
\clearpage

\input{sections/appA_related_work}

\clearpage
\input{sections/appB_reservoir}

\clearpage
\input{sections/appC_distillation}

\clearpage
\input{sections/appD_execution_protocol}

\clearpage
\input{sections/appE_six_axis_protocol}

\clearpage
\input{sections/appF_sensitivity}

\clearpage
\input{sections/appG_additional_results}

\clearpage
\input{sections/appH_evaluated_baselines}

\clearpage
\input{sections/appJ_llm_usage}

\end{document}

%% file: sections/00_abstract.tex
\begin{abstract}
Symbolic regression (SR) seeks concise and interpretable mathematical expressions from data for scientific equation discovery. Existing SR benchmarks face a tradeoff between evaluation cost and benchmark validity. Repeated evaluation of large task pools is expensive, and compact benchmarks lack systematic evidence of preserved task diversity and algorithm discriminability. \textbf{\arena{}} provides a unified infrastructure for benchmark distillation and dynamic evaluation. The framework standardizes 664 heterogeneous tasks with executable ground truth expressions and distills the \reservoir{} into \core{}, a validated benchmark of 50 tasks. The distillation process preserves task coverage and algorithm discrimination under explicit balance constraints. \arena{} applies a unified execution protocol to heterogeneous SR algorithms and produces comparable outputs and search trajectories. Multi Axis Evaluation characterizes numerical quality, symbolic quality, and search behavior. \core{} reduces evaluation workload by $92.5\%$ and maintains agreement with \reservoir{} evaluations. Experiments show that \arena{} achieves $72.6\%$ to $86.7\%$ lower approximation error than alternative selectors, further supporting its fidelity to the \reservoir{}. Evaluation reveals a substantial gap between numerical fitting and symbolic recovery across current SR methods, suggesting that reliable equation recovery remains an open challenge.
\end{abstract}

%% file: sections/01_introduction.tex
\section{Introduction}
Symbolic regression (SR) aims to recover compact and interpretable mathematical expressions from data, providing an approach for scientific equation discovery and symbolic modeling \citep{Langley1981,Schmidt2009}. Recent progress in SR has expanded from evolutionary search \citep{koza1992genetic,cranmer2023pysr} to neural generation \citep{landajuela2022unified,valipour2021symbolicgpt} and large language model assisted discovery \citep{grayeli2024lasr,shojaee2025llmsr}. The growing number of SR algorithms increases the demand for reliable and consistent evaluation.

Existing SR benchmarks provide resources for algorithm comparison. Large benchmark collections support broad task coverage at substantial cost. Compact collections reduce evaluation cost, yet evidence that compact collections preserve conclusions from larger task populations remains limited. A reliable compact benchmark requires validation beyond task diversity. Selected tasks maintain algorithm discrimination and preserve conclusions from a larger task population. A systematic strategy for constructing compact and reliable SR benchmarks remains unavailable.

The evaluation challenge becomes stronger as SR algorithms adopt diverse search mechanisms and output representations. Existing benchmarks mainly emphasize task collection and final performance measurement. Reliable comparison also requires consistent execution and evaluation criteria that characterize different aspects of algorithm behavior.

\textbf{\arena{}} provides a unified framework for benchmark distillation and evaluation in symbolic regression. Compact benchmark construction is formulated as a distillation problem. The framework combines task information and algorithm responses from 664 standardized SR tasks to construct \core{}, a validated benchmark of 50 tasks. The distillation process preserves task coverage and algorithm discrimination at a reduced evaluation budget.
Beyond benchmark construction, \arena{} provides a unified evaluation protocol for diverse SR algorithms. The framework standardizes execution and records search trajectories. Multi Axis Evaluation organizes metrics into numerical quality, symbolic quality, and search behavior.

Experiments on 15 representative SR algorithms validate the fidelity and diagnostic value of \arena{}. \core{} reduces approximation error to the \reservoir{} by $72.6\%$ to $86.7\%$ relative to alternative selectors. 
Across nine algorithms at both scales, Spearman rank correlation reaches $0.9833$ for ID and $0.9667$ for OOD. The OOD ordering of three validation algorithms, FePySR, SymbolFit, and JAXSR, is preserved. The main contributions are as follows.

\noindent\textbf{Benchmark Distillation.}
We formulate compact SR benchmark construction as a distillation problem and derive \core{}, a validated 50 task subset from 664 executable tasks.

\vspace{0.35em}

\noindent\textbf{Unified Evaluation.}
\arena{} standardizes heterogeneous SR algorithms in a unified execution protocol, enabling reproducible comparison across representative methods.

\vspace{0.35em}

\noindent\textbf{Multi Axis Evaluation.}
We evaluate numerical quality, symbolic quality, and search behavior, revealing tradeoffs beyond final numerical error.

%% file: sections/related_work.tex
\section{Related Work}

\paragraph{Symbolic Regression Benchmarks.}
SR evaluation ranges from classical equation sets to broader benchmark suites.
Nguyen~\citep{uy2011semantically}, Keijzer~\citep{keijzer2003improving}, Korns~\citep{korns2011accuracy}, and Vladislavleva~\citep{vladislavleva2009order} provide controlled settings for equation recovery and extrapolation.
SRBench~\citep{lacava2021contemporary} expands evaluation across synthetic and real world regression problems, and SRBench 2025~\citep{SRBench2025call} considers predictive accuracy and expression complexity.
SRSD~\citep{matsubara2024rethinking} focuses on scientific equation rediscovery, and LLM SRBench~\citep{shojaee2025llmsrbench} evaluates LLM based equation discovery.
\Cref{tab:benchmark_comparison} compares representative benchmarks and evaluation capabilities. Existing SR benchmarks provide broad task coverage and specialized evaluation protocols. Validation of compact benchmark fidelity remains limited, particularly for preserving task coverage and algorithm discrimination.

\begin{table}[htbp] 
\centering \caption{\textbf{Comparison with representative symbolic regression benchmarks.}} \label{tab:benchmark_comparison} \footnotesize \setlength{\tabcolsep}{2pt} \renewcommand{\arraystretch}{1.08} \begin{tabularx}{\columnwidth}{@{}>{\raggedright\arraybackslash}p{\dimexpr0.42\linewidth-13.5pt\relax} c @{\hspace{10pt}} >{\raggedright\arraybackslash}X c c c c@{}} \toprule \textbf{Benchmark} & \textbf{Tasks} & \textbf{Focus} & \textbf{HP} & \textbf{CP} & \textbf{VC} & \textbf{ST} \\ \midrule SRBench~\citep{lacava2021contemporary} & 252 & Algorithm comparison & $\checkmark$ & $\checkmark$ & $\times$ & $\times$ \\ SRBench 2025~\citep{SRBench2025call} & 24 & Accuracy and complexity & $\checkmark$ & $\checkmark$ & $\times$ & $\times$ \\ SRSD~\citep{matsubara2024rethinking} & 240 & Scientific equation rediscovery & $\times$ & $\times$ & $\times$ & $\times$ \\ LLM-SRBench~\citep{shojaee2025llmsrbench} & 240 & LLM based equation discovery & $\times$ & $\times$ & $\times$ & $\times$ \\ \makecell[l]{Nguyen \citep{uy2011semantically}\\ Keijzer~\citeyearpar{keijzer2003improving}\\ Korns~\citeyearpar{korns2011accuracy}\\ Vladislavleva~\citeyearpar{vladislavleva2009order}} & \makecell{12\\ 15\\ 15\\ 8} & Equation recovery and extrapolation & $\times$ & $\times$ & $\times$ & $\times$ \\ \textbf{\arena{}} & \textbf{664$\rightarrow$50} & \textbf{Validated benchmark distillation} & $\checkmark$ & $\checkmark$ & $\checkmark$ & $\checkmark$ \\ \bottomrule \end{tabularx} \vspace{2pt} \resizebox{\columnwidth}{!}{\footnotesize \textbf{HP}: heterogeneous task pool. \textbf{CP}: cross paradigm evaluation. \textbf{VC}: validated compact benchmark. \textbf{ST}: search trajectory recording.} 
\end{table}

\paragraph{Efficient Benchmarking.}
Efficient evaluation has received attention in machine learning.
tinyBenchmarks~\citep{polo2024tinybenchmarks}, Anchor Points~\citep{vivek2024anchorpoints}, and SubLIME~\citep{saranathan2025sublime} investigate compact subset selection for approximating benchmark performance or preserving model rankings.
Prior research concentrates on NLP and general machine learning.
Compact SR evaluation remains less explored due to heterogeneous tasks and diverse algorithm responses. A broader review of related work is provided in \Cref{app:related-work}.

%% file: sections/02_SymbolicArena.tex
\section{SymbolicArena}
\label{sec:method}
\begin{figure}[htbp]
	\centering
	\includegraphics[width=\linewidth]{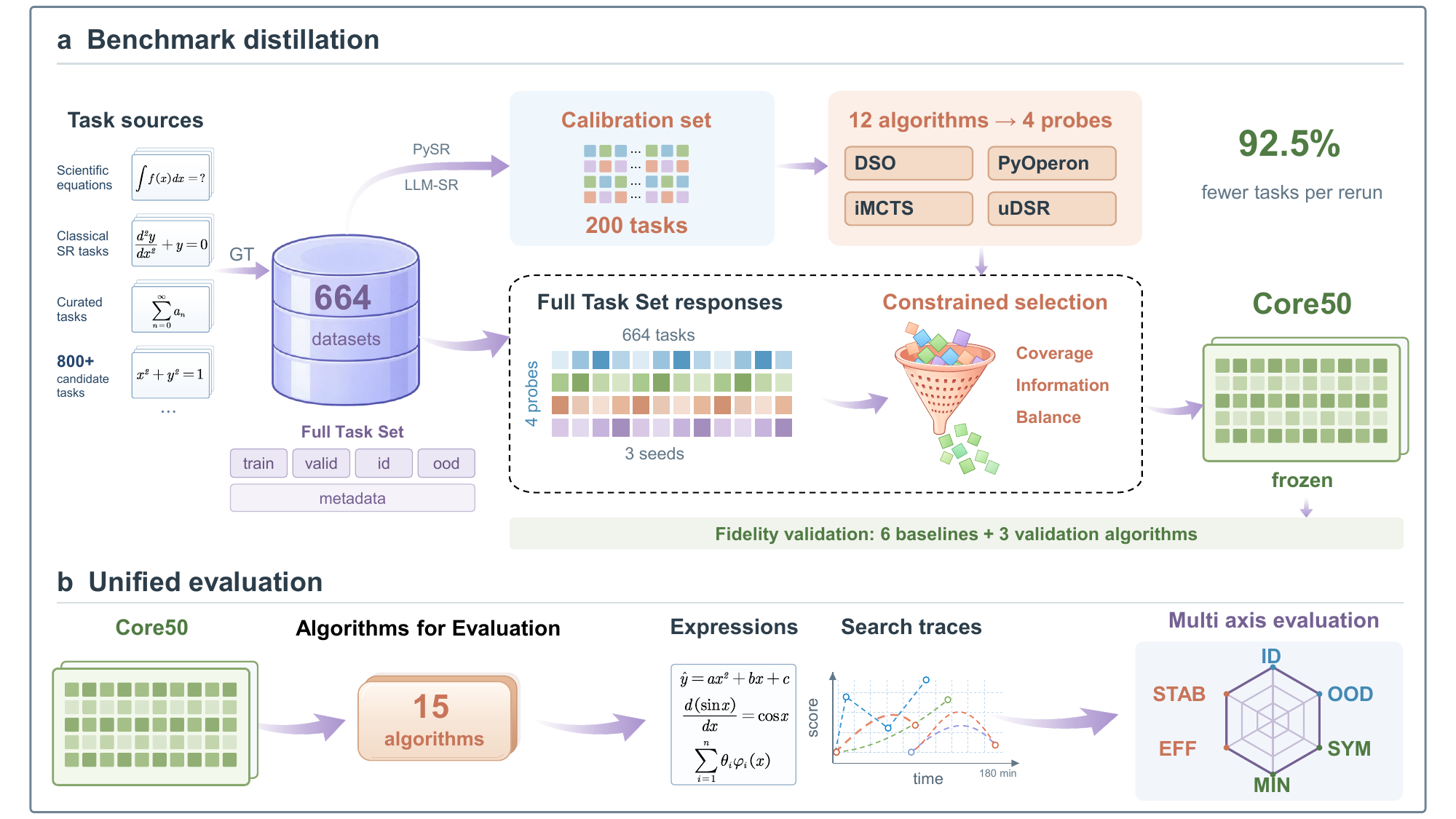}
    \caption{\textbf{Overview of \arena{}.}
    \textbf{a.} Benchmark distillation retains 664 tasks from 800 candidates, forming the \reservoir{}. The \calibset{} tasks support selection of the \probe{}. Task properties and probe responses across \reservoir{} guide selection of 50 tasks in \core{}.
    \textbf{b.} \arena{} runs 15 SR algorithms on \core{} using a common protocol and records outputs and search trajectories.
    Multi Axis Evaluation reports numerical quality (ID and OOD), symbolic quality (SYM and MIN), and search behavior (EFF and STAB).}
	\label{fig:symbolicarena-pipeline}
\end{figure}

\arena{} contains two stages for compact and reproducible SR evaluation, as shown in \Cref{fig:symbolicarena-pipeline}. \textbf{Benchmark distillation} constructs a compact task set that preserves task coverage and algorithm responses from the \reservoir{}. \textbf{Unified evaluation} applies a common execution protocol to the \core{} and measures numerical quality, symbolic quality, and search behavior.

\subsection{Benchmark Distillation}
\label{sec:benchmark-distillation}

We formulate compact SR evaluation as a subset selection problem. The goal is to identify a task subset whose aggregate scores and algorithm separation remain consistent with the \reservoir{}. The construction proceeds from task standardization to calibration selection, probe selection, and final benchmark distillation. The details and the task manifest appear in \Cref{app:core50-details} and \Cref{tab:table18_core50_ground_truth_manifest}.

\paragraph{\reservoir{} Construction.}
To support consistent evaluation across heterogeneous SR tasks, \arena{} maps each candidate task to a unified schema. The schema specifies feature ordering, target information, and fixed data splits. Ground truth expressions and structural metadata are retained for symbolic analysis and duplicate control.

A validation pipeline checks expression parsing, variable consistency, and target recomputation. The candidate pool exceeds 800 tasks. The 664 valid tasks form \reservoir{}, denoted by $\fullset$ :
\begin{equation*}
\fullset =
\left\{
\left(
f_i^{*},
D_i^{\mathrm{train}},
D_i^{\mathrm{ID}},
D_i^{\mathrm{OOD}},
m_i
\right)
\right\}_{i=1}^{664},
\end{equation*}
where $f_i^{*}$ denotes the ground truth expression,
$D_i^{\cdot}$ denotes the fixed data splits, and $m_i$ represents task
metadata. Detailed validation procedures are provided in
\Cref{app:reservoir}.

\paragraph{Calibration Task Selection.}

The \calibset{} is constructed to identify informative tasks from the
\reservoir{}. PySR and LLM-SR serve as initial probes because the search mechanisms produce complementary response patterns. Each probe runs once on every task in \reservoir{} to obtain an initial performance signal. Calibration selection only identifies informative tasks. 

Tasks are ranked according to the normalized performance difference between
PySR and LLM-SR on ID and OOD evaluation. The \calibset{} contains tasks
with strong probe disagreement and tasks solved by a single probe. Additional
tasks with moderate disagreement are included to maintain broader task
coverage. The construction audit is provided in
\Cref{app:candidate200-audit}.

\paragraph{Probe Selection.}

Probe selection maximizes mean calibration usability subject to coverage of at least three primary paradigms. Calibration usability combines finite result rate and resistance to numerical explosion. The 12 algorithms on the 200 calibration tasks remain eligible. Among four algorithm subsets covering at least three primary paradigms, the unique optimum is
\[
\probepanel=
\{\mathrm{DSO},\mathrm{PyOperon},\mathrm{iMCTS},\mathrm{uDSR}\}.
\]
The four probes cover Neural Policy Search, Evolutionary Search, and Structured Search. Definitions and selection results appear in
\Cref{app:probe4-selection}.

\paragraph{\core{} Construction.}
The four probes are evaluated on 664 tasks in the \reservoir{} using three random seeds. Error summaries, output validity, and seed variation characterize each task. Structural descriptors encode task properties. Response descriptors capture performance differences among the probes. Task informativeness combines discrimination and construction stability as $\mathrm{Info}_i=\mathrm{Disc}_i\,\cdot\ \mathrm{Stab}_i$. \core{} is selected as a feasible subset of 50 tasks by considering task coverage, mean task informativeness, and distributional balance. The selection maximizes the objective function 
\begin{equation*}
\begin{aligned}
J(S)&=0.45\,\mathrm{Coverage}(S)+0.35\,\mathrm{MeanInfo}(S)+0.20\,\mathrm{Balance}(S),
\qquad S\in\mathcal{F},
\end{aligned}
\end{equation*}
where $\mathrm{MeanInfo}(S)=|S|^{-1}\sum_{i\in S}\mathrm{Info}_i$, $|S|=50$.
The coverage is defined as follows
\[
\mathrm{Coverage}(S)=0.6\,\mathrm{StructuralCoverage}(S)
+0.4\,\mathrm{ResponseCoverage}(S).
\]
$\mathrm{Balance}(S)$ measures how well the subset represents the distributions of task difficulty and failure modes. The feasible set $\mathcal{F}$ controls source composition and duplicate membership. Additional constraints regulate difficulty and failure patterns. Details appear in \Cref{app:core50-scoring-search}.

We optimize the objective using greedy initialization followed by feasible single task swaps, and freeze the best observed subset as \core{} before fidelity validation and formal evaluation.

\subsection{Unified Evaluation Protocol}

\arena{} evaluates 15 SR algorithms on \core{} under a standardized execution protocol. Each algorithm uses a common wrapper. Inputs comprise the training split, feature names, and target information. ID and OOD test splits and ground truth expressions are reserved for evaluation.

Algorithm specific environments remain isolated. Data loading, metric
computation, and expression processing follow a shared implementation. The
execution layer records standardized outputs and trajectories of expressions
selected by each algorithm's native rule. The records support reproducible
comparison and search analysis. Implementation details are
provided in \Cref{app:execution-protocol}.

\subsection{Multi Axis Evaluation}
\label{sec:multi-axis-evaluation}
Multi Axis Evaluation assesses numerical quality with \idq{} and \oodg{}. \symf{} and \minn{} assess symbolic quality. \eff{} and \stab{} describe search behavior. Details are provided in
\Cref{app:six-axis-protocol}.

Opus 5 postprocessing simplifies final expressions and provides symbolic judgments for SYM and the structural component of STAB. MIN is computed from the frozen simplified expressions. Processed expressions and judgments are frozen before metric aggregation. The postprocessing procedure is detailed in \Cref{app:opus-postprocessing}.

The metrics are defined on a $[0,1]$ scale. All metrics except STAB use empirical averages across tasks and random seeds. STAB is aggregated across runs within each task and averaged across tasks. In the following formulas, $\mathbb{E}$ denotes the corresponding empirical average.

\paragraph{In-distribution quality} \idq{} measures numerical predictive quality on the ID test set
\begin{equation*}
\mathrm{Score}^{\mathrm{ID}} = \mathbb{E}\left[\,q^{\mathrm{ID}}\,\right], \qquad q^{\mathrm{ID}} = \phi(\mathrm{NMSE}^{\mathrm{ID}}),
\label{eq:quality-map}
\end{equation*}
where $\mathrm{NMSE}^{\mathrm{ID}}$ is the normalized mean squared error
on the ID test set. For a valid NMSE value $x$, the clipped log error is
\[
r(x)=\operatorname{clip}\!\left(
\log_{10}(\max(x,\epsilon)),\ell_{\min},\ell_{\max}\right).
\]
Numerical quality is obtained by linear rescaling
\begin{equation*}
\phi(x)=\frac{\ell_{\max}-r(x)}{\ell_{\max}-\ell_{\min}},
\qquad \phi(x)\in[0,1].
\label{eq:phi}
\end{equation*}
The bounds are $\ell_{\min}=-12$ and $\ell_{\max}=2$, with
$\epsilon=10^{-12}$. Lower error yields higher quality.
Invalid or unevaluable outputs receive quality zero.

\paragraph{Out-of-distribution quality} \oodg{} measures numerical predictive quality on the OOD test set with the same map
\begin{equation*}
\mathrm{Score}^{\mathrm{OOD}} = \mathbb{E}\left[\,q^{\mathrm{OOD}}\,\right], \qquad q^{\mathrm{OOD}} = \phi(\mathrm{NMSE}^{\mathrm{OOD}}),
\end{equation*}
where $\mathrm{NMSE}^{\mathrm{OOD}}$ is the normalized mean squared error on the OOD test set.

\paragraph{Symbolic fidelity} \symf{} prioritizes symbolic equivalence judgments
\begin{equation*}
\mathrm{Score}^{\mathrm{SYM}} = \mathbb{E}\left[\,m^{\mathrm{SYM}}\,\right], \qquad
m^{\mathrm{SYM}} =
\begin{cases}
1, & Eq=1,\\[3pt]
0.5(S_{\mathrm{tree}}F_vF_o)^{1/3}, & Eq=0,
\end{cases}
\label{eq:sym}
\end{equation*}
where the indicator $Eq$ records the evaluator's \texttt{equivalent} judgment for the
processed prediction and reference. The quantities $S_{\mathrm{tree}}$, $F_v$, and $F_o$ measure tree similarity, variable recovery, and operator recovery, respectively, as defined in
\Cref{app:Symbolic Quality}.

\paragraph{Minimality} \minn{} measures expression minimality relative to the reference
\begin{equation*}
\mathrm{Score}^{\mathrm{MIN}} = \mathbb{E}\left[\,m^{\mathrm{MIN}}\,\right], 
\qquad
m^{\mathrm{MIN}} = \min\left(1,\, \frac{C^{ref}}{C^{pred}}\right),
\label{eq:min}
\end{equation*}
where $C^{ref}$ and $C^{pred}$ denote the expression tree sizes of the simplified reference and prediction.

\paragraph{Efficiency}
\eff{} measures how rapidly search approaches the best numerical quality reached within a run. Let $T$ denote the number of checkpoints on the fixed time grid
\begin{equation*}
\mathrm{Score}^{\mathrm{EFF}}=\mathbb E[m^{\mathrm{EFF}}],
\qquad
m^{\mathrm{EFF}}=\frac1T\sum_{t=1}^{T}\frac{q(t)}{q^*},
\label{eq:eff}
\end{equation*}
where the numerical quality is $q(t)=(q^{\mathrm{ID}}(t)+q^{\mathrm{OOD}}(t))/2$,
and the normalization reference is $q^*=\max_{1\leq t\leq T}q(t)$. A run with $q^*=0$ receives $m^{\mathrm{EFF}}=0$. The experiment records one checkpoint per minute, and $T$ is expressed in minutes. At each checkpoint, $q(t)$ evaluates the expression selected by the algorithm's native rule. Test quality does not guide candidate selection. 

\paragraph{Stability}
\stab{} combines numerical consistency, output validity, and structural consistency
\begin{equation*}
\mathrm{Score}^{\mathrm{STAB}}
=
\mathbb{E}[m^{\mathrm{STAB}}],
\qquad
m^{\mathrm{STAB}}=(NVC)^{1/3},
\end{equation*}
where $N$, $V$, and $C$ denote numerical consistency, valid run fraction, and structural consistency.
Detailed definitions are provided in \Cref{app:metric-stab}.

%% file: sections/03_experiments.tex
\section{Experiments}
\label{sec:experiments}

Experiments examine benchmark fidelity, generalization beyond construction probes, and diagnostic value.

\subsection{Experimental Setup}
\label{sec:experimental-setup}

\paragraph{Construction and fidelity evaluation.}
\core{} construction follows \Cref{sec:benchmark-distillation}.
The four probes are evaluated on all 664 tasks using three random seeds.
Fidelity validation uses a separate one hour evaluation of nine algorithms on \reservoir{}.
FePySR~\citep{yu2026fepysr}, SymbolFit~\citep{tsoi2025symbolfit}, and JAXSR~\citep{kitchin2024jaxsr} are excluded from benchmark construction and serve as validation algorithms.
Scores on \core{} use the same evaluation records restricted to the frozen 50 tasks.
PySR and LLM-SR use seed 1314.
The remaining seven algorithms use seeds 520, 521, and 522.
Probe selection sensitivity appears in \Cref{app:sensitivity}.

\paragraph{Formal evaluation.}
The comparison evaluates 15 algorithms on frozen \core{} using seeds 520, 521, and 522. \Cref{app:evaluated-baselines} presents the algorithms and configurations.
Runs have a three hour budget.
Clean evaluation provides the six axis results.
Supplementary experiments perturb training labels at $1\%$ and $5\%$.
Construction, fidelity validation, and formal evaluation use separate records.
Metric definitions appear in \Cref{sec:multi-axis-evaluation}.
The execution protocol is described in \Cref{app:execution-protocol}.

\subsection{Fidelity to the Full Task Set}
\label{sec:benchmark-fidelity}

\paragraph{\core{} best preserves aggregate benchmark behavior.}

We evaluate how closely each 50 task subset reproduces aggregate probe scores on \reservoir{}.
Six alternative selectors are defined in \Cref{app:core50-validation}.
All selectors use the same construction records.
Let $A_p(S)$ denote the mean task score of probe $p$ on task set $S$.
Aggregate fidelity is measured by
\[
\operatorname{MAE}(S,\fullset)
=
\frac{1}{|\probepanel|}
\sum_{p\in\probepanel}
\left|A_p(S)-A_p(\fullset)\right|.
\]
Lower MAE indicates closer agreement with \reservoir{}.

\begin{figure}[htbp]
	\centering
	\captionsetup[subfigure]{labelformat=simple, labelsep=period}
	\begin{subfigure}[b]{0.32\linewidth}
		\centering
		\includegraphics[width=\linewidth]{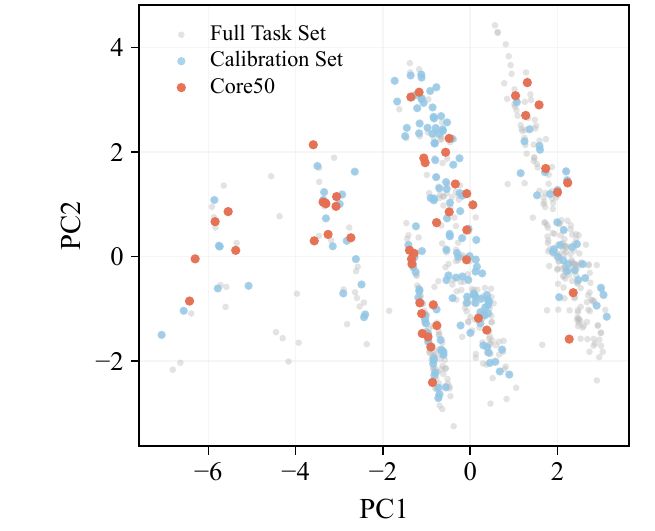}
		\caption{Coverage}
		\label{fig:core50-validation:coverage}
	\end{subfigure}\hfill
	\begin{subfigure}[b]{0.32\linewidth}
		\centering
		\includegraphics[width=\linewidth]{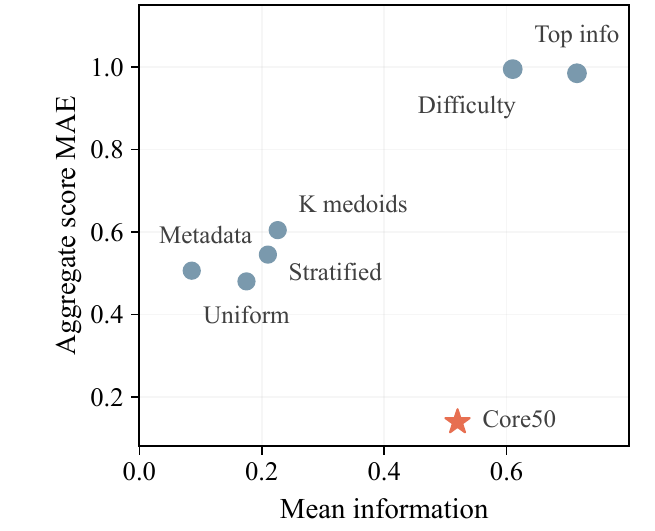}
		\caption{Tradeoff}
		\label{fig:core50-validation:tradeoff}
	\end{subfigure}\hfill
	\begin{subfigure}[b]{0.32\linewidth}
		\centering
		\includegraphics[width=\linewidth]{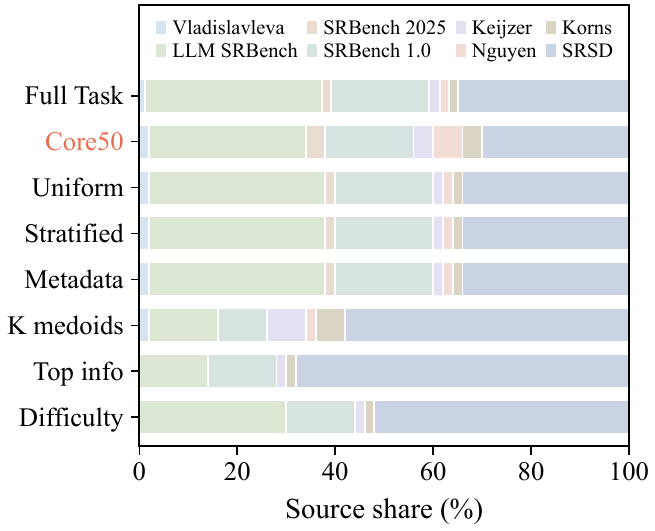}
		\caption{Composition}
		\label{fig:core50-validation:composition}
	\end{subfigure}
	\caption{\textbf{Aggregate fidelity and task representation.}
	\textbf{a.} PCA projection of structural and probe response descriptors for \reservoir{}, \calibset{}, and \core{}.
	\textbf{b.} Mean task information and aggregate score MAE for \core{} and six alternative selectors.
	\textbf{c.} Source proportions for \reservoir{} and the evaluated 50 task subsets.}
	\label{fig:core50-validation}
\end{figure}

\core{} achieves the lowest aggregate score MAE of $0.1388$.
In \Cref{fig:core50-validation:coverage}, \core{} occupies the major regions of the joint structural and response space, indicating broad representation of the task population.
In \Cref{fig:core50-validation:tradeoff}, top information and difficulty balanced selection achieve higher mean task information yet produce much larger MAE.
In \Cref{fig:core50-validation:composition}, several baselines closely match the source proportions of \reservoir{} yet retain higher MAE.
The results indicate that benchmark fidelity benefits from both representative coverage and informative algorithm responses. Optimizing either factor alone is insufficient among the evaluated selectors.

\begin{figure}[htbp]
	\centering
	\captionsetup[subfigure]{labelformat=simple, labelsep=period}
	\begin{subfigure}[t]{0.4\linewidth}
		\centering
		\includegraphics[width=\linewidth, trim=0pt 20pt 0pt 0pt, clip]{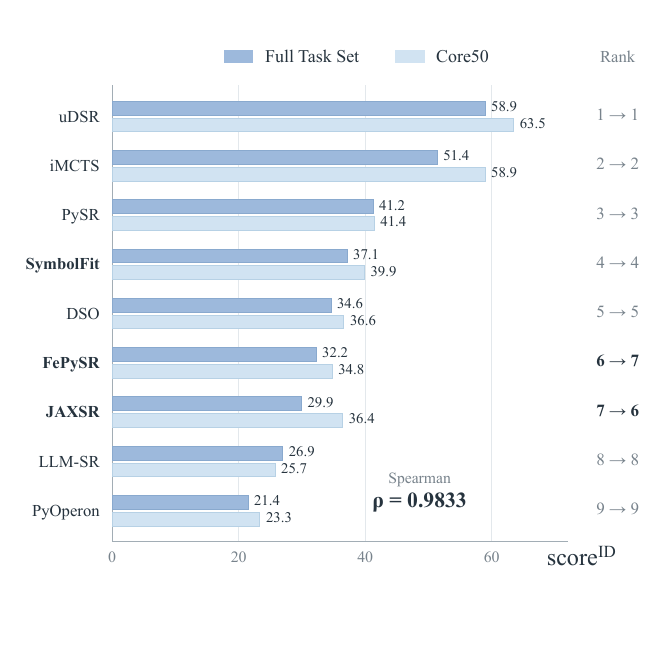}
		\caption{ID}
		\label{fig:heldout_validation:id}
	\end{subfigure}
	\hspace{0.05\linewidth}
	\begin{subfigure}[t]{0.4\linewidth}
		\centering
		\includegraphics[width=\linewidth, trim=0pt 20pt 0pt 0pt, clip]{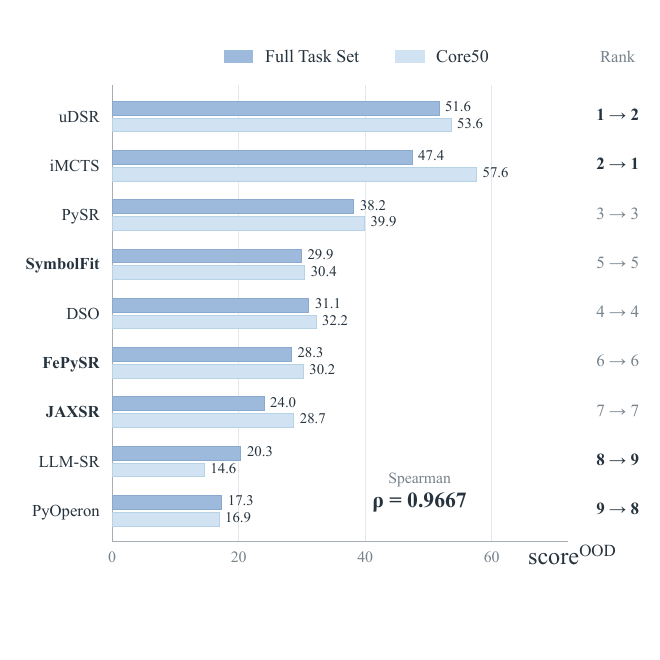}
		\caption{OOD}
		\label{fig:heldout_validation:ood}
	\end{subfigure}
	\caption{\textbf{Numerical score and rank agreement.}
	ID and OOD scores for nine algorithms on \reservoir{} and \core{}.
	FePySR, SymbolFit, and JAXSR are excluded from benchmark construction.
	Scores use identical evaluation records and a one hour budget.}
	\label{fig:heldout_validation}
\end{figure}

\paragraph{Fidelity on algorithms excluded from construction.}
We evaluate nine algorithms on both \reservoir{} and \core{} using identical evaluation records. \Cref{fig:heldout_validation:id} and \Cref{fig:heldout_validation:ood} report ID and OOD score and rank agreement across the two scales.
FePySR, SymbolFit, and JAXSR are excluded from benchmark construction and serve as validation algorithms.
Across all nine algorithms, Spearman correlations reach $\rho=0.9833$ for ID and $\rho=0.9667$ for OOD.
The three validation algorithms preserve their OOD ordering, and one ID pair changes order.
The strong agreement provides evidence that \core{} fidelity is not confined to the construction probes.

\subsection{Diagnostic Analysis of SR Methods}
\label{sec:algorithm-evaluation}

The validated \core{} enables diagnostic comparison of 15 representative SR algorithms.
\Cref{tab:six_axis_profiles} reports point estimates on clean \core{}.
Task bootstrap intervals are provided in \Cref{app:algorithm-axis-profiles}, and axis correlations appear in \Cref{app:axis-correlation}.

\input{tables/six_axis_table_clean}

\begin{figure}[htbp]
    \centering
    \includegraphics[width=0.86\linewidth]{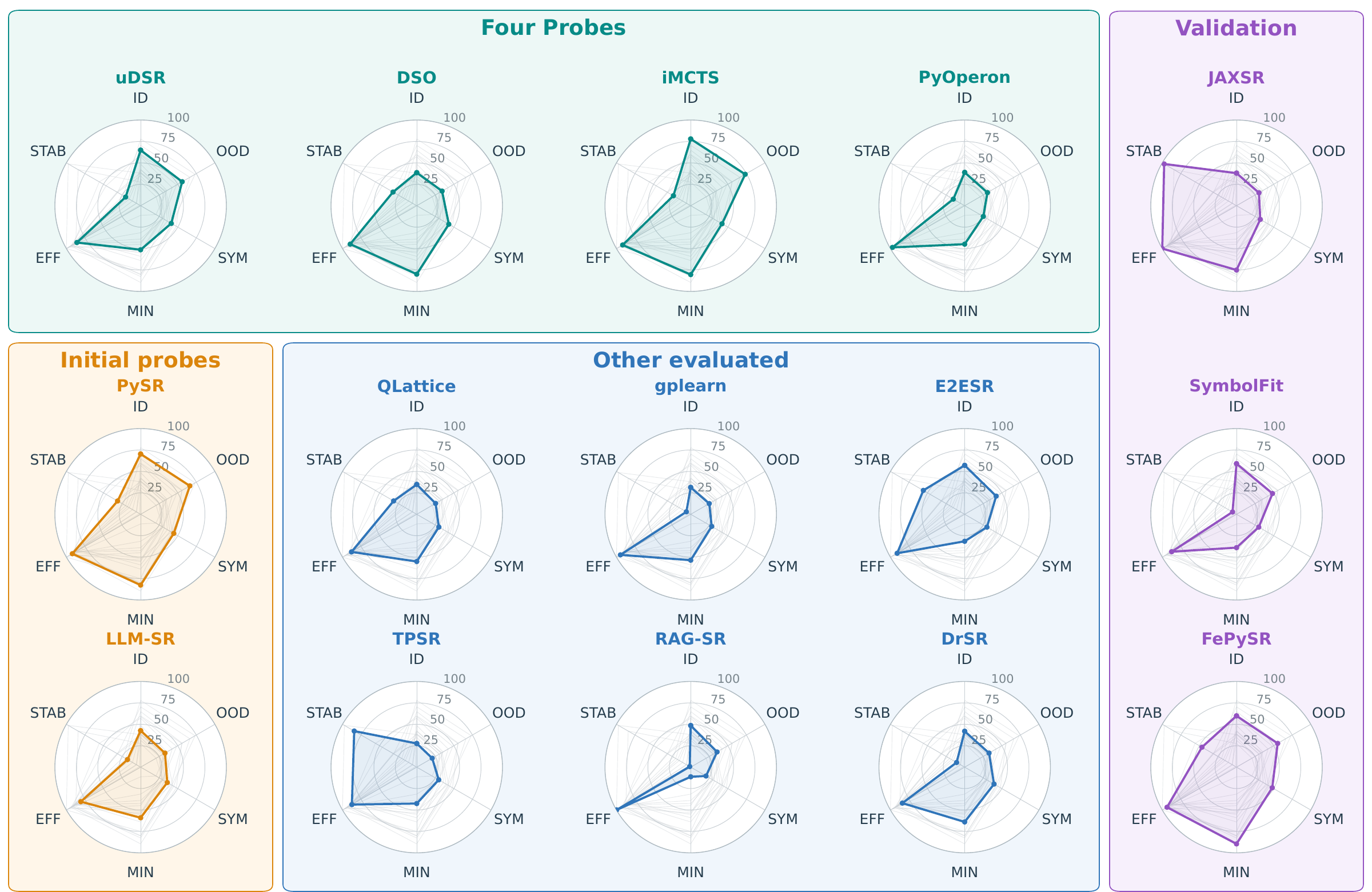}
    \caption{\textbf{Per algorithm six axis profiles on \core{}.}
    Each panel displays the six axis scores of one algorithm. Normalized
    scores are multiplied by 100 for a common 0 to 100 display scale.
    Gray profiles show the remaining algorithms for reference. The panels
    reveal differences beyond a single numerical ranking.}
    \label{fig:six-axis-radar-profiles}
\end{figure}

\paragraph{No evaluated method dominates all dimensions.}
\Cref{tab:six_axis_profiles} shows that the highest point estimates occur in different methods across numerical, symbolic, and search metrics.
iMCTS leads ID and OOD, FePySR leads SYM and MIN, and JAXSR leads EFF and STAB. PySR ranks second on ID, OOD, SYM, and MIN. \Cref{fig:six-axis-radar-profiles} further reveals distinct profile shapes across methods, indicating strong capability specialization.

\paragraph{Strong numerical accuracy does not guarantee symbolic recovery.}
\Cref{tab:six_axis_profiles} shows that strong numerical performance does not consistently correspond to strong symbolic fidelity.
iMCTS leads both numerical axes yet attains a SYM score of $42.15$.
A clean \texttt{Nguyen-9} case in \Cref{app:representative-symbolic-mismatch} provides a direct example.
Near zero numerical error can coexist with failure of strict symbolic equivalence.
Final numerical error alone is insufficient to characterize equation recovery.

\paragraph{Search behavior is distinct from attained quality.}
\Cref{tab:six_axis_profiles} also reveals a clear separation between search behavior and attained solution quality.
JAXSR reaches near maximal EFF and STAB yet attains moderate ID and OOD scores.
iMCTS shows a contrasting profile, leading both numerical axes with a STAB score of $23.31$.
A high EFF score can reflect early attainment of a modest peak because EFF is normalized to each run's observed maximum.
High STAB indicates repeatability and does not directly imply high attained quality.
Search progress and repeatability provide complementary information to final solution quality.

%% file: tables/six_axis_table_clean.tex
\begin{table*}[htbp]
\centering
\begingroup

\definecolor{SATableInk}{HTML}{000000}
\definecolor{SATableMuted}{HTML}{000000}
\definecolor{SATableRule}{HTML}{CED2D7}
\definecolor{SABestBackground}{HTML}{E6F2E8}
\definecolor{SABestInk}{HTML}{1B6B2B}
\definecolor{SASecondBackground}{HTML}{E2EDF8}
\definecolor{SASecondInk}{HTML}{244C78}

\setlength{\abovecaptionskip}{5pt}
\setlength{\belowcaptionskip}{7pt}
\caption{Six-axis profiles of 15 symbolic regression methods on \core{} (clean).
Methods are grouped by their primary search paradigm.
Higher values denote better performance on each axis.}
\label{tab:six_axis_profiles}

\fontsize{7.2}{9.2}\selectfont
\color{SATableInk}
\renewcommand{\arraystretch}{1.12}
\setlength{\tabcolsep}{0pt}
\setlength{\heavyrulewidth}{0.8pt}
\setlength{\lightrulewidth}{0.3pt}
\setlength{\cmidrulewidth}{0.3pt}
\setlength{\aboverulesep}{1.6pt}
\setlength{\belowrulesep}{1.6pt}

\edef\SAParadigmWidth{\the\dimexpr 0.13\linewidth\relax}
\edef\SAAlgorithmWidth{\the\dimexpr 0.14\linewidth\relax}
\edef\SAScoreWidth{\the\dimexpr(\linewidth-\SAParadigmWidth-\SAAlgorithmWidth)/6\relax}
\edef\SAHighlightWidth{\the\dimexpr\SAScoreWidth-11pt\relax}

\newcommand{\SAHighlight}[3]{%
  \begingroup
  \setlength{\fboxsep}{0.4pt}%
  \colorbox{#1}{\makebox[\SAHighlightWidth][c]{\textcolor{#2}{\bfseries #3}}}%
  \endgroup
}
\newcommand{\SABest}[1]{\SAHighlight{SABestBackground}{SABestInk}{#1}}
\newcommand{\SASecond}[1]{\SAHighlight{SASecondBackground}{SASecondInk}{#1}}
\newcommand{\SAGroup}[2]{%
  \multirow{#1}{\SAParadigmWidth}{%
    \color{SATableMuted}\fontsize{6.7}{8.6}\selectfont\raggedright #2}%
}
\newcommand{\SASeparator}{\arrayrulecolor{SATableRule}\midrule\arrayrulecolor{SATableInk}}

\begin{tabular}{@{}
  >{\raggedright\arraybackslash}m{\SAParadigmWidth}
  >{\raggedright\arraybackslash}m{\SAAlgorithmWidth}
  *{6}{>{\centering\arraybackslash}m{\SAScoreWidth}}
@{}}
\arrayrulecolor{SATableInk}
\toprule
& & \multicolumn{2}{c}{\textbf{Numerical}}
    & \multicolumn{2}{c}{\textbf{Symbolic}}
    & \multicolumn{2}{c}{\textbf{Search}} \\
\arrayrulecolor{SATableInk}
\cmidrule(lr){3-4}
\cmidrule(lr){5-6}
\cmidrule(lr){7-8}
\textbf{Paradigm} & \textbf{Algorithm}
  & \textbf{ID} $\uparrow$ & \textbf{OOD} $\uparrow$ & \textbf{SYM} $\uparrow$
  & \textbf{MIN} $\uparrow$ & \textbf{EFF} $\uparrow$ & \textbf{STAB} $\uparrow$ \\
\arrayrulecolor{SATableInk}
\midrule

\SAGroup{3}{Structured\\Search}
 & iMCTS & \SABest{77.91} & \SABest{73.35} & 42.15 & 80.37 & 91.66 & 23.31 \\
 & QLattice & 34.79 & 25.37 & 29.82 & 55.12 & 87.81 & 31.12 \\
 & JAXSR & 37.99 & 30.21 & 32.00 & 75.03 & \SABest{99.99} & \SABest{97.43} \\
\SASeparator

\SAGroup{4}{Evolutionary\\Search}
 & PySR & \SASecond{70.34} & \SASecond{66.28} & \SASecond{44.60} & \SASecond{82.62} & 91.94 & 31.03 \\
 & PyOperon & 38.89 & 30.64 & 25.06 & 44.88 & 97.28 & 15.34 \\
 & gplearn & 31.37 & 24.96 & 28.25 & 53.55 & 94.67 & 5.86 \\
 & SymbolFit & 59.04 & 48.46 & 29.96 & 38.92 & 87.30 & 5.38 \\
\SASeparator

\SAGroup{2}{Neural Policy\\Search}
 & uDSR & 64.84 & 55.94 & 41.25 & 51.41 & 85.83 & 20.20 \\
 & DSO & 38.56 & 34.14 & 43.19 & 79.90 & 89.60 & 31.92 \\
\SASeparator

\SAGroup{2}{Hybrid\\Search}
 & FePySR & 59.84 & 55.45 & \SABest{48.13} & \SABest{89.55} & 93.56 & 46.53 \\
 & RAG-SR & 48.58 & 35.45 & 20.66 & 11.25 & \SASecond{99.74} & 1.37 \\
\SASeparator

\SAGroup{2}{Transformer\\Methods}
 & E2ESR & 57.09 & 42.40 & 29.96 & 31.50 & 90.95 & 55.70 \\
 & TPSR & 27.60 & 20.84 & 29.60 & 42.46 & 87.33 & \SASecond{84.20} \\
\SASeparator

\SAGroup{2}{LLM Assisted\\Search}
 & DrSR & 41.86 & 32.82 & 39.51 & 63.94 & 83.93 & 10.92 \\
 & LLM-SR & 42.59 & 32.78 & 35.89 & 59.06 & 80.55 & 17.72 \\
\arrayrulecolor{SATableInk}
\bottomrule
\end{tabular}

\par\vspace{5pt}
{\fontsize{6.2}{7.6}\selectfont\color{SATableMuted}
\makebox[\linewidth][l]{%
  Display scale 0--100.\hfill
  {\setlength{\fboxsep}{0pt}\colorbox{SABestBackground}{\rule{0pt}{5pt}\hspace{8pt}}}%
  \hspace{3pt}\textcolor{SABestInk}{Best (1st)}\hspace{13pt}%
  {\setlength{\fboxsep}{0pt}\colorbox{SASecondBackground}{\rule{0pt}{5pt}\hspace{8pt}}}%
  \hspace{3pt}\textcolor{SASecondInk}{Second-best (2nd)}%
}\par
}

\endgroup
\end{table*}

%% file: sections/04_limitations.tex
\section{Limitations}

\paragraph{Benchmark scope and fidelity.}
\core{} is distilled from the current \reservoir{} using responses from the
\probe{}.
Fidelity is validated only for the current task distribution and the algorithmic paradigms represented in the evaluation.
New task families or substantially different search paradigms may require renewed fidelity validation of \core{}.
\core{} also inherits the coverage boundary of the \reservoir{}.
Scientific domains and operator families absent from the current 664 tasks remain outside the validated scope.

\paragraph{Algorithm configuration.}
The evaluation uses official default or recommended configurations when available and applies only adaptations required by the shared protocol.
Algorithm specific hyperparameter optimization is not included.
Absolute scores and relative rankings may change with dedicated tuning.
The reported results characterize each method in a standardized benchmark configuration.

%% file: sections/05_conclusion.tex
\section{Conclusion}

\arena{} provides a unified framework for efficient and reproducible symbolic
regression evaluation.
The framework standardizes 664 executable SR tasks and distills \core{}, reducing
the evaluation workload by $92.5\%$.
Relative to six alternative 50 task selectors, \core{} reduces aggregate score
MAE by $72.6\%$ to $86.7\%$ and preserves strong ranking agreement with
\reservoir{}.
Validation on algorithms excluded from benchmark construction further supports
the fidelity of the compact benchmark.

Evaluation of 15 representative SR methods reveals clear differences across
numerical quality, symbolic quality, and search behavior.
No evaluated method dominates all dimensions.
Strong numerical accuracy does not consistently correspond to faithful symbolic
recovery, and strong search progress or repeatability does not imply high attained
solution quality.
The results show that final numerical error alone provides an incomplete view of
SR performance.
\arena{} offers a reproducible benchmark and evaluation protocol for studying
such differences at substantially lower cost.

%% file: sections/reproducibility_statement.tex
\paragraph{Reproducibility Statement.}
\arena{} fixes task manifests, data splits, random seeds, execution budgets,
algorithm configurations, and evaluation protocols.
Appendices B to E document benchmark construction, execution procedures, and
metric definitions.
Appendix I documents model assisted symbolic evaluation, prompt templates, and
information access controls.
The supplementary material provides the manifests, scripts, configurations,
and frozen symbolic judgments required to reproduce the reported results.

%% file: sections/iclr2027_ai_use_statement.tex
\subsection*{AI use statement}
Generative AI tools assisted manuscript drafting and revision, LaTeX editing, figure preparation, literature retrieval, code inspection, and analysis of existing experimental records. AI tools also supported research discussion, experimental design review, consistency checks, and interpretation of experimental results. All methodological decisions, experimental designs, and reported results were reviewed and verified by the authors. Opus 5 assisted final expression simplification, equivalence assessment, and structural judgments for seed pairs in the evaluation pipeline. The resulting processed expressions and judgments were frozen before metric aggregation and retained for auditing.

%% file: sections/appA_related_work.tex
\section{Related Work}
\label{app:related-work}
\paragraph{Symbolic Regression Algorithms.}
Genetic programming and evolutionary optimization provide established approaches to symbolic regression \citep{burlacu2020operon,cranmer2023pysr}. Learning methods guide expression search using several complementary strategies. Deep Symbolic Regression uses reinforcement learning to optimize expression generation \citep{petersen2021dsr}. Deep Generative Symbolic Regression \citep{kamienny2023deep}, iMCTS \citep{huang2025imcts}, and TPSR \citep{shojaee2023tpsr} explore tree search and planning. Neural models also guide symbolic search \citep{landajuela2022unified,mundhenk2021dsoseeding}. NeSymReS and End-to-End Symbolic Regression generate expressions directly from data \citep{biggio2021nesymres,kamienny2022end}. LLM-SR \citep{shojaee2025llmsr} and LaSR \citep{grayeli2024lasr} use large language models for equation discovery. RAG-SR \citep{zhang2025ragsr} and DrSR \citep{wang2025drsr} explore further LLM based discovery strategies. The diversity of search strategies and inductive biases motivates a unified evaluation across paradigms.

\paragraph{Symbolic Regression Benchmarks.}
Existing SR benchmarks support broad algorithm comparison and controlled scientific equation discovery. The Keijzer benchmark~\citeyearpar{keijzer2003improving} evaluates numerical fitting, linear scaling, and extrapolation. The Vladislavleva benchmark~\citeyearpar{vladislavleva2009order} emphasizes model complexity, numerical accuracy, and extrapolation performance. The Korns benchmark~\citeyearpar{korns2011accuracy} assesses the exact recovery of complex multivariate expressions. Nguyen evaluates elementary analytical function recovery and genetic programming search capabilities \citep{uy2011semantically}. SRBench evaluates contemporary SR algorithms across synthetic and real-world regression problems \citep{lacava2021contemporary}. Its recent extension broadens algorithm coverage and compares predictive accuracy, expression complexity, and computational cost \citep{SRBench2025call}. SRSD reconstructs scientific equation-discovery tasks with physically meaningful sampling ranges and expression-level similarity measures \citep{matsubara2024rethinking}. LLM-SRBench evaluates LLM-based equation discovery on tasks designed to reduce direct formula memorization \citep{shojaee2025llmsrbench}. The benchmarks provide valuable datasets and evaluation protocols, yet construction, execution, and diagnostic coverage remain heterogeneous. A unified framework that preserves benchmark fidelity and supports cross-paradigm evaluation is still missing.

\paragraph{Benchmark Distillation.}
Efficient benchmarking seeks to reduce evaluation cost and preserve conclusions from the full benchmark. \citet{perlitz2024efficient} study how smaller evaluation sets can retain reliable model comparisons. tinyBenchmarks selects compact subsets to approximate full benchmark performance \citep{polo2024tinybenchmarks}. Anchor~Points identifies representative examples for data-efficient evaluation \citep{vivek2024anchorpoints}. SubLIME considers subset selection based on model-ranking preservation \citep{saranathan2025sublime}. Compact evaluation sets can support reliable model comparison. Symbolic regression introduces additional requirements because task structures and algorithm behaviors are heterogeneous.

\paragraph{Dynamic and Multi Axis Evaluation.}
Recent benchmarking studies assess model behavior beyond a single aggregate score. Dynabench introduced evaluation that evolves alongside model development \citep{kiela2021dynabench}. Dynaboard extended dynamic benchmarking toward multidimensional evaluation and standardized comparison across systems \citep{ma2021dynaboard}. DataPerf emphasized reproducible and data-centric benchmarking protocols \citep{mazumder2023dataperf}. SR benchmarks also provide numerical, symbolic, and computational diagnostics \citep{lacava2021contemporary,matsubara2024rethinking,shojaee2025llmsrbench}. Existing evaluation focuses largely on final outcomes using protocols specific to each benchmark. \arena{} complements existing protocols with standardized execution and Multi Axis Evaluation. The framework incorporates search trajectories and failure analysis.

%% file: sections/appB_reservoir.tex
\section{\reservoir{} Construction and Standardization}
\label{app:reservoir}

\reservoir{} is a population of 664 symbolic regression tasks collected from multiple benchmark sources. Construction covers task admission, representation standardization, and validation. Duplicate control and composition auditing complete the process. Ground truth formulas and static metadata are used during \reservoir{} construction. Algorithm response features are excluded.

\subsection{Task Sources and Admission} %
\label{app:reservoir-filtering}

The initial pool exceeds 800 symbolic regression tasks collected from modern benchmarks and classical equation sets.
Modern sources include SRSD~\citep{matsubara2024rethinking} and LLM-SRBench~\citep{shojaee2025llmsrbench}.
SRBench and its 2025 extension provide further tasks \citep{lacava2021contemporary,SRBench2025call}.
Classical sources comprise the Keijzer~\citeyearpar{keijzer2003improving}, Korns~\citeyearpar{korns2011accuracy}, and Vladislavleva~\citeyearpar{vladislavleva2009order} benchmarks. The Nguyen equation set provides further tasks \citep{uy2011semantically}.

Admission to \reservoir{} requires all four conditions below.
\begin{itemize}
	\item The ground truth formula is unambiguous, parseable as a symbolic expression tree, and executable as a numerical program.
	\item Variables, constants, and operators are resolvable and consistent with the dataset columns.
	\item Train, ID test, and OOD test splits are available or reproducibly generated from the executable ground truth program.
	\item Metadata for provenance, structural analysis, and split reconstruction are available. Duplicate control metadata are also available.
\end{itemize}

Tasks with ambiguous or unresolvable definitions are excluded. Admission yields the 664 tasks in \reservoir{}.

\subsection{Unified Task Representation} %
\label{app:reservoir-schema-splits}

Each admitted task is converted to a common executable schema. The main field groups are summarized in \Cref{tab:reservoir-schema}.

\begin{table}[H]
	\centering
	\small
	\caption{\textbf{Unified schema of \reservoir{}.}}
	\label{tab:reservoir-schema}
	\begin{tabular}{p{0.25\linewidth}p{0.68\linewidth}}
		\toprule
		\textbf{Field group} & \textbf{Stored information} \\
		\midrule
		Identity and provenance &
		\texttt{dataset\_id}, source family, and subgroup.\newline Original basename and source benchmark \\
		
		Variables and target &
		Ordered feature names and target name.\newline Number of features and dummy variable indicator \\
		
		Ground truth expression &
		Raw source formula and executable expression.\newline Normalized symbolic forms and expression tree \\
		
		Data splits &
		Fixed train, ID test, and OOD test samples.\newline Split sizes and parameters required for deterministic regeneration \\
		
		Structural metadata &
		Formula complexity and operator types.\newline Variable count bin and OOD type \\
		
		Duplicate metadata &
		Semantic duplicate group identifier and provenance links used for duplicate auditing \\
		\bottomrule
	\end{tabular}
\end{table}

The raw source formula is preserved for provenance. A normalized symbolic representation is generated separately for structural comparison across sources. Expression normalization provides a unified representation by standardizing equivalent expressions and enforcing deterministic serialization.

Structural analysis uses the resulting expression strings, tree representations, and normalized operator forms. Additional representations are retained for duplicate detection and recovery analysis. The representations support symbolic equivalence analysis, tree comparison, and structural consistency analysis. Recovery analysis covers variables and operators.

The training split serves as the input for symbolic expression search. The ID test split measures generalization within the observed training distribution, and the OOD test split measures extrapolation beyond it. Compatible official splits are retained. Missing splits are generated deterministically from the executable ground truth program. Task metadata record the parameters required to regenerate the splits.

All materialized splits remain fixed and are shared across algorithms. Official numerical metrics are computed on the frozen ID and OOD test splits.

\paragraph{Validation and duplicate control.}
All standardized tasks are validated by replaying executable ground truth expressions, checking fixed data splits, and reproducing the normalized symbolic representations. Tasks with unresolved inconsistencies are excluded. Semantically equivalent or closely related tasks are assigned a shared \texttt{semantic\_duplicate\_group}, which prevents multiple members of one group from entering \core{}.
The resulting \reservoir{} forms an executable and reproducible population for subsequent benchmark construction.

\subsection{\reservoir{} Composition} %
\label{app:reservoir-composition}

The 664 tasks cover eight benchmark sources, multiple formula types, and
three complexity levels, as summarized in \Cref{tab:reservoir-composition}.
All statistics are derived solely from task formulas and static metadata and
define the population available for subsequent \core{} distillation.

\begin{table}[htbp]
    \centering
    \small
    \caption{\textbf{Composition of \reservoir{} (664 tasks).}}
    \label{tab:reservoir-composition}
    \renewcommand{\arraystretch}{1.15}
    \begin{tabularx}{\linewidth}{@{}>{\raggedright\arraybackslash}p{0.21\linewidth}>{\raggedright\arraybackslash}X@{}}
        \toprule
        \textbf{Dimension} & \textbf{Categories and task counts} \\
        \midrule
        \textbf{Task sources} &
        LLM-SRBench (240), SRSD (232), SRBench 1.0 (133), Keijzer (15),
        SRBench 2025 first-principles subset (12), Nguyen (12), Korns (12),
        Vladislavleva (8) \\
        \addlinespace[3pt]
        \textbf{Formula types} &
        Rational (282), Trigonometric (137), Polynomial (124),
        Exponential / logarithmic (86), Mixed (34), Unknown (1) \\
        \addlinespace[3pt]
        \textbf{Formula complexity} &
        Simple (284), Moderate (188), Complex (192) \\
        \bottomrule
    \end{tabularx}
    \par\smallskip
    \begin{minipage}{\linewidth}
        \footnotesize
        Counts appear in parentheses. Each task belongs to one category
        within each dimension according to its recorded metadata.
    \end{minipage}
\end{table}

\paragraph{Operator count complexity.}
Formula complexity is characterized using an operator count extracted from the
Python abstract syntax tree of each task's \texttt{formula.py}.
For task $i$, the count is defined as
\[
C_i =
N_i^{\mathrm{BinOp}}
+
N_i^{\mathrm{UnaryOp}}
+
N_i^{\mathrm{Compare}},
\]
where $N_i^{\mathrm{BinOp}}$, $N_i^{\mathrm{UnaryOp}}$, and
$N_i^{\mathrm{Compare}}$ denote the numbers of binary operation, unary
operation, and comparison nodes in the complete source file.
Function calls are excluded from the operator count, and auxiliary functions
in \texttt{formula.py} contribute to $C_i$ when they contain any counted node
type.
For example, the stored implementation of Keijzer-1 has $C_i=7$.
The target expression $0.3x_1\sin(2\pi x_1)$ contributes four binary operation
nodes, and auxiliary logic in the same source file contributes three comparison
nodes, placing the task in the Moderate group.
The 664 tasks are partitioned according to the empirical one third and two
thirds quantiles of $C_i$, yielding boundaries of 5 and 8
\[
\mathrm{Simple}: C_i \leq 5,\qquad
\mathrm{Moderate}: 6 \leq C_i \leq 8,\qquad
\mathrm{Complex}: C_i \geq 9.
\]
Tasks with equal operator counts remain in the same group, producing 284
Simple, 188 Moderate, and 192 Complex tasks.
The unequal group sizes arise from ties near the quantile boundaries.
The resulting categories describe relative operator count complexity in the
stored source implementation.
Function calls such as \texttt{sin} and \texttt{exp} receive no additional
complexity weight, and auxiliary implementation logic can affect the count.
The categories serve as static descriptive metadata for \reservoir{}
composition, separate from empirical task difficulty.

%% file: sections/appC_distillation.tex
\section{\core{} Distillation and Validation}
\label{app:core50-details}

\core{} construction proceeds from \calibset{} mining to \probe{} selection and final task selection.
Construction records are separated from fidelity validation and formal evaluation.

\subsection{\calibset{} Mining}
\label{app:candidate200-audit}

\calibset{} contains 200 tasks selected to expose informative differences among SR algorithms.
PySR and LLM-SR serve as initial probes, with one evaluation per probe on every task in \reservoir{}.
Calibration selection is separate from the later selection of \probe{}.

For task $i$ and algorithm $a$, let $\ell_{i,a}^{\mathrm{id}}$ and
$\ell_{i,a}^{\mathrm{ood}}$ denote the clipped $\log \mathrm{NMSE}$ on the
ID and OOD test splits.
The performance gaps are
\[
g_i^{id}
=
\ell_{i,\mathrm{LLMSR}}^{id}
-
\ell_{i,\mathrm{PySR}}^{id},
\qquad
g_i^{ood}
=
\ell_{i,\mathrm{LLMSR}}^{ood}
-
\ell_{i,\mathrm{PySR}}^{ood}.
\]
A positive gap indicates an advantage for PySR, and a negative gap indicates
an advantage for LLM-SR.
The disagreement score combines the ID and OOD differences
\[
G_i
=
\frac{|g_i^{id}|+|g_i^{ood}|}{2}.
\]

\calibset{} contains three diagnostic groups.
The high disagreement group contributes 140 tasks from the largest response gaps.
One sided evaluability contributes 20 tasks for which only one initial probe obtains a successful solution.
The moderate gap group contributes 40 tasks from intermediate disagreement regions and broadens coverage beyond extreme cases.

Across source families, \calibset{} contains 70 tasks from SRSD, 70 from LLM-SRBench, and 33 from SRBench1.0.
Korns contributes 9 tasks.
Keijzer, Nguyen, and SRBench2025 contribute 6 tasks each.

\begin{figure}[htbp]
    \centering
    \includegraphics[width=0.68\linewidth]
    {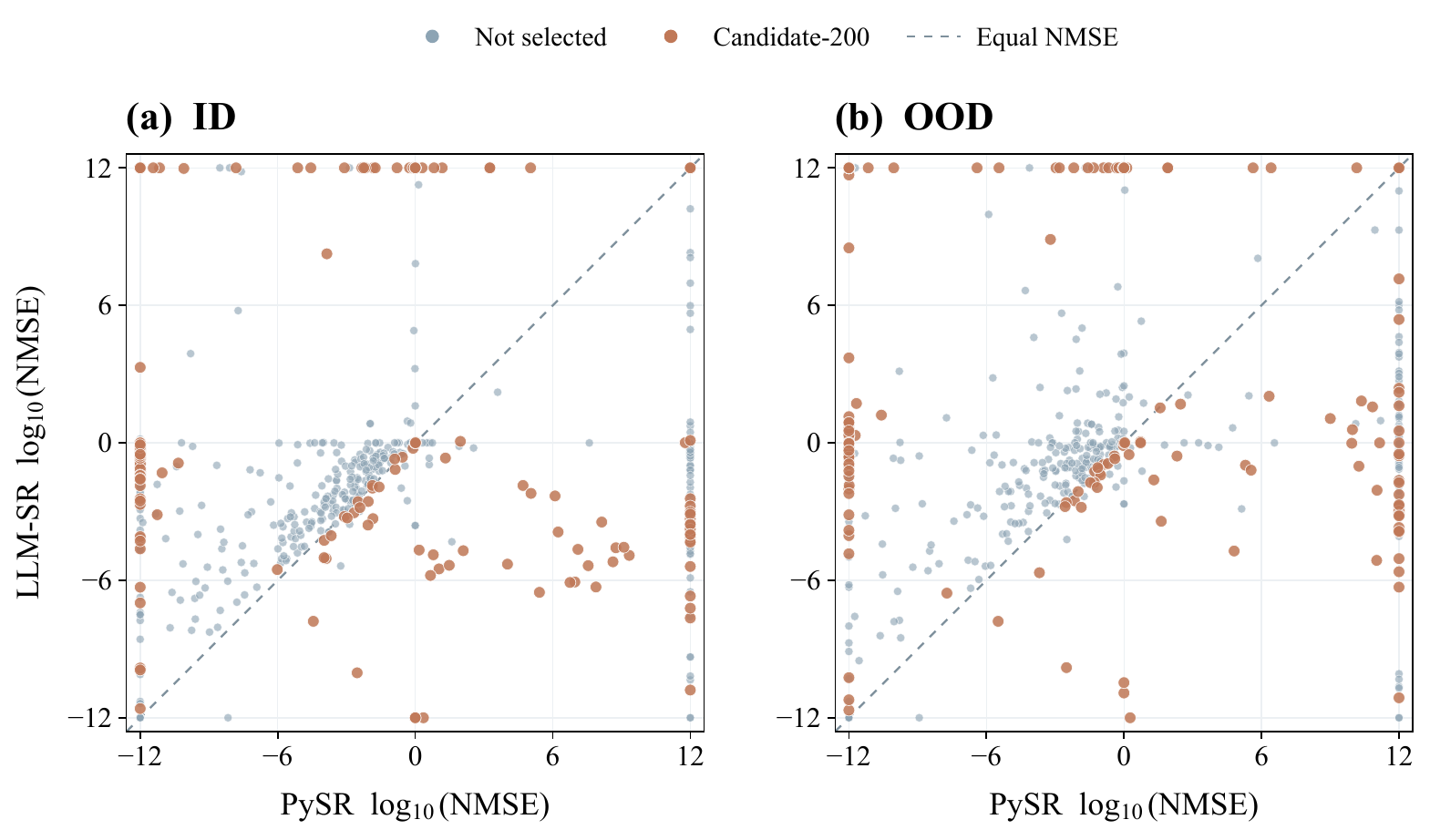}
    \caption{\textbf{Dual probe mining diagnostic.}
    Each point represents one task in the PySR and LLM-SR response space.
    Off diagonal points indicate large performance differences.
    Selected \calibset{} tasks cover distinct ID and OOD response regions.}
    \label{fig:dual-probe-scatter}
\end{figure}

\Cref{fig:dual-probe-scatter} shows that \calibset{} retains highly separable tasks and less extreme response regions.

\subsection{Selection of Four Probes}
\label{app:probe4-selection}

Probe selection uses one recorded run for each of 12 algorithms on the 200 calibration tasks.
The objective favors reliable responses subject to coverage of at least three primary paradigms.
The criterion prioritizes reliable response availability and coverage across search paradigms.
Task discriminability is measured later from the selected probes on \reservoir{}, keeping calibration usability separate from final task selection.

\paragraph{Calibration usability.}
Let $\mathcal{D}_{\mathrm{cal}}$ contain the $N=200$ calibration tasks, and
let $e_{i,a}^{s}$ denote the NMSE for algorithm $a$ on task $i$ and split $s$.
The tasks with jointly available errors are
\[
\mathcal{Q}_a=
\bigl\{
i\in\mathcal{D}_{\mathrm{cal}}:
e_{i,a}^{s}\in\mathbb{R}_{\geq0}
\text{ for all }
s\in\{\mathrm{train},\mathrm{ID},\mathrm{OOD}\}
\bigr\}.
\]
Membership in $\mathbb{R}_{\geq0}$ requires a finite and nonnegative value.
The finite result rate is
\[
q(a)=\frac{|\mathcal{Q}_a|}{N}.
\]

For a positive threshold $\tau$, define the explosion tasks as
\[
\mathcal{B}_{a,\tau}
=
\bigl\{
i\in\mathcal{D}_{\mathrm{cal}}:
\exists s\in\{\mathrm{ID},\mathrm{OOD}\},
e_{i,a}^{s}\in\mathbb{R}_{\geq0},
e_{i,a}^{s}>\tau
\bigr\}.
\]
The explosion rate is
\[
b_\tau(a)=\frac{|\mathcal{B}_{a,\tau}|}{N}.
\]

Both rates use all 200 tasks as the denominator.
An unavailable error on one split and an explosion on another can affect both rates.
Reaching the time limit does not reduce usability if evaluable outputs remain available.

Calibration usability is
\[
H_\alpha(a;\tau)
=
\alpha q(a)
+
(1-\alpha)\bigl[1-b_\tau(a)\bigr],
\qquad
0\leq\alpha\leq1.
\]
The default values are $\alpha=0.5$ and $\tau=100$.
The score combines numerical availability and resistance to numerical explosion.

\begin{figure}[htbp]
    \centering
    \includegraphics[width=0.6\linewidth]{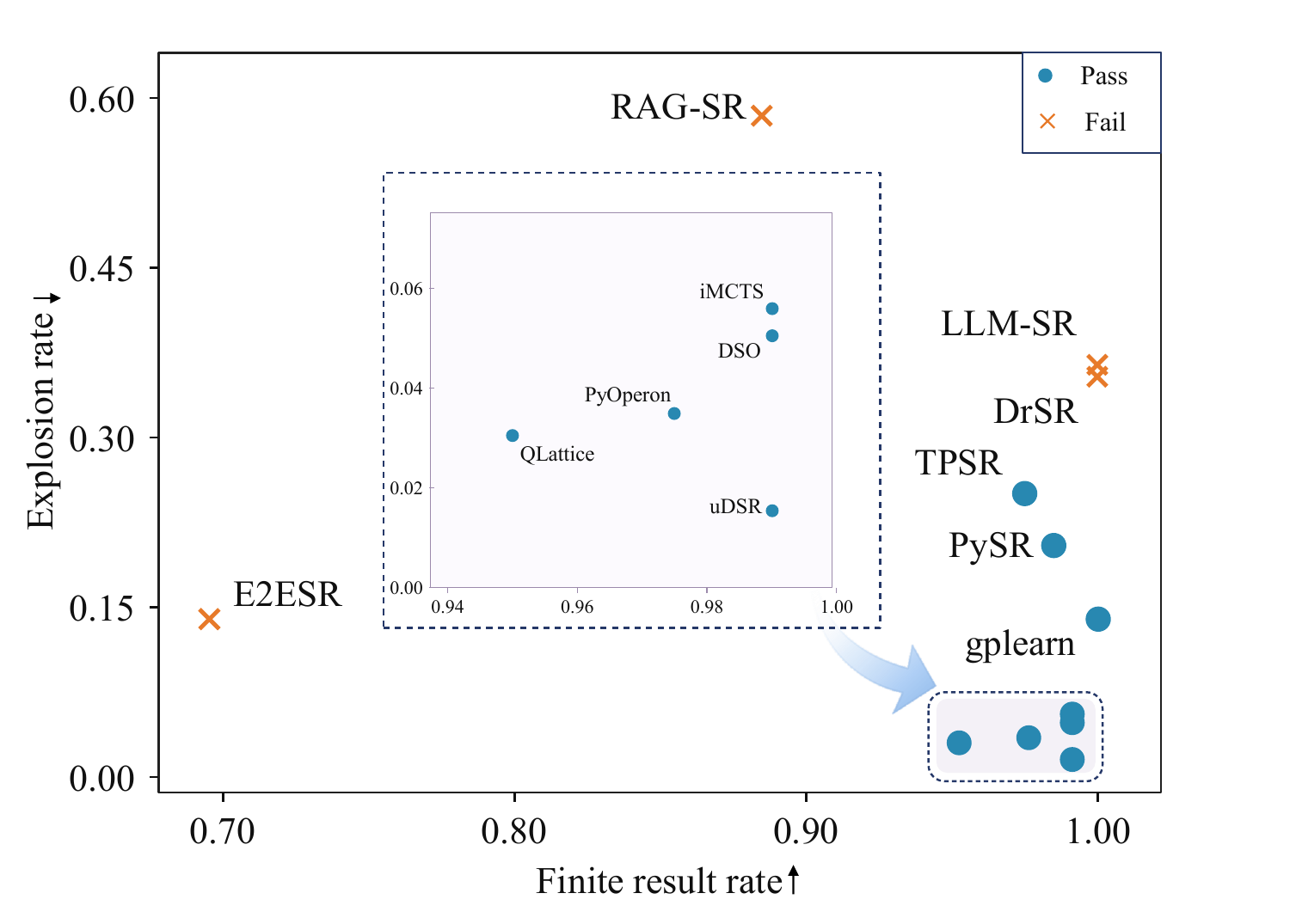}
    \caption{\textbf{Calibration usability of candidate algorithms.}
    Points show finite result rate $q(a)$ and explosion rate $b_{100}(a)$ on \calibset{}.
    The reference region marks high availability and low explosion rates.
    All 12 algorithms remain eligible for probe selection.}
    \label{fig:probe-usability-candidates}
\end{figure}

For visualization, $q(a)\geq0.9$ and $b_{100}(a)\leq0.3$ define a reference region.
\Cref{fig:probe-usability-candidates} visualizes the calibration usability of all 12 candidates at the default setting.
All 12 algorithms remain eligible in the subsequent subset search.

\paragraph{Probe objective and paradigm coverage.}
Let $\mathcal{A}_{12}$ denote the calibration algorithms and $g(a)$ denote the
primary paradigm label used in \Cref{tab:six_axis_profiles}.
DSO and uDSR belong to Neural and Policy Search.
iMCTS and QLattice belong to Structured Search.
The labels represent primary categories and do not imply mutually exclusive mechanisms.

For a minimum of $m$ primary paradigms, feasible subsets are
\[
\mathcal{P}_m=
\left\{
P\subseteq\mathcal{A}_{12}:
|P|=4,\quad
|\{g(a):a\in P\}|\geq m
\right\}.
\]
Subset usability is
\[
U(P;\alpha,\tau)
=
\frac{1}{|P|}
\sum_{a\in P}H_\alpha(a;\tau).
\]
The default rule selects
\[
\probepanel
\in
\operatorname*{arg\,max}_{P\in\mathcal{P}_3}
U(P;0.5,100).
\]

All $\binom{12}{4}=495$ subsets are enumerated, and 468 satisfy the three paradigm requirement.
Exact rational arithmetic preserves tied scores.
A four algorithm subset covering at least three paradigms contains at most two members from one primary paradigm.

\begin{table}[htbp]
    \centering
    \small
    \setlength{\tabcolsep}{4pt}
    \caption{\textbf{Highest scoring feasible four probe combinations.}
    Mean calibration usability uses $\alpha=0.5$ and $\tau=100$.
    Each reported subset covers three primary paradigms.}
    \label{tab:probe4-combination-audit}
    \begin{tabularx}{\linewidth}{@{}r >{\raggedright\arraybackslash}X r@{}}
        \toprule
        \textbf{Rank} & \textbf{Candidate combination} & \textbf{Mean usability}\\
        \midrule
        1 & \textbf{DSO, PyOperon, iMCTS, uDSR} & \textbf{0.971250}\\
        2 & DSO, PyOperon, QLattice, uDSR & 0.969375\\
        3 & PyOperon, iMCTS, QLattice, uDSR & 0.968750\\
        \bottomrule
    \end{tabularx}
\end{table}

\Cref{tab:probe4-combination-audit} gives the unique optimum
\[
\probepanel=
\{\mathrm{DSO},\mathrm{PyOperon},\mathrm{iMCTS},\mathrm{uDSR}\}.
\]
The selected subset covers three primary paradigms.
Sensitivity analysis appears in \Cref{app:sensitivity}.

\subsection{\core{} Construction}
\label{app:core50-scoring-search}
\core{} selects 50 tasks from \reservoir{} using probe responses and task descriptors.
The same eligibility rules apply to every source family.

\paragraph{Construction records and aggregation.}
Each record contains ID and OOD NMSE, output validity, and run status.
Valid expressions remain evaluable after timeout, while missing or invalid outputs retain explicit annotations.

Finite nonnegative NMSE values use
\[
\ell(x)=
\operatorname{clip}
\left(
\log_{10}(\max(x,10^{-12})),
-12,
12
\right).
\]
For each task $i$ and probe $p$, let
$\mathcal S_{i,p}^{\mathrm{valid}}$
contain seeds with valid evaluable outputs.
For split $a\in\{\mathrm{id},\mathrm{ood}\}$, the median log error and seed variability are
\[
m_{i,p}^{a}
=
\operatorname{median}_{s\in\mathcal S_{i,p}^{\mathrm{valid}}}
\ell(\mathrm{NMSE}_{i,p,s}^{a}),
\qquad
q_{i,p}^{a}
=
\operatorname{IQR}_{s\in\mathcal S_{i,p}^{\mathrm{valid}}}
\ell(\mathrm{NMSE}_{i,p,s}^{a}).
\]
The valid fraction $v_{i,p}$ uses all $K=3$ scheduled seeds.
An empty valid seed set leaves the corresponding median and IQR unavailable.

\paragraph{Task descriptors and information.}
Structural descriptors encode task properties and redundancy information.
Response descriptors encode probe performance, validity, and seed variation.

Let $\mathcal N$ denote normalization to $[0,1]$, and let $H_i$ measure the entropy of the valid output pattern across probes.
The discrimination score is
\[
\begin{aligned}
\mathrm{Disc}_i={}&
0.45\,\mathcal N
\left(
\operatorname{Var}_p(m^{\mathrm{id}}_{i,p})
\right)
+
0.45\,\mathcal N
\left(
\operatorname{Var}_p(m^{\mathrm{ood}}_{i,p})
\right)
\\
&+
0.10\,\mathcal N(H_i).
\end{aligned}
\]

Construction instability and stability are
\[
u_i
=
\operatorname{Mean}_p
\left[
q^{\mathrm{id}}_{i,p}
+
q^{\mathrm{ood}}_{i,p}
+
1-v_{i,p}
\right],
\qquad
\mathrm{Stab}_i
=
1-\mathcal N(u_i).
\]
Task information combines discrimination and stability
\[
\mathrm{Info}_i
=
\mathrm{Disc}_i
\,
\mathrm{Stab}_i.
\]
A high information score requires probe separation and consistency across seeds.
Construction stability is distinct from the final evaluation axis \stab{}.

\paragraph{Difficulty, failure mode, and eligibility.}
Tasks with unusable probe responses are excluded.
Partially failed or high variance tasks remain eligible subject to category limits.
A timeout alone does not determine eligibility if evaluable outputs remain available.

Difficulty is
\[
\mathrm{Difficulty}_i
=
\operatorname{Mean}_p
\left[
\frac{
m^{\mathrm{id}}_{i,p}
+
m^{\mathrm{ood}}_{i,p}
}{2}
\right].
\]
Difficulty thresholds define easy, medium, hard, and extreme tasks.
Validity and ID to OOD degradation define failure annotations.

\paragraph{\core{} objective and constraints.}
Coverage and mean task information are
\[
\begin{aligned}
\mathrm{Coverage}(S)
&=
0.6\,\mathrm{StructuralCoverage}(S)
+
0.4\,\mathrm{ResponseCoverage}(S),
\\
\mathrm{MeanInfo}(S)
&=
\frac{1}{|S|}
\sum_{i\in S}\mathrm{Info}_i.
\end{aligned}
\]

For feasible subsets satisfying $|S|=50$, the objective combines coverage,
mean task information, and distributional balance
\[
\boxed{
J(S)
=
0.45\,\mathrm{Coverage}(S)
+
0.35\,\mathrm{MeanInfo}(S)
+
0.20\,\mathrm{Balance}(S)
}.
\]

The feasible set requires
\begin{itemize}
    \item exactly 50 tasks
    \item at most one task per semantic duplicate group
    \item at most one task per basename
    \item at most 6 tasks per subgroup
    \item family counts within the shared quota formula
    \item 4 to 10 easy tasks, 14 to 24 medium tasks, 12 to 22 hard tasks, and 3 to 6 extreme tasks
    \item at most 8 limited quota tasks
    \item at most 2 one sided tasks
    \item at most 5 unstable tasks
    \item at most 5 all struggle tasks
    \item 3 to 20 tasks per winning probe
\end{itemize}

Let $n=50$, $N=664$, and $F$ denote the number of source families.
Let $n_f$ denote the number of \reservoir{} tasks from family $f$.
Family targets use
\[
t_f
=
n
\left(
\beta\frac{n_f}{N}
+
(1-\beta)\frac1F
\right),
\qquad
\beta=0.6,
\]
with bounds
\[
q_f^{\min}
=
\max(1,\lfloor t_f-1\rfloor),
\qquad
q_f^{\max}
=
\lceil t_f+2\rceil.
\]
The same quota rule and eligibility criteria apply to every family.

\paragraph{Search and freezing.}
Greedy initialization constructs feasible subsets using coverage and task information.
Feasible single task swaps refine the objective across multiple initializations.
The highest observed feasible solution is frozen as \core{} before fidelity validation.

\subsection{Selection Baselines and Fidelity Validation}
\label{app:core50-validation}

We compare the frozen \core{} with six alternative selectors that each produce 50 tasks.

\begin{itemize}[leftmargin=*]
\item \textbf{random-50} selects 50 distinct dataset IDs uniformly from all 664 tasks.

\item \textbf{family-stratified random-50} samples distinct dataset IDs within source families using the smoothed targets $t_f$ defined above.
Largest remainder rounding assigns integer quotas totaling 50 tasks.

\item \textbf{metadata-diverse-50} selects diverse tasks using metadata only.
Inputs include source family, feature count, and formula complexity.
Sample count is also included.

\item \textbf{response-space K-medoids-50} selects medoids in the response space of \probe{}.
The response representation contains ID and OOD behavior, validity, and stability.

\item \textbf{top-information-50} selects the 50 tasks with the largest
$\mathrm{Info}_i$ values.

\item \textbf{difficulty-balanced-50} applies quotas across difficulty bins and uses task information to break ties.
\end{itemize}

\paragraph{Aggregate score fidelity.}
For task $i$ and probe $p$, let $K$ be the number of scheduled runs and
$f_{i,p}$ the number of runs with no valid evaluable output.
For a task and probe pair with at least one valid run,
\[
\bar m_{i,p}
=
\frac{
m^{\mathrm{id}}_{i,p}
+
m^{\mathrm{ood}}_{i,p}
}{2}.
\]
The task score is
\[
z_{i,p}
=
-\bar m_{i,p}
-
\lambda\frac{f_{i,p}}{K},
\qquad
f_{i,p}<K.
\]
The evaluation uses $K=3$ and $\lambda=2$.
A task and probe pair with no valid run receives $z_{i,p}=-12$.

Probe scores are averaged across all tasks in $S$
\[
A_p(S)
=
\frac{1}{|S|}
\sum_{i\in S}z_{i,p}.
\]
Aggregate fidelity is measured by
\[
\boxed{
\operatorname{MAE}(S,\fullset)
=
\frac{1}{|\probepanel|}
\sum_{p\in\probepanel}
\left|
A_p(S)
-
A_p(\fullset)
\right|,
\qquad
|\probepanel|=4
}.
\]
All selectors use the same construction records and scoring rule.
NMAE divides MAE by the range of the four aggregate probe scores on $\fullset$.
At the default setting, \core{} achieves an MAE of $0.1388$ and an NMAE of $2.33\%$.

Penalty sensitivity is evaluated at
$\lambda\in\{0,0.5,1,2,4\}$.
\core{} membership and the $-12$ assignment for zero valid runs remain fixed.
All five settings preserve the probe ordering on \reservoir{}.
NMAE ranges from $2.29\%$ to $2.55\%$.

\paragraph{Repeated random selection.}
Each random selector is evaluated on 100 independently sampled subsets of 50 tasks.
MAE is computed separately for each draw using fixed probe responses.
\Cref{tab:core50-repeated-selection} reports the mean and sample standard deviation.
Empirical 2.5th and 97.5th percentiles describe variation across sampled subsets.

Two of the 100 uniform draws obtain lower MAE than the fixed \core{}.
No draw from the smoothed family stratified selector reaches the \core{} value in the current sample.
A proportional family quota check yields a mean MAE of $0.4931$ and a standard deviation of $0.1959$.
Complete memberships and per draw results accompany the supplementary data.

The remaining four selectors are reported as fixed point estimates.

\input{tables/Appendix/core50_selector_repeats}

\paragraph{Ranking validation.}
Fidelity is also assessed using algorithm rankings.
Validation considers Spearman and Kendall correlations and pairwise order agreement.
The four construction probes define six unordered pairs.
Validation algorithms excluded from benchmark construction provide additional evidence beyond \probe{}.
Ranking results are reported in the main experiments.

\subsection{\core{} Manifest}
\label{app:core50-manifest}

\Cref{tab:table18_core50_ground_truth_manifest} presents the \core{} membership and ground truth expressions.
\core{} includes 16 LLM-SRBench tasks, 15 SRSD tasks, 9 SRBench1.0 tasks, 3 Nguyen tasks, 2 each from SRBench2025, Korns, and Keijzer, and 1 Vladislavleva task.

\input{tables/Appendix/table18_core50_ground_truth_manifest}

%% file: tables/Appendix/core50_selector_repeats.tex
\begin{table}[htbp]
\centering
\small
\setlength{\tabcolsep}{5pt}
\renewcommand{\arraystretch}{1.12}
\caption{\textbf{Aggregate score fidelity of 50 task selectors.}
Random selectors report mean MAE $\pm$ standard deviation and empirical 95\% ranges across 100 draws from \reservoir{}.
Other selectors report point estimates.
\core{} is fixed, and the final row gives a proportional family quota check.}
\label{tab:core50-repeated-selection}
\begin{tabularx}{\linewidth}{@{}Xcc@{}}
\toprule
\textbf{Selector} & \textbf{MAE} & \textbf{Empirical 95\% range} \\
\midrule
Uniform random & $0.5298\pm 0.2550$ & $[0.1605,\,1.1819]$ \\
Smoothed family stratified & $0.5438\pm 0.2072$ & $[0.2257,\,1.0102]$ \\
Metadata diverse & $0.5060$ & \textemdash \\
Response K-medoids & $0.6044$ & \textemdash \\
Top information & $1.0339$ & \textemdash \\
Difficulty balanced & $1.0430$ & \textemdash \\
\rowcolor{gray!15} \textbf{\core{}} & $\mathbf{0.1388}$ & \textemdash \\
\midrule
Proportional family quota check & $0.4931\pm 0.1959$ & $[0.2459,\,0.9728]$ \\
\bottomrule
\end{tabularx}
\end{table}

%% file: tables/Appendix/table18_core50_ground_truth_manifest.tex
\begingroup
\small
\setlength{\tabcolsep}{3pt}
\renewcommand{\arraystretch}{1.3}
\setlength{\emergencystretch}{3em}
\arrayrulecolor{black!15}
\setlength{\aboverulesep}{0pt}
\setlength{\belowrulesep}{0pt}
\sloppy
\begin{longtable}{@{}>{\centering\arraybackslash}m{0.05\textwidth} >{\raggedright\arraybackslash}m{0.27\textwidth} >{\raggedright\arraybackslash}m{0.20\textwidth} >{\raggedright\arraybackslash}m{0.44\textwidth}@{}}
\caption{\textbf{\core{} ground truth formula manifest.} Tasks are grouped by source family. Variables follow the recorded order, and constants use at most four decimal places. Exact expressions and task identifiers are provided in the released metadata.}
\label{tab:table18_core50_ground_truth_manifest}\\
\toprule
\textbf{Idx} & \textbf{Task} & \textbf{Variables} & \textbf{Ground Truth} \\
\midrule
\endfirsthead
\caption[]{\textbf{\core{} ground truth formula manifest (continued).}}\\
\toprule
\textbf{Idx} & \textbf{Task} & \textbf{Variables} & \textbf{Ground Truth} \\
\midrule
\endhead
\bottomrule
\endlastfoot
\rowcolor{black!8}\multicolumn{4}{@{}l}{\textbf{SRSD} \quad 15 tasks}\\
\addlinespace[2pt]
1 & Feynman II.27.18 & $x_0,x_1,x_2$ & $8.854\times10^{-12}x_0^2$ \\
\midrule[0.3pt]
2 & Feynman I.43.31 & $x_0,x_1$ & $1.3806\times10^{-23}x_0x_1$ \\
\midrule[0.3pt]
3 & Feynman II.34.11 & $x_0,\allowbreak\dots,\allowbreak x_6$ & $\frac{x_0x_2x_4}{2x_5}$ \\
\midrule[0.3pt]
4 & Feynman I.18.16 & $x_0,\dots,x_5$ & $x_0x_1x_3\sin x_4 $ \\
\midrule[0.3pt]
5 & Feynman II.34.2a & $x_0,\dots,x_4$ & $\frac{x_1x_3}{2\pi x_4}$ \\
\midrule[0.3pt]
6 & Feynman I.39.22 & $x_0,\dots,x_4$ & $\frac{1.3806\times10^{-23}x_1x_3}{x_4}$ \\
\midrule[0.3pt]
7 & Feynman I.11.19 & $x_0,\allowbreak\dots,\allowbreak x_7$ & $x_0x_1+x_2x_3+x_5x_6$ \\
\midrule[0.3pt]
8 & Feynman II.4.23 & $x_0,x_1$ & $2.824\times10^{10}\,\frac{x_0}{\pi x_1}$ \\
\midrule[0.3pt]
9 & Feynman III.7.38 & $x_0,x_1$ & $6.0368\times10^{33}\,\pi x_0x_1$ \\
\midrule[0.3pt]
10 & Feynman I.13.12 & $x_0,\dots,x_3$ & $6.6743\times10^{-11}x_0x_1\left(\frac{1}{x_2}-\frac{1}{x_3}\right)$ \\
\midrule[0.3pt]
11 & Feynman III.10.19 & $x_0,\dots,x_5$ & $x_2\sqrt{x_3^2+x_4^2+x_5^2}$ \\
\midrule[0.3pt]
12 & Feynman I.32.5 & $x_0,x_1$ & $\frac{6.9862\times10^{-16}x_0^2x_1^2}{\pi}$ \\
\midrule[0.3pt]
13 & Feynman I.29.16 & $x_0,\dots,x_3$ & $\sqrt{x_0^2+2x_0x_1\cos(x_2-x_3)+x_1^2}$ \\
\midrule[0.3pt]
14 & Feynman I.15.3t & $x_0,\dots,x_4$ & $\frac{x_1-1.1126\times10^{-17}x_3x_4}{\sqrt{1-1.1126\times10^{-17}x_3^2}}$ \\
\midrule[0.3pt]
15 & Feynman Bonus 20 & $x_0,\dots,x_5$ & $\frac{7.8370\times10^{-29}x_2^2\left(\frac{x_2}{x_3}-\sin^2x_5+\frac{x_3}{x_2}\right)}{\pi x_3^2}$ \\
\addlinespace[4pt]
\rowcolor{black!8}\multicolumn{4}{@{}l}{\textbf{SRBench 1.0} \quad 9 tasks}\\
\addlinespace[2pt]
16 & Feynman I.27.6 & $d_1,d_2,n$ & $\frac{1}{\frac{1}{d_1}+\frac{n}{d_2}}$ \\
\midrule[0.3pt]
17 & Feynman II.15.4 & $mom,B,\theta$ & $-mom\,B\cos\theta$ \\
\midrule[0.3pt]
18 & Strogatz Shear Flow 1 & $x,y$ & $\frac{\cos x}{\tan y}$ \\
\midrule[0.3pt]
19 & Strogatz Predator Prey 2 & $x,y$ & $y\left(\frac{x}{1+x}-0.075y\right)$ \\
\midrule[0.3pt]
20 & Strogatz Bar Mag 2 & $x,y$ & $0.5\sin(y-x)-\sin y$ \\
\midrule[0.3pt]
21 & Strogatz Bar Mag 1 & $x,y$ & $0.5\sin(x-y)-\sin x$ \\
\midrule[0.3pt]
22 & Feynman I.12.11 & $q,\allowbreak E_f,\allowbreak B,\allowbreak v,\allowbreak\theta$ & $q\left(E_f+Bv\sin\theta\right)$ \\
\midrule[0.3pt]
23 & Strogatz Van der Pol 1 & $x,y$ & $10\left(y-\frac{1}{3}x^3+\frac{1}{3}x\right)$ \\
\midrule[0.3pt]
24 & Feynman Test 15 & $c,\allowbreak v,\allowbreak\omega,\allowbreak\theta$ & $\frac{\sqrt{1-v^2/c^2}\,\omega}{1+\frac{v}{c}\cos\theta}$ \\
\addlinespace[4pt]
\rowcolor{black!8}\multicolumn{4}{@{}l}{\textbf{SRBench 2025} \quad 2 tasks}\\
\addlinespace[2pt]
25 & Hubble Law & $D$ & $73.3\,D$ \\
\midrule[0.3pt]
26 & Leavitt Law & $\log P$ & $-2.084\log P+15.65$ \\
\addlinespace[4pt]
\rowcolor{black!8}\multicolumn{4}{@{}l}{\textbf{LLM-SRBench} \quad 16 tasks}\\
\addlinespace[2pt]
27 & II.34.2 & $mom,\,q,\,r$ & $\frac{2\,mom}{q\,r}$ \\
\midrule[0.3pt]
28 & III.21.20 & $j,\allowbreak\rho_{c,0},\allowbreak q,\allowbreak A_{\mathrm{vec}}$ & $-\frac{A_{\mathrm{vec}}\,q\,\rho_{c,0}}{j}$ \\
\midrule[0.3pt]
29 & I.11.19 & $A,\allowbreak x_1,\allowbreak x_2,\allowbreak x_3,\allowbreak y_1,\allowbreak y_3$ & $\frac{-A+x_1y_1-x_3y_3}{x_2}$ \\
\midrule[0.3pt]
30 & II.11.27 & $P_{ol},\allowbreak n,\allowbreak\varepsilon,\allowbreak E_f$ & $\frac{3P_{ol}}{n(3E_f\varepsilon+P_{ol})}$ \\
\midrule[0.3pt]
31 & Chemical Reaction 11 & $t,A$ & $-0.8817A^2+0.8817\sin\sqrt{A}$ \\
\midrule[0.3pt]
32 & Chemical Reaction 10 & $t,A$ & $-0.1749A^2+0.1749\sin\ln(A+1)$ \\
\midrule[0.3pt]
33 & Chemical Reaction 34 & $t,A$ & $-0.1689A^2+0.1689A^{0.3333}$ \\
\midrule[0.3pt]
34 & Physical Oscillation 19 & $x,t,v$ & $-0.2v-x\,e^{-|x|}$ \\
\midrule[0.3pt]
35 & Chemical Reaction 0 & $t,A$ & $-0.1899A^2+\frac{0.1899A^2}{0.7497A^4+1}$ \\
\midrule[0.3pt]
36 & Physical Oscillation 41 & $x,t,v$ & $-0.3254(1-x^2)v-0.3333^2x\,e^{-|x|}$ \\
\midrule[0.3pt]
37 & Chemical Reaction 33 & $t,A$ & $-0.1184\sqrt{A}-0.1184A^2+0.1184\,t\sin\ln(A+1)$ \\
\midrule[0.3pt]
38 & II.6.15b & $E_f,\allowbreak\varepsilon,\allowbreak p_d,\allowbreak\theta$ & $\frac{6^{1/3}}{2\pi^{1/3}}\left(\frac{p_d\sin\theta\cos\theta}{E_f\varepsilon}\right)^{1/3}$ \\
\midrule[0.3pt]
39 & Bio Population Growth 5 & $t,P$ & $0.9198P\left(1-\frac{P}{84.0283}\right)+\frac{0.9198P^2}{7.5314P+1}$ \\
\midrule[0.3pt]
40 & Materials Science 21 & $\varepsilon,T$ & $28.5804\varepsilon^2-0.2857(T-281.8702)+4.2784\varepsilon^3(T-281.8702)$ \\
\midrule[0.3pt]
41 & Bio Population Growth 3 & $t,P$ & $0.8450P\left(\frac{P}{5.1153}-1\right)\left(1-\frac{P}{34.4388}\right)+0.8450P\left(1-e^{-0.0968P}\right)$ \\
\midrule[0.3pt]
42 & Bio Population Growth 9 & $t,P$ & $0.1699P\left(\frac{P}{1.0472}-1\right)\left(1-\frac{P}{10.2105}\right)+0.1699P\left(1-\frac{P}{10.2105}\right)+0.1699P\left(1-e^{-0.0970P}\right)$ \\
\addlinespace[4pt]
\rowcolor{black!8}\multicolumn{4}{@{}l}{\textbf{Nguyen} \quad 3 tasks}\\
\addlinespace[2pt]
43 & Nguyen 12 & $x_1,x_2$ & $x_1^4-x_1^3+\frac{x_2^2}{2}-x_2$ \\
\midrule[0.3pt]
44 & Nguyen 6 & $x_1$ & $\sin(x_1)+\sin(x_1+x_1^2)$ \\
\midrule[0.3pt]
45 & Nguyen 9 & $x_1,x_2$ & $\sin(x_1)+\sin(x_2^2)$ \\
\addlinespace[4pt]
\rowcolor{black!8}\multicolumn{4}{@{}l}{\textbf{Korns} \quad 2 tasks}\\
\addlinespace[2pt]
46 & Korns 4 & $x_1,\dots,x_5$ & $0.13\sin(x_3)-2.3$ \\
\midrule[0.3pt]
47 & Korns 2 & $x_1,\dots,x_5$ & $0.23+14.2\,\operatorname{pdiv}(x_4+x_2,3x_5)$ \\
\addlinespace[4pt]
\rowcolor{black!8}\multicolumn{4}{@{}l}{\textbf{Keijzer} \quad 2 tasks}\\
\addlinespace[2pt]
48 & Keijzer 11 & $x_1,x_2$ & $x_1x_2+\sin\!\big((x_1-1)(x_2-1)\big)$ \\
\midrule[0.3pt]
49 & Keijzer 2 & $x_1$ & $0.3\,x_1\sin(2\pi x_1)$ \\
\addlinespace[4pt]
\rowcolor{black!8}\multicolumn{4}{@{}l}{\textbf{Vladislavleva} \quad 1 task}\\
\addlinespace[2pt]
50 & Vladislavleva 4 & $x_1,\dots,x_5$ & $\frac{10}{5+\sum_{i=1}^{5}(x_i-3)^2}$ \\
\end{longtable}
\parbox{\linewidth}{\footnotesize
For Korns 2, protected division is defined as
$\operatorname{pdiv}(a,b)=a/b$ for $|b|>10^{-3}$ and $1$ otherwise.
}
\endgroup

%% file: sections/appD_execution_protocol.tex
\section{Unified Execution Protocol}
\label{app:execution-protocol}

The protocol provides a shared contract for heterogeneous SR algorithms.
Each run specifies an algorithm, task, seed, and time budget.
Algorithms receive training data and permitted task information.
Fixed ID and OOD test splits and ground truth expressions remain reserved for evaluation.
Algorithm specific configurations are reported separately.

\subsection{Wrapper and Execution Architecture}
\label{app:wrapper-architecture}

\begin{figure}[htbp]
	\centering
	\includegraphics[width=0.95\linewidth]{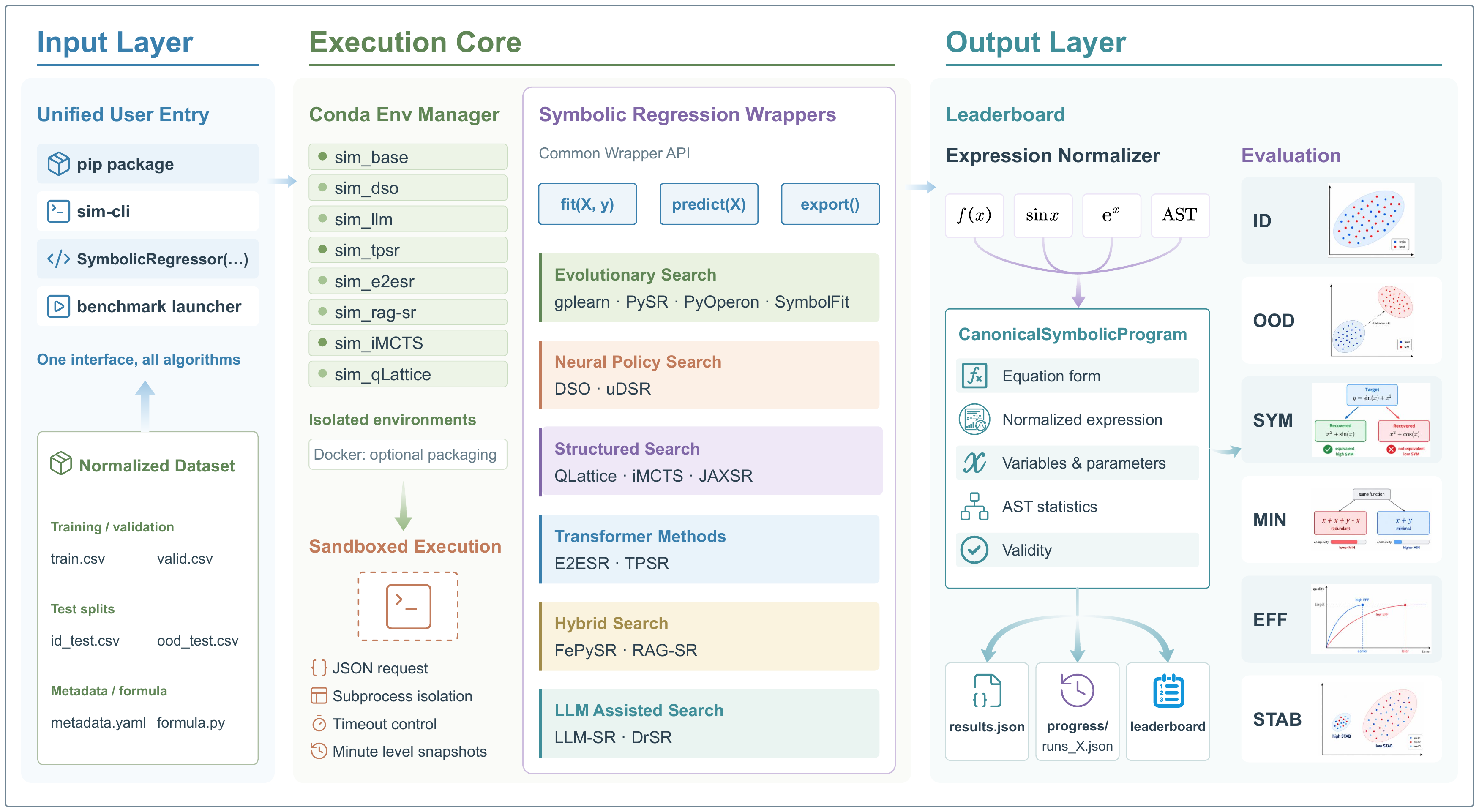}
	\caption{\textbf{\arena{} execution architecture.}
	A shared task contract connects isolated algorithm wrappers to standardized
	evaluation artifacts.}
	\label{fig:sa-infrastructure}
\end{figure}

\Cref{fig:sa-infrastructure} presents the execution architecture of \arena{}.
All entry points share one dataset contract and evaluation interface.
Algorithm dependencies remain isolated.
Wrappers expose common fitting, prediction, and expression retrieval operations.
Native search procedures and model selection rules remain unchanged.

The execution core maps each algorithm to an isolated environment.
Conda provides isolated environments, and Docker supports optional packaging.
A structured request specifies task inputs and run configuration.
Subprocess supervision isolates execution, enforces the run budget, and records termination status.

\Cref{alg:sr_adapter} defines the common adapter interface.
Each wrapper maps native operations to shared fitting, prediction, and expression
retrieval functions.

\begin{algorithm}[htbp]
\small
\caption{Unified SR adapter}
\label{alg:sr_adapter}
\begin{algorithmic}[1]
\Require Native method $A$, identifier $a$, environment $E_a$
\State $W_a \gets \Call{CreateAdapter}{A}$
\State $\Call{Bind}{W_a.\texttt{fit}, A.\texttt{fit}}$
\State $\Call{Bind}{W_a.\texttt{predict}, A.\texttt{predict}}$
\State $\Call{Bind}{W_a.\texttt{get\_optimal\_equation}, A}$
\State $\Call{Bind}{W_a.\texttt{get\_total\_equations}, A}$
\State $\Call{Register}{a, E_a, W_a}$
\end{algorithmic}
\end{algorithm}

Native outputs are converted to a shared symbolic representation before metric
computation.
The representation binds expression content to variables and fitted parameters.
Structural statistics and validity information support symbolic evaluation and
output checks.
Raw outputs are retained for auditing.
Symbolic simplification and adjudication follow
\Cref{app:six-axis-protocol}.

\subsection{Unified Run Procedure}
\label{app:run-procedure}

\Cref{alg:sr-framework} specifies one run defined by an algorithm, task, seed,
and evaluation condition.

\begin{algorithm}[htbp]
\small
\caption{Unified symbolic regression run procedure}
\label{alg:sr-framework}
\begin{algorithmic}[1]
\Require Algorithm $a$, dataset $d$, seed $s$, condition $c$,
         budget $B$, checkpoint interval $\Delta$
\State $D \gets \Call{LoadAndValidate}{d}$
\State $\theta \gets \Call{BuildConfig}{a,s,B,\mathrm{Schema}(D)}$
\State $\widetilde D_{\rm tr}
       \gets \Call{PerturbTrainingLabels}{D_{\rm tr},c,s}$
\State $w \gets \Call{StartIsolatedFit}{a,\widetilde D_{\rm tr},\theta}$
\State $\mathcal T \gets \Call{EmptyTrajectory}{}$; $t \gets 1$
\While{$\Call{Running}{w} \land t\Delta \leq B$}
    \State \Call{WaitUntilCheckpoint}{$t\Delta$}
    \State $(f_t,u_t,e_t) \gets \Call{ReadNativeIncumbent}{w}$
    \If{$e_t$ provides an auditable candidate}
        \State $p_t \gets \Call{ExportProgram}{f_t}$
        \State $m_t \gets \Call{EvaluateNumeric}{p_t,D_{\rm ID},D_{\rm OOD}}$
        \State $\mathcal T[t] \gets (p_t,u_t,m_t,e_t)$
    \Else
        \State $\mathcal T[t] \gets \Call{Unavailable}{e_t}$
    \EndIf
    \State $t \gets t+1$
\EndWhile
\State $\mathcal T \gets \Call{CompleteTrajectory}{\mathcal T,B,\Delta}$
\State $(\hat f,\sigma) \gets \Call{FinalizeOrRecover}{w,\mathcal T}$
\If{$\Call{Valid}{\hat f}$}
    \State $p \gets \Call{ExportProgram}{\hat f}$
    \State $M \gets \Call{EvaluateNumeric}{p,D_{\rm ID},D_{\rm OOD}}$
\Else
    \State $p \gets \mathrm{Unavailable}$; $M \gets \mathrm{NA}$
\EndIf
\State \Call{Record}{$a,d,s,c,p,\mathcal T,M,\sigma$}
\end{algorithmic}
\end{algorithm}

Let $D=(D_{\rm tr},D_{\rm val},D_{\rm ID},D_{\rm OOD})$, and let $\theta$ denote the algorithm specific configuration.
At each checkpoint, the wrapper records the native incumbent and its provenance.
An auditable incumbent is exported to the shared representation and evaluated on the fixed ID and OOD splits.

ID and OOD scores remain external to search and candidate selection.
Training label noise modifies $D_{\rm tr}$ only.
Failed or unavailable outputs retain explicit status labels.

\subsection{Recorded Results and Search Trajectories}
\label{app:result-artifacts}

Each run produces a final artifact and search trajectory.
The final artifact stores the selected expression, evaluation results, and run status.
Trajectory records store checkpoint expressions and provenance.

Formal evaluation uses 180 minutes with one checkpoint per minute.
Each checkpoint stores the native best so far expression supported by auditable evidence.
The latest auditable expression is carried forward between updates and after early termination.
A checkpoint lacking sufficient evidence remains marked as \texttt{unavailable}.

Recorded artifacts support Multi Axis Evaluation.
Final expressions support numerical and symbolic evaluation.
Search trajectories provide temporal evidence for EFF, and repeated seed runs provide the records required by STAB.

\subsection{Execution Isolation and Reproducibility Boundary}
\label{app:environment-isolation}

Separate environments and subprocess isolation reduce dependency conflicts
across algorithms.
Frozen task inputs, execution rules, environment specifications, and recorded
artifacts define the reproducibility boundary.
Additional boundaries for language model assisted search and symbolic
postprocessing are described in \Cref{app:llm-usage}.

%% file: sections/appE_six_axis_protocol.tex
\section{Multi Axis Evaluation Details}
\label{app:six-axis-protocol}

The six axes are fixed independently of the evaluated algorithm set.
ID and OOD measure numerical quality, SYM and MIN measure symbolic quality, and EFF and STAB characterize search behavior.

Formal evaluation uses 50 tasks and three random seeds
\[
\seedset=\{520,521,522\}.
\]
All axes except STAB are computed per task and seed and averaged across 150 runs.
STAB is computed per task across the three seeds and averaged across 50 tasks.
Unless stated otherwise, $\mathbb{E}[\cdot]$ denotes the empirical average associated with a metric.
Metrics use the $[0,1]$ scale and the 0 to 100 display scale.

\subsection{Numerical Quality Mapping}
\label{app:metric-common}

ID and OOD use the same bounded transformation of NMSE.
EFF and the numerical component of STAB reuse the mapping.
For a valid NMSE value $x$, the clipped log error is
\[
r(x)=\operatorname{clip}\!\left(
\log_{10}(\max(x,\epsilon)),
\ell_{\min},
\ell_{\max}
\right).
\]

Numerical quality is
\[
\phi(x)
=
\frac{\ell_{\max}-r(x)}
{\ell_{\max}-\ell_{\min}}.
\]

The bounds are $\ell_{\min}=-12$ and $\ell_{\max}=2$, with
$\epsilon=10^{-12}$.
The mapping satisfies $\phi(x)\in[0,1]$.
Lower error yields higher quality.
Invalid or unevaluable outputs receive quality zero.

For a final expression
\[
q^{\mathrm{ID}}
=
\phi(\mathrm{NMSE}^{\mathrm{ID}}),
\qquad
q^{\mathrm{OOD}}
=
\phi(\mathrm{NMSE}^{\mathrm{OOD}}).
\]

\subsection{Numerical Quality}

ID and OOD measure final numerical quality on the fixed evaluation splits
\[
\boxed{
\mathrm{Score}^{\mathrm{ID}}
=
\mathbb{E}[q^{\mathrm{ID}}],
\qquad
\mathrm{Score}^{\mathrm{OOD}}
=
\mathbb{E}[q^{\mathrm{OOD}}]
}.
\]

ID to OOD degradation is reported separately for diagnostic analysis
\[
\log_{10}
\frac{
\mathrm{NMSE}^{\mathrm{OOD}}+\epsilon
}{
\mathrm{NMSE}^{\mathrm{ID}}+\epsilon
}.
\]
The OOD score excludes the degradation measure.

\subsection{Opus Assisted Symbolic Postprocessing}
\label{app:opus-postprocessing}

Opus 5 assists symbolic postprocessing following search completion.
Each request contains the selected final expression, fixed variable mappings, and coefficients.
Domain and operator assumptions are also supplied.
Ground truth expressions are processed separately and remain fixed across algorithms.
Model assisted processing remains external to search and candidate selection.

\paragraph{Simplification and validation.}
The model returns a structured JSON record with outcome
\texttt{simplified}, \texttt{unchanged}, or \texttt{unable}.
An optional expression and brief justification are also returned.
The prompt requires preservation of mathematical meaning and excludes coefficient refitting, variable renaming, and additional domain assumptions.

Responses must satisfy a fixed schema.
Candidate expressions pass syntax and variable checks.
Symbolic comparisons and numerical probes provide diagnostic evidence.
An \texttt{unable} outcome retains the original expression and an explicit fallback flag.

Ordinary expressions use normalized expression trees implemented with SymPy \citep{meurer2017sympy}.
gplearn retains protected operators as typed prefix nodes and uses its native evaluator near guard boundaries.
Protected operators remain distinct from ordinary division, logarithms, and square roots.

\paragraph{Adjudication.}
Separate requests compare each processed prediction with its reference for SYM.
Seed pairs from the same algorithm, task, and condition are compared for the structural component of STAB.
Ground truth is excluded from the seed pair comparison.

Structural judgments distinguish mathematical equivalence, agreement after abstraction of numerical constants, and different or undetermined structure.
MIN and partial SYM use deterministic expression tree statistics.
A positive model judgment provides evaluation evidence and does not constitute formal mathematical proof.

\paragraph{Freezing and reuse.}
Processed expressions and judgments are frozen before metric aggregation.
Artifacts retain original and processed expressions, input and response hashes, and model identifiers.
Prompt and schema versions are also retained.
Cached judgments are reused only after matching frozen expression dependencies and run provenance.
Audit corrections remain linked to original responses.
Metric aggregation requires no additional model calls.

\subsection{Symbolic Quality}
\label{app:Symbolic Quality}

\paragraph{SYM.}

SYM measures agreement between the recovered expression and the underlying symbolic relation.
Ground truth and predicted expressions use the same simplification procedure
\[
f^{\mathrm{ref}}
=
\mathcal{S}(f^{\mathrm{gt}}),
\qquad
\widetilde f
=
\mathcal{S}(\widehat f).
\]

Let $d_{\mathrm{eq}}$ denote the evaluator's equivalence judgment.
The equivalence indicator is
\[
Eq
=
\mathbf{1}
\left[
d_{\mathrm{eq}}
=
\texttt{equivalent}
\right].
\]

For predictions lacking a positive equivalence judgment, partial recovery uses expression structure and symbolic content.
Tree similarity is
\[
S_{\mathrm{tree}}
=
1-
\mathrm{NED}
\left(
\widetilde f,
f^{\mathrm{ref}}
\right),
\]
where $\mathrm{NED}$ denotes normalized expression tree distance.

Variable and operator recovery are
\[
F_v
=
\mathrm{F1}_{\mathrm{variable}},
\qquad
F_o
=
\mathrm{F1}_{\mathrm{operator}},
\]
where both scores compare the corresponding symbolic sets of
$\widetilde f$ and $f^{\mathrm{ref}}$.

For an invalid or unparsable expression
\[
Eq
=
S_{\mathrm{tree}}
=
F_v
=
F_o
=
0.
\]

The symbolic score for one task and seed is
\[
\boxed{
m^{\mathrm{SYM}}
=
\begin{cases}
1,
& Eq=1,
\\[3pt]
0.5
\left(
S_{\mathrm{tree}}F_vF_o
\right)^{1/3},
& Eq=0.
\end{cases}
}
\]

The final SYM score is
\[
\boxed{
\mathrm{Score}^{\mathrm{SYM}}
=
\mathbb{E}[m^{\mathrm{SYM}}]
}.
\]

\paragraph{MIN.}

MIN measures expression complexity relative to the simplified reference used by SYM.
Complexity is the number of nodes in the expression tree
\[
C^{\mathrm{ref}}
=
C(f^{\mathrm{ref}}),
\qquad
C^{\mathrm{pred}}
=
C(\widetilde f).
\]

The minimality score for one task and seed is
\[
\boxed{
m^{\mathrm{MIN}}
=
\min
\left(
1,
\frac{
C^{\mathrm{ref}}
}{
C^{\mathrm{pred}}
}
\right)
}.
\]

An invalid or unparsable final expression receives
\[
m^{\mathrm{MIN}}=0.
\]

The final MIN score is
\[
\boxed{
\mathrm{Score}^{\mathrm{MIN}}
=
\mathbb{E}[m^{\mathrm{MIN}}]
}.
\]

A larger MIN score indicates lower expression complexity relative to the reference.

\subsection{Search Behavior}

\paragraph{EFF.}
\label{app:metric-eff}

EFF measures search progress relative to the best numerical quality reached in each run.
One numerical evaluation is recorded per minute.
Symbolic judgments remain external to EFF.

For
\[
t\in\{1,\ldots,T\},
\qquad
T=180,
\]
the evaluator measures the ID and OOD quality of the expression selected by the algorithm's native objective or model selection rule.
Test results remain external to candidate selection.
Numerical quality at minute $t$ is
\[
q(t)
=
\frac{
q^{\mathrm{ID}}(t)
+
q^{\mathrm{OOD}}(t)
}{2},
\]
where
\[
q^{\mathrm{ID}}(t)
=
\phi
\left(
\mathrm{NMSE}^{\mathrm{ID}}(t)
\right),
\qquad
q^{\mathrm{OOD}}(t)
=
\phi
\left(
\mathrm{NMSE}^{\mathrm{OOD}}(t)
\right).
\]

The maximum observed quality is
\[
q^{*}
=
\max_{1\leq t\leq T}
q(t).
\]

For a complete auditable trajectory, the run score is
\[
\boxed{
m^{\mathrm{EFF}}
=
\frac{1}{T}
\sum_{t=1}^{T}
\frac{q(t)}{q^{*}}
}.
\]

A run with $q^{*}=0$ receives
\[
m^{\mathrm{EFF}}=0.
\]

The final EFF score is
\[
\boxed{
\mathrm{Score}^{\mathrm{EFF}}
=
\mathbb{E}[m^{\mathrm{EFF}}]
}.
\]

The trajectory $q(t)$ may decrease after expression updates.
The value $q^{*}$ is computed at run end and used only for normalization.
Unavailable checkpoints follow the trajectory completion rule in
\Cref{app:result-artifacts}.

\paragraph{STAB.}
\label{app:metric-stab}

STAB measures consistency across repeated random seed runs on the same task.
For $K=|\seedset|$ runs, consistency is evaluated across
$\binom{K}{2}$ seed pairs.
The experiment uses $K=3$.
STAB combines numerical consistency, output validity, and structural consistency.

\textbf{Numerical consistency.}
For seeds $i$ and $j$
\[
\delta^{\mathrm{num}}_{ij}
=
\frac{
\left|
q_i^{\mathrm{ID}}
-
q_j^{\mathrm{ID}}
\right|
+
\left|
q_i^{\mathrm{OOD}}
-
q_j^{\mathrm{OOD}}
\right|
}{2}.
\]

Numerical consistency for one task is
\[
N
=
1-
\frac{1}{
\binom{|\seedset|}{2}
}
\sum_{i<j}
\delta^{\mathrm{num}}_{ij}.
\]

\textbf{Output validity.}
The valid run fraction for one task is
\[
V
=
\frac{
\#\{\text{valid seeds}\}
}{
|\seedset|
}.
\]

\textbf{Structural consistency.}
The evaluator judges every scheduled pair of frozen final expressions.
A valid pair receives a positive judgment for mathematical equivalence or matching structure after abstraction of numerical constants.
The indicator is
\[
I^{\mathrm{struct}}_{ij}
=
\begin{cases}
1,
& \text{valid pair with a positive symbolic judgment},
\\
0,
& \text{otherwise}.
\end{cases}
\]

Invalid pairs and judgments labeled
\texttt{undetermined} or \texttt{non\_applicable} receive zero.
Structural consistency is
\[
C
=
\frac{1}{
\binom{K}{2}
}
\sum_{i<j}
I^{\mathrm{struct}}_{ij}.
\]

The stability score for one task is
\[
\boxed{
m^{\mathrm{STAB}}
=
(NVC)^{1/3}
}.
\]

The final STAB score is
\[
\boxed{
\mathrm{Score}^{\mathrm{STAB}}
=
\mathbb{E}[m^{\mathrm{STAB}}]
}.
\]

The geometric mean requires all three components to remain high.

\textbf{Adjudication limitations.}
An audit found inconsistent pairwise symbolic judgments for 3 clean, 2 one percent noise, and 3 five percent noise algorithm task groups.
The cases limit interpretation of the structural component of STAB.

\subsection{Bootstrap Confidence Intervals}
\label{app:metric-bootstrap}

Confidence intervals use 1000 task bootstrap replicates.
Each replicate samples 50 tasks from \core{} with replacement and retains the three recorded runs for each sampled task.
The same task draws are used across algorithms, axes, and evaluation conditions.

For the five axes defined per run, each replicate averages the sampled run scores.
STAB averages the sampled task scores.
The 2.5th and 97.5th percentiles define the 95\% interval on the 0 to 100 display scale.

Evaluation outputs and symbolic judgments remain fixed during resampling.
The intervals quantify task resampling variability within the empirical \core{} sample and are not interpreted as significance tests.

\subsection{Supplementary Training Noise Robustness}
\label{app:metric-robu}

Training noise robustness is reported separately from the six formal evaluation axes.
Two supplementary conditions are evaluated
\[
\sigma\in\{0.01,0.05\}.
\]

Training labels are perturbed according to
\[
y'
=
y
+
\sigma
\operatorname{std}(y)
\epsilon,
\qquad
\epsilon
\sim
\mathcal{N}(0,1).
\]

The values of $\sigma$ set the Gaussian noise standard deviation to
$1\%$ and $5\%$ of the training target standard deviation.
ID and OOD test sets remain clean.

Clean and noisy numerical quality is reported on both evaluation splits.
Performance degradation is reported for each noise condition.

%% file: sections/appF_sensitivity.tex
\section{Sensitivity Analysis}
\label{app:sensitivity}

The analysis examines two construction decisions in \arena{}: Probe4
selection and \core{} construction.
Each decision is re-evaluated across controlled perturbations.
\subsection{Probe4 Selection Stability and Replacement Structure}
\label{app:probe4-stability}

All 12 calibration algorithms remain eligible for probe selection. The observed selection mass concentrates on a small competitive set, motivating an audit of panel stability and replacement structure. We examine the default panel
\[
\mathcal{A}_4=\{\mathrm{DSO},\mathrm{PyOperon},\mathrm{iMCTS},\mathrm{uDSR}\}
\]
across variation in the calibration sample and selection rule, followed by direct member removal tests. All analyses reuse the 2,400 Calib200 records from Appendix~C.2 and require no additional SR runs.

\paragraph{Calibration resampling.}
We perform 1,000 family stratified task bootstrap replicates. Each replicate samples task blocks with replacement and preserves the original source family counts. The 12 algorithm records associated with each sampled task move as one block. Probe selection is repeated at the default setting $\alpha=0.5$, $\tau=100$, and $m=3$, with all tied optima retained. The default Probe4 is the unique optimum in 63.4\% of replicates and belongs to the optimal set in 73.4\%. Its tie averaged selection mass is 68.3\%, and its median competition rank is 1 with an interquartile range of 1--2.

Table~\ref{tab:probe4-competitive-set} shows a concentrated competitive set. uDSR and PyOperon receive nearly all bootstrap inclusion mass. DSO and iMCTS remain frequent members, and QLattice is the principal alternative. gplearn receives limited mass. The other candidates receive none. The pattern emerges from the complete 12 algorithm selection procedure and is not imposed by prior filtering.

\begin{table}[H]
\centering
\caption{
Calibration usability and bootstrap inclusion of the 12 eligible probe candidates.
Inclusion frequency averages tied optima uniformly.
}
\label{tab:probe4-competitive-set}

\small
\setlength{\tabcolsep}{5.2pt}

\begin{tabular}{l c c c c}
\toprule
Algorithm &
Probe4 &
$H$ &
Inclusion (\%) &
Any optimum (\%) \\
\midrule

uDSR      & Yes & 0.9875 & 100.0 & 100.0 \\
PyOperon  & Yes & 0.9600 & 98.8  & 99.3 \\
DSO       & Yes & 0.9700 & 88.1  & 91.5 \\
iMCTS     & Yes & 0.9675 & 81.4  & 85.4 \\
QLattice  &     & 0.9600 & 30.5  & 35.0 \\
gplearn   &     & 0.9300 & 1.2   & 1.9 \\
PySR      &     & 0.8750 & 0.0   & 0.0 \\
TPSR      &     & 0.8625 & 0.0   & 0.0 \\
DrSR      &     & 0.8225 & 0.0   & 0.0 \\
LLM-SR    &     & 0.8175 & 0.0   & 0.0 \\
E2ESR     &     & 0.7775 & 0.0   & 0.0 \\
RAG-SR    &     & 0.6500 & 0.0   & 0.0 \\

\bottomrule
\end{tabular}
\end{table}

\paragraph{Selection rule perturbation.}
We vary $\alpha$ from 0.30 to 0.70 in increments of 0.05 and evaluate
$\tau\in\{10,30,100,300,1000\}$ with $m=3$ fixed.
Across 45 settings, Probe4 is the unique optimum in 31 and belongs to
the optimal set in 33.
Its competition rank remains within the top three for every tested setting.
The result indicates stable competition across moderate changes in the
usability definition.

\paragraph{Member removal.}
Each Probe4 member is removed in turn, and the complete selection procedure is rerun on the remaining 11 algorithms. QLattice replaces DSO, iMCTS, or uDSR, and gplearn replaces PyOperon. Every fallback panel preserves three paradigms. Table~\ref{tab:probe4-replacement} reports the resulting replacement panels and usability losses. The largest usability loss is 0.007500, corresponding to 0.75 percentage points on the $[0,1]$ usability scale. The result shows that the selection procedure admits structured fallback choices with limited loss in calibration usability.

\begin{table}[t]
\centering
\caption{
Probe4 replacement analysis.
$U$ denotes panel usability, and $\Delta U$ measures the loss from the
default Probe4 score $U=0.971250$.
}
\label{tab:probe4-replacement}

\small
\setlength{\tabcolsep}{7pt}
\renewcommand{\arraystretch}{1.08}

\begin{tabular}{l l c c c}
\toprule
Removed &
Replacement &
Best $U$ &
$\Delta U$ &
Paradigms \\
\midrule

iMCTS
& QLattice
& 0.969375
& 0.001875
& 3 \\

DSO
& QLattice
& 0.968750
& 0.002500
& 3 \\

uDSR
& QLattice
& 0.964375
& 0.006875
& 3 \\

PyOperon
& gplearn
& 0.963750
& 0.007500
& 3 \\

\bottomrule
\end{tabular}
\end{table}

\paragraph{Paradigm coverage.}
The minimum paradigm requirement isolates the role of search diversity in probe selection. At $m=2$, the default panel shares the optimum with a second panel at the same best usability of 0.971250. Requiring three paradigms removes the tie and preserves the same best usability. At $m=4$, the default panel becomes infeasible, and the best feasible panel changes to
\[
\{\mathrm{PyOperon},\mathrm{iMCTS},\mathrm{TPSR},\mathrm{uDSR}\},
\]
with usability 0.944375. The three paradigm setting resolves the ambiguity at $m=2$ with no reduction in maximum usability.

Overall, all 12 candidates enter the selection procedure, yet the evidence consistently concentrates on a small competitive set. Probe4 remains dominant or near dominant across calibration resampling and rule perturbations. Member removal produces structured fallback panels with small usability loss and preserved paradigm coverage. The analysis supports the stability of the selection process and exposes its principal alternatives.

\subsection{\core{} Construction}
\label{app:core50-construction-sensitivity}

The \core{} selection objective is
\[
S^*
=
\arg\max_{S\subset\fullset,\;|S|=50}
\left[
w_C\,\mathrm{Coverage}(S)
+w_I\,\mathrm{MeanInfo}(S)
+w_B\,\mathrm{Balance}(S)
\right],
\]
where Coverage measures task coverage in the structural and probe response spaces, MeanInfo measures task-level discriminative power across algorithms and stability, and Balance controls distributional shifts in difficulty and failure modes. The default weights are $\boldsymbol{w}_0=(0.45,0.35,0.20)$.

\paragraph{Single-parameter weight sensitivity.}
Each weight $w_j$ is adjusted to a given value $v$, and the other two weights
are renormalized to sum to one.
The examined ranges are $w_C\in[0.40,0.50]$, $w_I\in[0.30,0.40]$, and
$w_B\in[0.15,0.25]$.

For each setting, 30 independent random restarts are performed, yielding 450
selection results in total.
The full-ranking rate is the frequency with which the complete ordering
$\mathrm{uDSR}>\mathrm{iMCTS}>\mathrm{DSO}>\mathrm{PyOperon}$ is preserved.

\begin{table}[htbp]
    \centering
    \small
    \setlength{\tabcolsep}{4pt}
    \renewcommand{\arraystretch}{1.15}
    \caption{\textbf{Single-parameter weight perturbation results.}
    Jaccard similarity and overlapping tasks are measured against the
    default-weight selection under 30 random restarts per setting.
    Within-group Jaccard similarity compares the 30 restarts of one setting.}
    \label{tab:core50-weight-single}
    \begin{tabular}{@{}l c c c c c@{}}
        \toprule
        Perturbed weight & Weight range
        & \shortstack{Median Jaccard\\vs. default}
        & \shortstack{Overlapping\\tasks}
        & \shortstack{Median within-\\group Jaccard}
        & \shortstack{Full-ranking\\rate} \\
        \midrule
        Coverage & $[0.40,0.50]$ & 0.724 & 42 & 0.667 & $100\%$ \\
        MeanInfo & $[0.30,0.40]$ & 0.710 & 41.5 & 0.667 & $100\%$ \\
        Balance  & $[0.15,0.25]$ & 0.710 & 41.5 & 0.667 & $100\%$ \\
        \bottomrule
    \end{tabular}
\end{table}

\Cref{tab:core50-weight-single} shows that weight perturbations mainly affect
tasks near the selection boundary.
The median Jaccard similarity between the reselected 50 task sets and the default
set is approximately $0.710$, corresponding to 41 to 42 overlapping tasks.
The median within-group Jaccard similarity across the 30 restarts under the same
weight setting is $0.667$, indicating that approximately 40 high-quality tasks
are selected consistently, and the remaining positions are occupied by
interchangeable marginal tasks.
These results demonstrate that the \core{} objective is robust to moderate
perturbations of the default weights.

\paragraph{Joint random weight analysis.}
To examine joint variations in multiple weights, the sampling region is
\[
\mathcal{W}_{\mathrm{box}}
=
\left\{
\boldsymbol{w}
\;\middle|\;
\begin{aligned}
w_C &\in [0.40,0.50],\\
w_I &\in [0.30,0.40],\\
w_B &= 1-w_C-w_I \in [0.15,0.25]
\end{aligned}
\right\}.
\]
5,000 weight vectors are sampled uniformly from $\mathcal{W}_{\mathrm{box}}$.
For each sampled vector, all candidate 50 task sets are rescored, and the
highest-scoring set is retained.
\Cref{tab:core50-optimal-sets} shows that four distinct optimal sets are
obtained across the 5,000 samples, with the default-weight optimal set
accounting for $40.24\%$ of the selections.

\begin{table}[htbp]
    \centering
    \small
    \setlength{\tabcolsep}{5pt}
    \renewcommand{\arraystretch}{1.15}
    \caption{\textbf{Distribution of optimal sets under joint weight sampling.}
    Selections report the number of the 5,000 sampled weight vectors that
    produce each optimal set. S0 denotes the default-weight selection.}
    \label{tab:core50-optimal-sets}
    \begin{tabular}{@{}l c c c@{}}
        \toprule
        Optimal set & Selections & Share & \shortstack{Jaccard\\vs. default} \\
        \midrule
        S0 (\core{}) & 2012 & $40.24\%$ & 1.000 \\
        S1 & 1803 & $36.06\%$ & 0.724 \\
        S2 & 1173 & $23.46\%$ & 0.786 \\
        S3 & 12 & $0.24\%$ & 0.754 \\
        \bottomrule
    \end{tabular}
\end{table}

\Cref{tab:core50-joint-weights} reports a mean Jaccard similarity to the default
set of $0.850$, corresponding to approximately 46 overlapping tasks on average.
All 5,000 selected sets preserve the ranking of the four probes on \reservoir{}.

Task inclusion frequencies further reveal a stable core: 37 tasks are selected
in all 5,000 runs, 38 appear in at least $90\%$ of the runs, and 42 appear in at
least $75\%$.

\begin{table}[htbp]
    \centering
    \small
    \setlength{\tabcolsep}{5pt}
    \renewcommand{\arraystretch}{1.15}
    \caption{\textbf{Joint random weight analysis results.}
    Statistics summarize the 5,000 sampled weight vectors.}
    \label{tab:core50-joint-weights}
    \begin{tabular}{@{}l c@{}}
        \toprule
        Statistic & Result \\
        \midrule
        Mean Jaccard vs. default & 0.850 \\
        Minimum overlapping tasks & 42 / 50 \\
        Full-ranking rate & $100\%$ \\
        Tasks selected in all runs & 37 \\
        Tasks in $\geq 90\%$ of runs & 38 \\
        Tasks in $\geq 75\%$ of runs & 42 \\
        \bottomrule
    \end{tabular}
\end{table}

Under joint variations of all three weights, the optimal sets continue to
overlap substantially with \core{}, and the resulting algorithm ranking remains
unchanged.

%% file: sections/appG_additional_results.tex
\section{Additional Experimental Results}
\label{app:additional-results}

The appendix provides supporting evidence for Multi Axis Evaluation in
\arena{}. The analyses examine redundancy among evaluation axes and quantify
uncertainty from task sampling. Additional results assess robustness to
training noise and examine a representative gap between numerical
approximation and symbolic recovery.

\subsection{Correlation Among Evaluation Axes}
\label{app:axis-correlation}

\begin{figure}[htbp]
    \centering
    \includegraphics[width=0.4\linewidth]{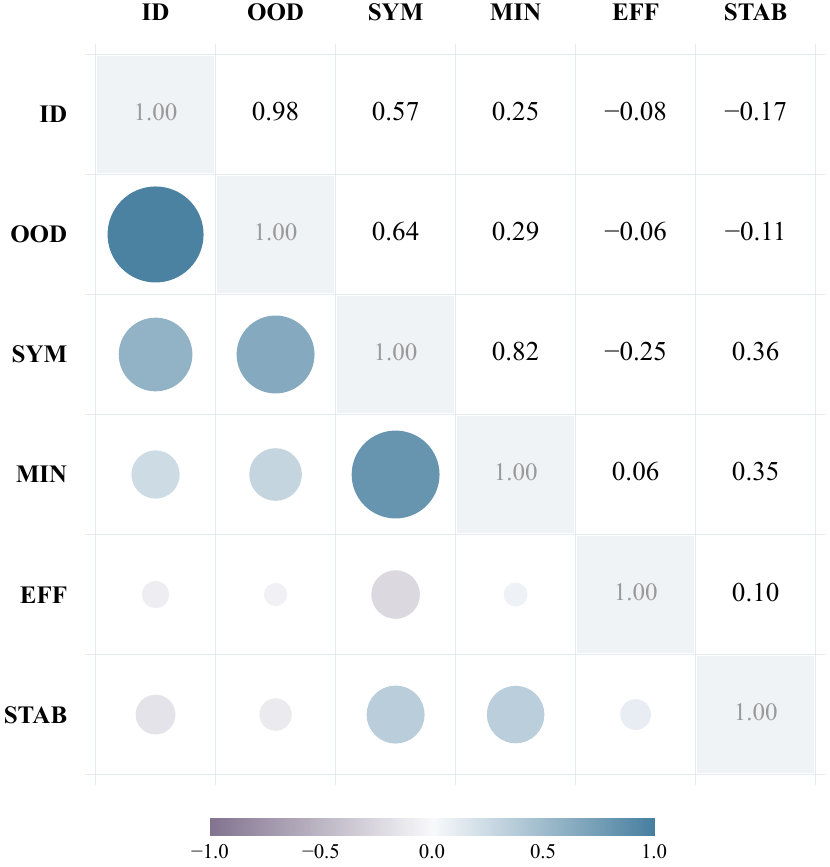}
    \caption{Spearman correlations among the six evaluation axes across
    15 algorithms on clean \core{}.}
    \label{fig:appendix-axis-correlation}
\end{figure}

\Cref{fig:appendix-axis-correlation} shows distinct association patterns
among the six axes. ID and OOD are strongly correlated
($\rho=0.98$), indicating similar numerical rankings across the two
evaluation distributions. SYM and MIN also show a strong association
($\rho=0.82$). Numerical quality has moderate correlation with SYM
($\rho=0.57$ to $0.64$) and weaker correlation with MIN
($\rho=0.25$ to $0.29$).

EFF shows little association with final numerical quality, with correlations
of $-0.08$ for ID and $-0.06$ for OOD. STAB also has weak association
with numerical quality. The observed structure provides empirical support
for retaining separate dimensions for numerical quality, symbolic quality,
and search behavior.

\subsection{Uncertainty in Six Axis Scores}
\label{app:algorithm-axis-profiles}
\label{app:aggregate-score-uncertainty}

\begin{figure}[htbp]
    \centering
    \includegraphics[width=0.9\linewidth]{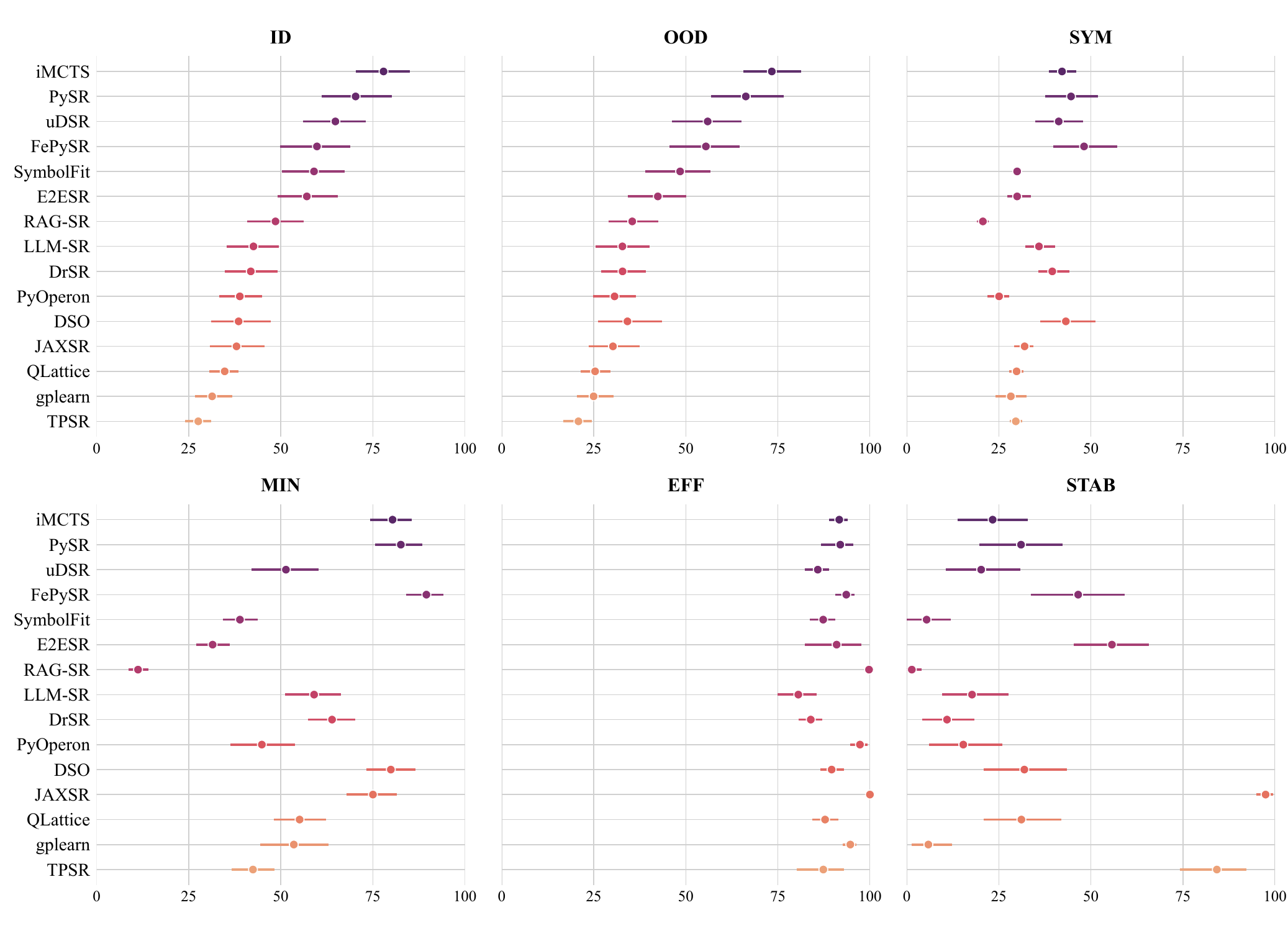}
    \caption{\textbf{Mean scores and 95\% task bootstrap intervals on clean
    \core{}.}
    Points denote mean scores and bars denote percentile intervals from
    1,000 resamples of 50 task blocks. Each sampled block retains the three
    recorded runs. Scores use the 0 to 100 display scale.}
    \label{fig:clean-six-axis-ci}
\end{figure}

\Cref{fig:clean-six-axis-ci} quantifies task sampling uncertainty for all six
axes. Interval widths differ across algorithms and metrics, indicating
different sensitivity to the sampled tasks. Small gaps between point estimates
should consequently be interpreted alongside their bootstrap intervals.
The analysis complements the six axis profiles reported in the main text.

\subsection{Robustness to Training Noise}
\label{app:noisy-training-results}

Training label perturbations extend the clean evaluation to corrupted
training signals.

\input{tables/noise_robustness_table}

\Cref{tab:noise_robustness} shows substantial variation in numerical
robustness across algorithms. High clean OOD performance does not consistently
correspond to small degradation under noise. iMCTS retains the highest OOD
score across all evaluated conditions, whereas DSO shows a smaller absolute
drop from its clean score. FePySR remains competitive at both noise levels.
The pattern distinguishes clean numerical quality from robustness to training
corruption.

\input{tables/six_axis_table_noise1}

\input{tables/six_axis_table_noise5}

Tables~\ref{tab:six_axis_profiles_noise1} and
\ref{tab:six_axis_profiles_noise5} report the complete six axis profiles at
$\sigma=0.01$ and $\sigma=0.05$. The tables extend the numerical robustness
analysis to symbolic quality and search behavior, using the same metric
definitions as the clean evaluation. Clean profiles are reported in
\Cref{tab:six_axis_profiles}.

\subsection{Numerical Approximation and Symbolic Recovery}
\label{app:representative-symbolic-mismatch}

A clean \texttt{Nguyen-9} run from PySR illustrates the distinction between
numerical approximation and symbolic recovery. For seed 520, the reference
expression and processed prediction are
\[
\begin{aligned}
f^{\mathrm{gt}}(x_1,x_2)
&=\sin(x_1)+\sin(x_2^2),\\
\widetilde f(x_1,x_2)
&=f^{\mathrm{gt}}(x_1,x_2)+1.1685188\times10^{-8}.
\end{aligned}
\]

The recorded NMSE values are $4.54\times10^{-15}$ on the fixed ID split
and $8.28\times10^{-17}$ on the fixed OOD split. Despite the near exact
numerical fit, the evaluator assigns \texttt{not\_equivalent}, producing
$m^{\mathrm{SYM}}=0.4886$.

The prediction preserves the principal functional structure and differs from
the reference by a small additive residual. Exact symbolic equivalence is
lost despite negligible numerical error. The example provides a concrete
motivation for evaluating numerical quality and symbolic recovery separately
in \arena{}.

%% file: tables/noise_robustness_table.tex
\begin{table*}[t]
\centering

\captionsetup{skip=0pt, aboveskip=0pt, belowskip=0.20em}
\caption{
Numerical robustness of 15 symbolic regression methods under training label noise.
ID and OOD scores are reported under clean, 1\%, and 5\% noise conditions.
$\Delta_{5\%}$ denotes the score change from clean to 5\% noise.
}
\label{tab:noise_robustness}

\begingroup
{\fontsize{7.25}{8.6}\selectfont
\renewcommand{\arraystretch}{1.13}
\setlength{\tabcolsep}{3.5pt}

\begin{tabular*}{0.985\textwidth}{
@{\extracolsep{\fill}}
>{\raggedright\arraybackslash}p{1.75cm}
>{\raggedright\arraybackslash}p{1.45cm}
c c c c c c c c
@{}
}
\arrayrulecolor{black}
\toprule

& & \multicolumn{4}{c}{\textbf{ID}} & \multicolumn{4}{c}{\textbf{OOD}} \\
\cmidrule(lr){3-6}\cmidrule(lr){7-10}

\textbf{Paradigm} & \textbf{Algorithm}
& \textbf{Clean} $\uparrow$ & \textbf{1\%} $\uparrow$ & \textbf{5\%} $\uparrow$ & $\boldsymbol{\Delta_{5\%}}$ $\uparrow$
& \textbf{Clean} $\uparrow$ & \textbf{1\%} $\uparrow$ & \textbf{5\%} $\uparrow$ & $\boldsymbol{\Delta_{5\%}}$ $\uparrow$ \\
\arrayrulecolor{black}\midrule

\multirow{3}{*}{\makecell[l]{Structured\\Search}}
& \textbf{iMCTS}
& \textbf{77.91}
& \textbf{55.08}
& \textbf{46.69}
& -31.22
& \textbf{73.35}
& \textbf{49.40}
& \textbf{40.73}
& $-32.62$ \\

& \textbf{QLattice} & 34.79 & 32.36 & 29.18 & $-5.61$ & 25.37 & 23.09 & 20.27 & $-5.10$ \\
& \textbf{JAXSR}    & 37.99 & 32.62 & 31.00 & $-6.99$ & 30.21 & 25.91 & 24.56 & $-5.65$ \\

\arrayrulecolor{black!28}\specialrule{0.25pt}{1.0pt}{1.0pt}\arrayrulecolor{black}

\multirow{4}{*}{\makecell[l]{Evolutionary\\Search}}
& \textbf{PySR}
& 70.34
& 48.42
& 41.05 & $-29.29$
& 66.28
& 41.50 & 35.53 & $-30.75$ \\

& \textbf{PyOperon}  & 38.89 & 36.29 & 32.97 & $-5.92$  & 30.64 & 27.85 & 24.42 & $-6.22$ \\
& \textbf{gplearn}   & 31.37 & 30.14 & 29.93 & $-1.44$  & 24.96 & 23.97 & 23.84 & $\mathbf{-1.12}$ \\
& \textbf{SymbolFit} & 59.04 & 44.38 & 38.42 & $-20.62$ & 48.46 & 33.87 & 28.26 & $-20.20$ \\

\arrayrulecolor{black!28}\specialrule{0.25pt}{1.0pt}{1.0pt}\arrayrulecolor{black}

\multirow{2}{*}{\makecell[l]{Neural Policy\\Search}}
& \textbf{uDSR} & 64.84 & 44.61 & 38.87 & $-25.97$ & 55.94 & 32.86 & 27.44 & $-28.50$ \\
& \textbf{DSO}  & 38.56 & 37.27 & 37.61 & $\mathbf{-0.95}$  & 34.14 & 32.95 & 32.84 & $-1.30$ \\

\arrayrulecolor{black!28}\specialrule{0.25pt}{1.0pt}{1.0pt}\arrayrulecolor{black}

\multirow{2}{*}{\makecell[l]{Hybrid\\Search}}
& \textbf{FePySR}
& 59.84 & 46.69
& 41.69
& $-18.15$
& 55.45
& 42.77
& 36.92
& $-18.53$ \\

& \textbf{RAG-SR} & 48.58 & 39.95 & 36.33 & $-12.25$ & 35.45 & 27.26 & 24.35 & $-11.10$ \\

\arrayrulecolor{black!28}\specialrule{0.25pt}{1.0pt}{1.0pt}\arrayrulecolor{black}

\multirow{2}{*}{\makecell[l]{Transformer\\Methods}}
& \textbf{E2ESR} & 57.09 & 43.64 & 36.31 & $-20.78$ & 42.40 & 31.36 & 25.97 & $-16.43$ \\
& \textbf{TPSR}  & 27.60 & 26.83 & 24.48 & $-3.12$  & 20.84 & 19.81 & 18.05 & $-2.79$ \\

\arrayrulecolor{black!28}\specialrule{0.25pt}{1.0pt}{1.0pt}\arrayrulecolor{black}

\multirow{2}{*}{\makecell[l]{LLM Assisted\\Search}}
& \textbf{DrSR}   & 41.86 & 37.67 & 31.76 & $-10.10$ & 32.82 & 28.55 & 23.44 & $-9.38$ \\
& \textbf{LLM-SR} & 42.59 & 35.12 & 33.36 & $-9.23$  & 32.78 & 26.08 & 23.00 & $-9.78$ \\

\arrayrulecolor{black}\bottomrule
\end{tabular*}
}

\vspace{0.18em}

{\fontsize{6.5}{7.6}\selectfont
\noindent\makebox[\linewidth][l]{%
Display scale: 0--100.\qquad
$\Delta_{5\%}=\mathrm{Score}_{5\%}-\mathrm{Score}_{clean}$.}
}

\endgroup
\end{table*}

%% file: tables/six_axis_table_noise1.tex
\begin{table*}[t]
\centering
\begingroup

\definecolor{SATableInk}{HTML}{000000}
\definecolor{SATableMuted}{HTML}{000000}
\definecolor{SATableRule}{HTML}{CED2D7}
\definecolor{SABestBackground}{HTML}{E6F2E8}
\definecolor{SABestInk}{HTML}{1B6B2B}
\definecolor{SASecondBackground}{HTML}{E2EDF8}
\definecolor{SASecondInk}{HTML}{244C78}

\setlength{\abovecaptionskip}{5pt}
\setlength{\belowcaptionskip}{7pt}
\caption{Six-axis profiles of 15 symbolic regression methods on \core{} at $1\%$ training-label noise ($\sigma=0.01$).
Methods are grouped by their primary search paradigm.
Higher values denote better performance on each axis.}
\label{tab:six_axis_profiles_noise1}

\fontsize{7.2}{9.2}\selectfont
\color{SATableInk}
\renewcommand{\arraystretch}{1.12}
\setlength{\tabcolsep}{0pt}
\setlength{\heavyrulewidth}{0.8pt}
\setlength{\lightrulewidth}{0.3pt}
\setlength{\cmidrulewidth}{0.3pt}
\setlength{\aboverulesep}{1.6pt}
\setlength{\belowrulesep}{1.6pt}

\edef\SAParadigmWidth{\the\dimexpr 0.13\linewidth\relax}
\edef\SAAlgorithmWidth{\the\dimexpr 0.14\linewidth\relax}
\edef\SAScoreWidth{\the\dimexpr(\linewidth-\SAParadigmWidth-\SAAlgorithmWidth)/6\relax}
\edef\SAHighlightWidth{\the\dimexpr\SAScoreWidth-11pt\relax}

\newcommand{\SAHighlight}[3]{%
  \begingroup
  \setlength{\fboxsep}{0.4pt}%
  \colorbox{#1}{\makebox[\SAHighlightWidth][c]{\textcolor{#2}{\bfseries #3}}}%
  \endgroup
}
\newcommand{\SABest}[1]{\SAHighlight{SABestBackground}{SABestInk}{#1}}
\newcommand{\SASecond}[1]{\SAHighlight{SASecondBackground}{SASecondInk}{#1}}
\newcommand{\SAGroup}[2]{%
  \multirow{#1}{\SAParadigmWidth}{%
    \color{SATableMuted}\fontsize{6.7}{8.6}\selectfont\raggedright #2}%
}
\newcommand{\SASeparator}{%
  \arrayrulecolor{SATableRule}%
  \specialrule{\lightrulewidth}{\aboverulesep}{\belowrulesep}%
  \arrayrulecolor{SATableInk}%
}

\renewcommand{\scorecell}[3]{#3}
\renewcommand{\bestscore}[3]{\SABest{#3}}
\renewcommand{\secondscore}[3]{\SASecond{#3}}

\begin{tabular}{@{}
  >{\raggedright\arraybackslash}m{\SAParadigmWidth}
  >{\raggedright\arraybackslash}m{\SAAlgorithmWidth}
  *{6}{>{\centering\arraybackslash}m{\SAScoreWidth}}
@{}}
\arrayrulecolor{SATableInk}
\toprule
& & \multicolumn{2}{c}{\textbf{Numerical}}
    & \multicolumn{2}{c}{\textbf{Symbolic}}
    & \multicolumn{2}{c}{\textbf{Search}} \\
\arrayrulecolor{SATableInk}
\cmidrule(lr){3-4}
\cmidrule(lr){5-6}
\cmidrule(lr){7-8}
\textbf{Paradigm} & \textbf{Algorithm}
  & \textbf{ID} $\uparrow$ & \textbf{OOD} $\uparrow$ & \textbf{SYM} $\uparrow$
  & \textbf{MIN} $\uparrow$ & \textbf{EFF} $\uparrow$ & \textbf{STAB} $\uparrow$ \\
\arrayrulecolor{SATableInk}%
\specialrule{\lightrulewidth}{\aboverulesep}{\belowrulesep}

\SAGroup{3}{Structured\\Search}
&
\textbf{iMCTS}
&
\bestscore{NumBlue}{41}{55.08}
&
\bestscore{NumBlue}{41}{49.40}
&
\secondscore{SymPurple}{33}{37.22}
&
\scorecell{SymPurple}{36}{72.88}
&
\scorecell{SearchGreen}{33}{89.34}
&
\scorecell{SearchGreen}{14}{12.82}
\\

&
\textbf{QLattice}
&
\scorecell{NumBlue}{21}{32.36}
&
\scorecell{NumBlue}{19}{23.09}
&
\scorecell{SymPurple}{15}{29.59}
&
\scorecell{SymPurple}{28}{54.60}
&
\scorecell{SearchGreen}{24}{90.87}
&
\scorecell{SearchGreen}{20}{38.32}
\\

&
\textbf{JAXSR}
&
\scorecell{NumBlue}{27}{32.62}
&
\scorecell{NumBlue}{24}{25.91}
&
\scorecell{SymPurple}{24}{30.88}
&
\scorecell{SymPurple}{38}{76.16}
&
\bestscore{SearchGreen}{41}{99.95}
&
\bestscore{SearchGreen}{41}{91.33}
\\

\SASeparator

\SAGroup{4}{Evolutionary\\Search}
&
\textbf{PySR}
&
\secondscore{NumBlue}{33}{48.42}
&
\scorecell{NumBlue}{33}{41.50}
&
\scorecell{SymPurple}{21}{30.53}
&
\scorecell{SymPurple}{30}{60.26}
&
\scorecell{SearchGreen}{23}{86.08}
&
\scorecell{SearchGreen}{10}{6.02}
\\

&
\textbf{PyOperon}
&
\scorecell{NumBlue}{22}{36.29}
&
\scorecell{NumBlue}{19}{27.85}
&
\scorecell{SymPurple}{9}{25.14}
&
\scorecell{SymPurple}{22}{43.76}
&
\scorecell{SearchGreen}{31}{97.30}
&
\scorecell{SearchGreen}{13}{15.92}
\\

&
\textbf{gplearn}
&
\scorecell{NumBlue}{25}{30.14}
&
\scorecell{NumBlue}{23}{23.97}
&
\scorecell{SymPurple}{10}{28.03}
&
\scorecell{SymPurple}{29}{53.30}
&
\scorecell{SearchGreen}{38}{94.24}
&
\scorecell{SearchGreen}{13}{7.77}
\\

&
\textbf{SymbolFit}
&
\scorecell{NumBlue}{35}{44.38}
&
\scorecell{NumBlue}{31}{33.87}
&
\scorecell{SymPurple}{22}{29.48}
&
\scorecell{SymPurple}{19}{36.93}
&
\scorecell{SearchGreen}{33}{84.49}
&
\scorecell{SearchGreen}{9}{2.74}
\\

\SASeparator

\SAGroup{2}{Neural Policy\\Search}
&
\textbf{uDSR}
&
\scorecell{NumBlue}{35}{44.61}
&
\scorecell{NumBlue}{28}{32.86}
&
\scorecell{SymPurple}{20}{27.79}
&
\scorecell{SymPurple}{17}{32.64}
&
\scorecell{SearchGreen}{33}{83.80}
&
\scorecell{SearchGreen}{10}{4.14}
\\

&
\textbf{DSO}
&
\scorecell{NumBlue}{30}{37.27}
&
\scorecell{NumBlue}{30}{32.95}
&
\bestscore{SymPurple}{41}{43.32}
&
\secondscore{SymPurple}{39}{81.81}
&
\scorecell{SearchGreen}{35}{90.21}
&
\scorecell{SearchGreen}{18}{32.25}
\\

\SASeparator

\SAGroup{2}{Hybrid\\Search}
&
\textbf{FePySR}
&
\scorecell{NumBlue}{30}{46.69}
&
\secondscore{NumBlue}{28}{42.77}
&
\scorecell{SymPurple}{22}{35.82}
&
\bestscore{SymPurple}{41}{83.67}
&
\scorecell{SearchGreen}{34}{89.49}
&
\secondscore{SearchGreen}{15}{43.75}
\\

&
\textbf{RAG-SR}
&
\scorecell{NumBlue}{8}{39.95}
&
\scorecell{NumBlue}{8}{27.26}
&
\scorecell{SymPurple}{8}{19.35}
&
\scorecell{SymPurple}{8}{9.64}
&
\secondscore{SearchGreen}{8}{99.35}
&
\scorecell{SearchGreen}{8}{0.00}
\\

\SASeparator

\SAGroup{2}{Transformer\\Methods}
&
\textbf{E2ESR}
&
\scorecell{NumBlue}{10}{43.64}
&
\scorecell{NumBlue}{10}{31.36}
&
\scorecell{SymPurple}{25}{27.65}
&
\scorecell{SymPurple}{28}{28.21}
&
\scorecell{SearchGreen}{37}{89.44}
&
\scorecell{SearchGreen}{23}{4.09}
\\

&
\textbf{TPSR}
&
\scorecell{NumBlue}{10}{26.83}
&
\scorecell{NumBlue}{10}{19.81}
&
\scorecell{SymPurple}{22}{29.36}
&
\scorecell{SymPurple}{23}{41.17}
&
\scorecell{SearchGreen}{13}{87.37}
&
\scorecell{SearchGreen}{13}{8.13}
\\

\SASeparator

\SAGroup{2}{LLM Assisted\\Search}
&
\textbf{DrSR}
&
\scorecell{NumBlue}{29}{37.67}
&
\scorecell{NumBlue}{24}{28.55}
&
\scorecell{SymPurple}{24}{34.65}
&
\scorecell{SymPurple}{24}{64.07}
&
\scorecell{SearchGreen}{29}{84.74}
&
\scorecell{SearchGreen}{10}{11.73}
\\

&
\textbf{LLM-SR}
&
\scorecell{NumBlue}{31}{35.12}
&
\scorecell{NumBlue}{26}{26.08}
&
\scorecell{SymPurple}{18}{30.85}
&
\scorecell{SymPurple}{20}{50.86}
&
\scorecell{SearchGreen}{26}{81.69}
&
\scorecell{SearchGreen}{9}{8.23}
\\

\bottomrule
\end{tabular}

\par\vspace{5pt}
{\fontsize{6.2}{7.6}\selectfont\color{SATableMuted}
\makebox[\linewidth][l]{%
  Display scale 0--100.\hfill
  {\setlength{\fboxsep}{0pt}\colorbox{SABestBackground}{\rule{0pt}{5pt}\hspace{8pt}}}%
  \hspace{3pt}\textcolor{SABestInk}{Best (1st)}\hspace{13pt}%
  {\setlength{\fboxsep}{0pt}\colorbox{SASecondBackground}{\rule{0pt}{5pt}\hspace{8pt}}}%
  \hspace{3pt}\textcolor{SASecondInk}{Second-best (2nd)}%
}\par
}

\endgroup
\end{table*}

%% file: tables/six_axis_table_noise5.tex
\begin{table*}[t]
\centering
\begingroup

\definecolor{SATableInk}{HTML}{000000}
\definecolor{SATableMuted}{HTML}{000000}
\definecolor{SATableRule}{HTML}{CED2D7}
\definecolor{SABestBackground}{HTML}{E6F2E8}
\definecolor{SABestInk}{HTML}{1B6B2B}
\definecolor{SASecondBackground}{HTML}{E2EDF8}
\definecolor{SASecondInk}{HTML}{244C78}

\setlength{\abovecaptionskip}{5pt}
\setlength{\belowcaptionskip}{7pt}
\caption{Six-axis profiles of 15 symbolic regression methods on \core{} at $5\%$ training-label noise ($\sigma=0.05$).
Methods are grouped by their primary search paradigm.
Higher values denote better performance on each axis.}
\label{tab:six_axis_profiles_noise5}

\fontsize{7.2}{9.2}\selectfont
\color{SATableInk}
\renewcommand{\arraystretch}{1.12}
\setlength{\tabcolsep}{0pt}
\setlength{\heavyrulewidth}{0.8pt}
\setlength{\lightrulewidth}{0.3pt}
\setlength{\cmidrulewidth}{0.3pt}
\setlength{\aboverulesep}{1.6pt}
\setlength{\belowrulesep}{1.6pt}

\edef\SAParadigmWidth{\the\dimexpr 0.13\linewidth\relax}
\edef\SAAlgorithmWidth{\the\dimexpr 0.14\linewidth\relax}
\edef\SAScoreWidth{\the\dimexpr(\linewidth-\SAParadigmWidth-\SAAlgorithmWidth)/6\relax}
\edef\SAHighlightWidth{\the\dimexpr\SAScoreWidth-11pt\relax}

\newcommand{\SAHighlight}[3]{%
  \begingroup
  \setlength{\fboxsep}{0.4pt}%
  \colorbox{#1}{\makebox[\SAHighlightWidth][c]{\textcolor{#2}{\bfseries #3}}}%
  \endgroup
}
\newcommand{\SABest}[1]{\SAHighlight{SABestBackground}{SABestInk}{#1}}
\newcommand{\SASecond}[1]{\SAHighlight{SASecondBackground}{SASecondInk}{#1}}
\newcommand{\SAGroup}[2]{%
  \multirow{#1}{\SAParadigmWidth}{%
    \color{SATableMuted}\fontsize{6.7}{8.6}\selectfont\raggedright #2}%
}
\newcommand{\SASeparator}{%
  \arrayrulecolor{SATableRule}%
  \specialrule{\lightrulewidth}{\aboverulesep}{\belowrulesep}%
  \arrayrulecolor{SATableInk}%
}

\renewcommand{\scorecell}[3]{#3}
\renewcommand{\bestscore}[3]{\SABest{#3}}
\renewcommand{\secondscore}[3]{\SASecond{#3}}

\begin{tabular}{@{}
  >{\raggedright\arraybackslash}m{\SAParadigmWidth}
  >{\raggedright\arraybackslash}m{\SAAlgorithmWidth}
  *{6}{>{\centering\arraybackslash}m{\SAScoreWidth}}
@{}}
\arrayrulecolor{SATableInk}
\toprule
& & \multicolumn{2}{c}{\textbf{Numerical}}
    & \multicolumn{2}{c}{\textbf{Symbolic}}
    & \multicolumn{2}{c}{\textbf{Search}} \\
\arrayrulecolor{SATableInk}
\cmidrule(lr){3-4}
\cmidrule(lr){5-6}
\cmidrule(lr){7-8}
\textbf{Paradigm} & \textbf{Algorithm}
  & \textbf{ID} $\uparrow$ & \textbf{OOD} $\uparrow$ & \textbf{SYM} $\uparrow$
  & \textbf{MIN} $\uparrow$ & \textbf{EFF} $\uparrow$ & \textbf{STAB} $\uparrow$ \\
\arrayrulecolor{SATableInk}%
\specialrule{\lightrulewidth}{\aboverulesep}{\belowrulesep}

\SAGroup{3}{Structured\\Search}
&
\textbf{iMCTS}
&
\bestscore{NumBlue}{41}{46.69}
&
\bestscore{NumBlue}{41}{40.73}
&
\scorecell{SymPurple}{32}{36.43}
&
\scorecell{SymPurple}{32}{66.26}
&
\scorecell{SearchGreen}{30}{86.11}
&
\scorecell{SearchGreen}{9}{5.24}
\\

&
\textbf{QLattice}
&
\scorecell{NumBlue}{23}{29.18}
&
\scorecell{NumBlue}{21}{20.27}
&
\scorecell{SymPurple}{16}{29.19}
&
\scorecell{SymPurple}{27}{54.99}
&
\scorecell{SearchGreen}{24}{85.95}
&
\scorecell{SearchGreen}{19}{29.66}
\\

&
\textbf{JAXSR}
&
\scorecell{NumBlue}{30}{31.00}
&
\scorecell{NumBlue}{28}{24.56}
&
\scorecell{SymPurple}{27}{31.29}
&
\scorecell{SymPurple}{39}{77.47}
&
\bestscore{SearchGreen}{41}{99.95}
&
\bestscore{SearchGreen}{41}{91.95}
\\

\SASeparator

\SAGroup{4}{Evolutionary\\Search}
&
\textbf{PySR}
&
\scorecell{NumBlue}{34}{41.05}
&
\scorecell{NumBlue}{35}{35.53}
&
\scorecell{SymPurple}{23}{30.66}
&
\scorecell{SymPurple}{29}{59.16}
&
\scorecell{SearchGreen}{25}{85.28}
&
\scorecell{SearchGreen}{11}{6.73}
\\

&
\textbf{PyOperon}
&
\scorecell{NumBlue}{26}{32.97}
&
\scorecell{NumBlue}{23}{24.42}
&
\scorecell{SymPurple}{8}{23.60}
&
\scorecell{SymPurple}{20}{39.83}
&
\scorecell{SearchGreen}{33}{97.27}
&
\scorecell{SearchGreen}{13}{15.33}
\\

&
\textbf{gplearn}
&
\scorecell{NumBlue}{29}{29.93}
&
\scorecell{NumBlue}{28}{23.84}
&
\scorecell{SymPurple}{9}{27.30}
&
\scorecell{SymPurple}{29}{53.52}
&
\scorecell{SearchGreen}{38}{93.24}
&
\scorecell{SearchGreen}{14}{3.90}
\\

&
\textbf{SymbolFit}
&
\scorecell{NumBlue}{36}{38.42}
&
\scorecell{NumBlue}{32}{28.26}
&
\scorecell{SymPurple}{22}{29.23}
&
\scorecell{SymPurple}{19}{36.38}
&
\scorecell{SearchGreen}{33}{83.24}
&
\scorecell{SearchGreen}{8}{2.77}
\\

\SASeparator

\SAGroup{2}{Neural Policy\\Search}
&
\textbf{uDSR}
&
\scorecell{NumBlue}{36}{38.87}
&
\scorecell{NumBlue}{29}{27.44}
&
\scorecell{SymPurple}{19}{27.88}
&
\scorecell{SymPurple}{17}{32.53}
&
\scorecell{SearchGreen}{33}{82.47}
&
\scorecell{SearchGreen}{11}{7.05}
\\

&
\textbf{DSO}
&
\scorecell{NumBlue}{34}{37.61}
&
\scorecell{NumBlue}{35}{32.84}
&
\bestscore{SymPurple}{41}{42.83}
&
\secondscore{SymPurple}{40}{80.35}
&
\scorecell{SearchGreen}{35}{90.29}
&
\scorecell{SearchGreen}{18}{29.73}
\\

\SASeparator

\SAGroup{2}{Hybrid\\Search}
&
\textbf{FePySR}
&
\secondscore{NumBlue}{32}{41.69}
&
\secondscore{NumBlue}{31}{36.92}
&
\secondscore{SymPurple}{26}{36.44}
&
\bestscore{SymPurple}{41}{84.89}
&
\scorecell{SearchGreen}{34}{88.83}
&
\secondscore{SearchGreen}{15}{40.14}
\\

&
\textbf{RAG-SR}
&
\scorecell{NumBlue}{8}{36.33}
&
\scorecell{NumBlue}{8}{24.35}
&
\scorecell{SymPurple}{8}{18.79}
&
\scorecell{SymPurple}{8}{8.66}
&
\secondscore{SearchGreen}{8}{99.47}
&
\scorecell{SearchGreen}{8}{0.00}
\\

\SASeparator

\SAGroup{2}{Transformer\\Methods}
&
\textbf{E2ESR}
&
\scorecell{NumBlue}{11}{36.31}
&
\scorecell{NumBlue}{11}{25.97}
&
\scorecell{SymPurple}{24}{27.30}
&
\scorecell{SymPurple}{27}{26.51}
&
\scorecell{SearchGreen}{37}{89.18}
&
\scorecell{SearchGreen}{17}{1.99}
\\

&
\textbf{TPSR}
&
\scorecell{NumBlue}{11}{24.48}
&
\scorecell{NumBlue}{11}{18.05}
&
\scorecell{SymPurple}{22}{28.62}
&
\scorecell{SymPurple}{22}{42.08}
&
\scorecell{SearchGreen}{18}{88.48}
&
\scorecell{SearchGreen}{9}{2.75}
\\

\SASeparator

\SAGroup{2}{LLM Assisted\\Search}
&
\textbf{DrSR}
&
\scorecell{NumBlue}{29}{31.76}
&
\scorecell{NumBlue}{24}{23.44}
&
\scorecell{SymPurple}{22}{33.36}
&
\scorecell{SymPurple}{22}{61.38}
&
\scorecell{SearchGreen}{29}{83.43}
&
\scorecell{SearchGreen}{9}{12.44}
\\

&
\textbf{LLM-SR}
&
\scorecell{NumBlue}{32}{33.36}
&
\scorecell{NumBlue}{27}{23.00}
&
\scorecell{SymPurple}{32}{31.26}
&
\scorecell{SymPurple}{27}{51.32}
&
\scorecell{SearchGreen}{27}{80.96}
&
\scorecell{SearchGreen}{10}{4.05}
\\

\bottomrule
\end{tabular}

\par\vspace{5pt}
{\fontsize{6.2}{7.6}\selectfont\color{SATableMuted}
\makebox[\linewidth][l]{%
  Display scale 0--100.\hfill
  {\setlength{\fboxsep}{0pt}\colorbox{SABestBackground}{\rule{0pt}{5pt}\hspace{8pt}}}%
  \hspace{3pt}\textcolor{SABestInk}{Best (1st)}\hspace{13pt}%
  {\setlength{\fboxsep}{0pt}\colorbox{SASecondBackground}{\rule{0pt}{5pt}\hspace{8pt}}}%
  \hspace{3pt}\textcolor{SASecondInk}{Second-best (2nd)}%
}\par
}

\endgroup
\end{table*}

%% file: sections/appH_evaluated_baselines.tex
\section{Evaluated Algorithms and Experimental Settings}
\label{app:evaluated-baselines}

The study evaluates 15 symbolic regression algorithms across six primary
search paradigms.
FePySR~\citep{yu2026fepysr}, SymbolFit~\citep{tsoi2025symbolfit}, and
JAXSR~\citep{kitchin2024jaxsr} are excluded from benchmark construction and
serve as validation algorithms.
All methods are evaluated on the frozen \core{} benchmark.

\paragraph{Algorithm panel.}
\Cref{tab:algorithm-panel} characterizes the algorithmic coverage of the
evaluation by grouping the 15 methods according to their primary search
paradigms and search mechanisms.

\begin{table}[htbp]
    \centering
    \caption{
    \textbf{Symbolic regression algorithms evaluated in \arena{}.}
    Bold marks validation algorithms.
    }
    \label{tab:algorithm-panel}

    \scriptsize
    \setlength{\tabcolsep}{4pt}
    \renewcommand{\arraystretch}{1.10}

    \begin{tabularx}{\linewidth}{
        @{}
        >{\raggedright\arraybackslash}p{0.24\linewidth}
        >{\raggedright\arraybackslash}p{0.33\linewidth}
        >{\raggedright\arraybackslash}X
        @{}
    }
        \toprule
        \textbf{Paradigm} &
        \textbf{Method} &
        \textbf{Description} \\
        \midrule

        \multirow[t]{3}{=}{Structured Search}
        & iMCTS~\citep{huang2025imcts}
        & Monte Carlo tree search for guided expression exploration. \\

        & QLattice~\citep{brolos2021feyn}
        & Probabilistic graph search for compact symbolic models. \\

        & \textbf{JAXSR}~\citep{kitchin2024jaxsr}
        & Structured symbolic model construction with optimization in JAX. \\

        \midrule[0.3pt]

        \multirow[t]{2}{=}{Neural Policy Search}
        & uDSR~\citep{landajuela2022unified}
        & Neural symbolic search with evolutionary refinement. \\

        & DSO~\citep{petersen2021dsr}
        & Policy search using risk seeking optimization. \\

        \midrule[0.3pt]

        \multirow[t]{4}{=}{Evolutionary Search}
        & PySR~\citep{cranmer2023pysr}
        & Population search with complexity aware model selection. \\

        & PyOperon~\citep{burlacu2020operon}
        & Evolutionary tree search with constant optimization. \\

        & gplearn~\citep{stephens2016gplearn}
        & Genetic programming with protected operators and parsimony control. \\

        & \textbf{SymbolFit}~\citep{tsoi2025symbolfit}
        & Hybrid symbolic search with numerical parameter optimization. \\

        \midrule[0.3pt]

        \multirow[t]{2}{=}{Transformer Methods}
        & E2ESR~\citep{kamienny2022end}
        & Pretrained neural generation of expressions from numerical data. \\

        & TPSR~\citep{shojaee2023tpsr}
        & Pretrained expression generation guided by planning. \\

        \midrule[0.3pt]

        \multirow[t]{2}{=}{Hybrid Search}
        & \textbf{FePySR}~\citep{yu2026fepysr}
        & PySR based symbolic regression with feature enhancement. \\

        & RAG-SR~\citep{zhang2025ragsr}
        & Retrieval guided symbolic search with evolutionary optimization. \\

        \midrule[0.3pt]

        \multirow[t]{2}{=}{LLM Assisted Search}
        & DrSR~\citep{wang2025drsr}
        & LLM symbolic regression with iterative candidate refinement. \\

        & LLM-SR~\citep{shojaee2025llmsr}
        & LLM equation proposals with numerical parameter refinement. \\

        \bottomrule
    \end{tabularx}
\end{table}

\paragraph{Evaluation configurations.}
\Cref{tab:core50-hyperparameters-compact} records the configurations used in the
formal \core{} evaluation.
The evaluated uDSR configuration uses the fixed \emph{uDSR trunk variant},
which combines a DSO style controller, the LINEAR poly token, and GP meld.

\begingroup
\scriptsize
\setlength{\tabcolsep}{2pt}
\renewcommand{\arraystretch}{1.8}

\begin{longtable}{
@{}
>{\raggedright\arraybackslash}p{0.13\linewidth}
>{\raggedright\arraybackslash}p{0.17\linewidth}
>{\raggedright\arraybackslash}p{0.20\linewidth}
>{\raggedright\arraybackslash}p{0.46\linewidth}
@{}
}
\caption{\textbf{Algorithm specific configurations for the formal \core{} evaluation.}
Gray rows mark validation algorithms.}
\label{tab:core50-hyperparameters-compact}\\

\toprule
\textbf{Algorithm} &
\textbf{Search setting} &
\textbf{Operators and basis} &
\textbf{Fixed settings} \\
\midrule
\endfirsthead

\caption[]{\textbf{Algorithm specific configurations for the formal \core{} evaluation (continued).}}\\
\toprule
\textbf{Algorithm} &
\textbf{Search setting} &
\textbf{Operators and basis} &
\textbf{Fixed settings} \\
\midrule
\endhead

\bottomrule
\endlastfoot

\textbf{gplearn}
&
Population 1000.
&
add, sub, mul, div, sqrt, log, sin, cos
&
1 job, tournament 20, initial depth 2 to 6, MAE metric,
parsimony 0.001, crossover 0.9, subtree, hoist, and point
mutation 0.01.
\\

\addlinespace[8pt]

\textbf{PyOperon}
&
Population 500, pool 500.
&
add, sub, mul, div, aq, exp, log, sin, cos, tanh, sqrt,
square, constants, variables
&
4 threads, max length 50, max depth 10, tournament 5,
keep best reinsertion, LM optimizer,
local search probability 1.0.
\\

\addlinespace[8pt]

\textbf{PySR}
&
8 populations of size 64, 500 cycles per iteration.
&
Binary $+,-,\times,/$. Unary square, cube, exp, log, sin, cos.
&
Serial execution with 1 process, max size 30, max depth 10,
parsimony 0.001, deterministic mode, precision 32,
best model selection.
\\

\addlinespace[8pt]

\textbf{QLattice}
&
BIC criterion.
&
QLattice/Feyn internal symbolic basis
&
4 threads, regression mode, BIC model selection,
4 significant digits.
\\

\addlinespace[8pt]

\textbf{DSO}
&
Batch size 1000.
&
add, sub, mul, div, sin, cos, exp, log
&
4 batch cores, reward inv NRMSE, $\epsilon=0.05$,
learning rate $5\times10^{-4}$, entropy weight 0.03,
expression length 4 to 64.
\\

\addlinespace[8pt]

\textbf{uDSR}
&
Batch size 1000, GP meld population 100 for 20 generations.
&
add, sub, mul, div, sin, cos, exp, log, sqrt, 1.0,
constants, poly token
&
1 batch core, reward inv NRMSE, $\epsilon=0.05$,
learning rate $5\times10^{-4}$, entropy weight 0.03,
expression length 4 to 100, GP meld and linear poly enabled.
\\

\addlinespace[8pt]

\textbf{iMCTS}
&
$K=500$, max depth 6.
&
$+,-,\times,/$, sin, cos, exp, log, real constant token
&
Exploration constant 4.0, $\gamma=0.5$,
GP rate 0.2, mutation 0.1, exploration 0.2,
max constants 10, LN Nelder--Mead optimization.
\\

\addlinespace[8pt]

\textbf{E2ESR}
&
200 input points, 10 bags, refine 10 trees,
stop refinement after 1.
&
Pretrained E2E decoder vocabulary with arithmetic,
inverse and powers, log and exp, trigonometric and
inverse trigonometric functions.
&
Single Torch thread, rescaling and execution forced to CPU,
top $k$ feature selection and relabeling,
missing variables filled with zero.
\\

\addlinespace[8pt]

\textbf{TPSR}
&
200 input points, 10 bags, refine 10 trees,
beam size 10, width 3, rollout 3, horizon 200.
&
Pretrained E2E decoder vocabulary.
Unsupported variables projected to zero.
&
4 CPU threads and 1 interop thread, E2E backbone,
sampling beam, one beam, $\lambda=0.1$,
no value training, reward sample limit 2048,
prefix cache disabled, CPU mode.
\\

\addlinespace[8pt]

\textbf{LLM-SR}
&
4 samples per iteration, max 10 fitted parameters.
&
Python expression skeletons from an LLM.
Constants are fitted.
&
Semantic prompt injection and prompt variables enabled,
sample persistence disabled,
\mbox{DeepInfra Llama-3.1-8B-Instruct-Turbo},
max 1024 tokens, temperature 0.6,
top $p$ 0.3, top $k$ 30.
\\

\addlinespace[8pt]

\textbf{DrSR}
&
4 samples per iteration, 400000 proposals,
max 10 fitted parameters.
&
Python expression skeletons from an LLM.
Constants are fitted.
&
Same prompt contract as LLM-SR,
one sampler and one evaluator,
sample persistence disabled,
\mbox{DeepInfra Llama-3.1-8B-Instruct-Turbo},
max 1024 tokens, temperature 0.6,
top $p$ 0.3, top $k$ 30.
\\

\addlinespace[8pt]

\textbf{RAG-SR}
&
Population 200, gene number 10.
&
Add, Sub, Mul, AQ, Sqrt, AbsLog, Abs, Square,
RSin, RCos, Max, Min, Neg
&
4 CPU threads, automatic lexicase selection,
crossover 0.9, mutation 0.1, max height 10,
RidgeCV, MinMax normalization, $R^2$ score,
external archive enabled,
\mbox{\texttt{number\_of\_invokes}=0.}
\\

\specialrule{0.4pt}{3pt}{1pt}

\rowcolor{gray!4}
\textbf{SymbolFit}
&
Max size and complexity 25.
&
Binary $+,-,\times,/$. Unary sin, cos, exp, log.
&
Serial deterministic search with one process,
model selection based on accuracy, input rescaling,
mean target scaling, max stderr 20,
fixed target uncertainty 1, no uncertainty fitting.
\\

\rowcolor{gray!4}
\textbf{FePySR}
&
8 experiments, FMN 30 epochs with batch size 64,
PySR uses 4 populations of 40 and 100 cycles per iteration.
&
Binary $+,-,\times,/$. Unary sin, cos, exp, log.
&
4 workers, FMN learning rate 0.1,
10 extracted features, 6 PySR features,
max size 20, max depth 8.
\\

\rowcolor{gray!4}
\textbf{JAXSR}
&
Max 5 terms, CV with 5 folds, greedy forward selection.
&
Constant and linear bases, polynomials of degree 3,
and interactions of order 2.
No transcendental or ratio terms.
&
BIC model selection, no regularization,
constant and linear bases enabled,
polynomial and interaction bases enabled.
\\

\end{longtable}
\endgroup

%% file: sections/appJ_llm_usage.tex
\section{LLM Usage and Information Boundaries}
\label{app:llm-usage}

\arena{} uses language models in two separate roles.
Llama participates in equation search for LLM-SR and DrSR.
Opus~5 performs symbolic postprocessing after search completion.
Search models cannot access ground truth expressions or ID and OOD evaluation targets.
\Cref{app:execution-protocol} defines the execution boundary.
The appendix specifies model roles, prompt settings, and information access for language model components.

\lstset{
  basicstyle=\ttfamily\footnotesize,
  breaklines=true,
  backgroundcolor=\color{black!6},
  frame=single,
  framerule=0.3pt,
  rulecolor=\color{black!25},
  framesep=5pt,
  xleftmargin=6pt,
  xrightmargin=6pt
}

\Cref{tab:llm-roles} summarizes the model roles and their relation to search.

\begin{table}[htbp]
    \centering
    \small
    \setlength{\tabcolsep}{4pt}
    \renewcommand{\arraystretch}{1.12}
    \caption{\textbf{Language model roles in \arena{}.}
    Opus~5 operates only during evaluator side postprocessing.}
    \label{tab:llm-roles}
    \begin{tabular}{@{}llll@{}}
        \toprule
        \textbf{Model} & \textbf{Stage} & \textbf{Primary role} & \textbf{Affects search} \\
        \midrule
        Opus~5 & Postprocessing & Symbolic processing and judgment & No \\
        Llama & LLM-SR search & Equation structure generation & Yes \\
        Llama & DrSR search & Generation and diagnostic feedback & Yes \\
        \bottomrule
    \end{tabular}
\end{table}

\subsection{Llama in LLM-SR and DrSR}
\label{app:llama-search}

The formal \core{} evaluation uses Llama 3.1 8B Instruct
\citep{grattafiori2024llama3}, served as
\texttt{meta-llama/Meta-Llama-3.1-8B-Instruct-Turbo} by
DeepInfra\footnote{\url{https://deepinfra.com/meta-llama/Meta-Llama-3.1-8B-Instruct-Turbo}} for
LLM-SR and DrSR.
Generation uses
\texttt{temperature=0.6},
\texttt{top\_p=0.3}, and
\texttt{max\_tokens=1024}.
Model weights remain fixed.
Candidate selection and coefficient fitting use training objectives.
ID and OOD splits remain reserved for evaluation.

\paragraph{LLM-SR.}
\label{app:llmsr-llama}

LLM-SR iteratively requests equation structures from Llama.
Prompts contain task context and variable descriptions.
Historical candidates provide additional search context.
Multiple islands maintain separate candidate histories.
Each generation prompt includes up to two functions selected by training score.
Each iteration requests four new structures.

For each structure, BFGS optimizes coefficients in \texttt{params} using training MSE.
All coefficients are initialized to one.
Candidates with higher training scores can enter the history pool.

The shared generation instruction is:

\begin{lstlisting}[basicstyle=\ttfamily\footnotesize,breaklines=true]
You are a helpful assistant tasked with discovering mathematical
function structures for scientific systems.
Complete the 'equation' function below, considering the physical
meaning and relationships of inputs.
\end{lstlisting}

The LLM-SR user message follows:

\begin{lstlisting}[basicstyle=\ttfamily\footnotesize,breaklines=true]
[fixed generation instruction]

[task background]
[variable descriptions]

[code preamble]

def equation_v0(...):
    ...

def equation_v1(...):
    """Improved version of `equation_v0`."""
\end{lstlisting}

Historical functions come from candidates recorded during the current run.
The final empty \texttt{equation\_vN} function requests the next structure.
Candidate evaluation occurs outside the Llama request.

\paragraph{DrSR.}
\label{app:drsr-llama}

DrSR uses the same equation generation mechanism and candidate pool design.
Additional Llama calls provide experience feedback and residual analysis.

\textbf{Experience feedback.}
Candidate outcomes are categorized using training score changes or execution failures.
Selected outcomes are converted into compact guidance for later generation.

\textbf{Residual analysis.}
Improved candidates can trigger a residual analysis request containing training data, the current expression, and prediction residuals.
The numerical matrix is
\[
[X,\;y,\;y-y_{\mathrm{pred}}],
\]
and displayed values use three decimal places.
The generated analysis can enter later generation prompts.

DrSR uses the task header:

\begin{lstlisting}[basicstyle=\ttfamily\footnotesize,breaklines=true]
Find the mathematical function skeleton that represents
{dependent}, given data on {independent}.
\end{lstlisting}

DrSR adds a single line output requirement to the shared generation instruction:

\begin{lstlisting}[basicstyle=\ttfamily\footnotesize,breaklines=true]
Write the final formula as a single-line return statement only.
Example: return params[0] + params[1] * x0 + params[2] * x1
\end{lstlisting}

A generation message follows:

\begin{lstlisting}[basicstyle=\ttfamily\footnotesize,breaklines=true]
[task header]

[optional residual analysis]

[optional previous experience]

[fixed generation instruction]

Variables:
{variables_block}

Background:
{background_text}

[candidate functions]

def equation_vN(...):
    ...
\end{lstlisting}

Candidate coefficients are evaluated locally.
The formal DrSR evaluator uses five
\(\operatorname{Uniform}(-1,1)\) initializations for BFGS and retains the
solution with the lowest training MSE.
Optimization uses
\texttt{maxiter=200},
\texttt{gtol=1e-10}, and
\texttt{eps=1e-12}.

\subsection{Opus~5 in Symbolic Evaluation}
\label{app:opus-symbolic-evaluation}

Claude Opus 5 \citep{anthropic2026opus5} operates only on completed runs.
Requests use \texttt{claude-opus-5}, contain no conversation history, and process final expressions only.
The configuration uses adaptive thinking, extra high effort, and
\texttt{max\_tokens=65536}.
Minute level trajectories used by EFF never invoke Opus~5.

Detailed simplification and adjudication procedures appear in
\Cref{app:opus-postprocessing}.
Opus~5 supports expression simplification, equivalence assessment, and pairwise structural assessment.
Ground truth and predicted expressions are processed independently before comparison.
Protected operator semantics and declared domain assumptions remain fixed.

Opus~5 does not directly compute evaluation scores.
Deterministic code aggregates frozen expressions and stored judgments.
\Cref{tab:opus-metric-role} records its contribution to each axis.

\begin{table}[htbp]
    \centering
    \small
    \setlength{\tabcolsep}{5pt}
    \renewcommand{\arraystretch}{1.12}
    \caption{\textbf{Role of Opus~5 in Multi Axis Evaluation.}}
    \label{tab:opus-metric-role}
    \begin{tabular}{@{}cl@{}}
        \toprule
        \textbf{Axis} & \textbf{Role of Opus~5} \\
        \midrule
        ID   & None \\
        OOD  & None \\
        SYM  & Simplification and equivalence judgment \\
        MIN  & Simplification before deterministic tree size computation \\
        EFF  & None \\
        STAB & Pairwise structural judgment of final expressions \\
        \bottomrule
    \end{tabular}
\end{table}

\subsection{Information Isolation and Leakage Control}
\label{app:llm-leakage-control}

Search models cannot access ground truth expressions or ID and OOD evaluation targets.
Formal runs map variable names to
\(x_0,x_1,\ldots\) and the target to \(y\).
Declared semantic descriptions can be supplied as algorithm inputs.
Such descriptions remain separate from ground truth formulas and evaluation targets.

For tasks lacking semantic metadata, the generic background is:

\begin{lstlisting}[basicstyle=\ttfamily\footnotesize,breaklines=true]
Find the mathematical function skeleton that represents
{target_desc}, given data on {feature_text}.
\end{lstlisting}

For SRSD tasks containing distractor variables, semantic roles are exposed as an unordered multiset.
The association between roles and individual variables remains hidden:

\begin{lstlisting}[basicstyle=\ttfamily\footnotesize,breaklines=true]
There are {n_total} candidate variables in an unknown order.
The unordered semantic-role multiset is: {role_text}.
The mapping from semantic roles to variable names is intentionally
hidden; do not assume which x_i corresponds to which semantic role.
\end{lstlisting}

The masking prevents direct identification of the numerical column associated with a semantic role.

\paragraph{Benchmark construction.}
The initial LLM-SR probe used for \calibset{} mining disables semantic prompt injection and uses a generic symbolic regression background.
The final \probe{} consists of DSO, PyOperon, iMCTS, and uDSR.
All four construction probes operate without language models.
Final \reservoir{} responses used for \core{} selection therefore contain no LLM generated probe responses.

\paragraph{Evaluator separation.}
Opus~5 receives ground truth only during symbolic processing of completed runs.
Its output cannot alter candidate generation or recorded search trajectories.
Processed expressions and judgments are frozen before metric aggregation.
Model identifiers and prompt hashes are retained for auditing.

%% file: references_iclr2027.bib
@book{koza1992genetic,
	title     = {Genetic Programming: On the Programming of Computers by Means of Natural Selection},
	author    = {Koza, John R.},
	year      = {1992},
	publisher = {MIT Press}
}

@misc{valipour2021symbolicgpt,
  title = {{SymbolicGPT}: A Generative Transformer Model for Symbolic Regression},
  author = {Valipour, Mojtaba and You, Bowen and Panju, Maysum and Ghodsi, Ali},
  year = {2021},
  url = {https://arxiv.org/abs/2106.14131}
}

@article{Langley1981,
	title={Data-driven discovery of physical laws},
	author={Langley, Pat},
	journal={Cognitive Science},
	volume={5},
	number={1},
	pages={31--54},
	year={1981}
}

@article{Schmidt2009,
  title = {Distilling free-form natural laws from experimental data},
  author = {Schmidt, Michael and Lipson, Hod},
  journal = {Science},
  volume = {324},
  number = {5923},
  pages = {81--85},
  year = {2009}
}

@inproceedings{kamienny2023deep,
  title = {Deep Generative Symbolic Regression with {M}onte-{C}arlo Tree Search},
  author = {Kamienny, Pierre-Alexandre and Lample, Guillaume and Lamprier, Sylvain and Virgolin, Marco},
  booktitle = {Proceedings of the 40th International Conference on Machine Learning},
  pages = {15655--15668},
  year = {2023},
  volume = {202},
  series = {Proceedings of Machine Learning Research},
  publisher = {P{ML}R},
  url = {https://proceedings.mlr.press/v202/kamienny23a.html}
}

@inproceedings{grayeli2024lasr,
  title = {Symbolic Regression with a Learned Concept Library},
  author = {Grayeli, Arya and Sehgal, Atharva and Costilla-Reyes, Omar and Cranmer, Miles and Chaudhuri, Swarat},
  booktitle = {Advances in Neural Information Processing Systems},
  volume = {37},
  year = {2024},
  pages = {44678--44709},
  doi = {10.52202/079017-1419}
}

@inproceedings{keijzer2003improving,
  author = {Keijzer, Maarten},
  title = {Improving Symbolic Regression with Interval Arithmetic and Linear Scaling},
  booktitle = {Genetic Programming: Proceedings of EuroGP 2003},
  series = {Lecture Notes in Computer Science},
  volume = {2610},
  pages = {70--82},
  year = {2003},
  publisher = {Springer},
  doi = {10.1007/3-540-36599-0_7}
}

@article{vladislavleva2009order,
  author = {Vladislavleva, Ekaterina J. and Smits, Guido F. and den Hertog, Dick},
  title = {Order of Nonlinearity as a Complexity Measure for Models Generated by Symbolic Regression via {Pareto} Genetic Programming},
  journal = {IEEE Transactions on Evolutionary Computation},
  volume = {13},
  number = {2},
  pages = {333--349},
  year = {2009},
  doi = {10.1109/TEVC.2008.926486}
}

@incollection{korns2011accuracy,
  author = {Korns, Michael F.},
  title = {Accuracy in Symbolic Regression},
  booktitle = {Genetic Programming Theory and Practice IX},
  series = {Genetic and Evolutionary Computation},
  pages = {129--151},
  year = {2011},
  publisher = {Springer},
  doi = {10.1007/978-1-4614-1770-5_8}
}

@article{uy2011semantically,
  author = {Uy, Nguyen Quang and Hoai, Nguyen Xuan and O'Neill, Michael and McKay, R. I. and Galv{\'a}n-L{\'o}pez, Edgar},
  title = {Semantically-Based Crossover in Genetic Programming: Application to Real-Valued Symbolic Regression},
  journal = {Genetic Programming and Evolvable Machines},
  volume = {12},
  number = {2},
  pages = {91--119},
  year = {2011},
  doi = {10.1007/s10710-010-9121-2}
}

@inproceedings{SRBench2025call,
  author = {Imai Aldeia, Guilherme Seidyo and Zhang, Hengzhe and Bomarito, Geoffrey and Cranmer, Miles and Fonseca, Alcides and Burlacu, Bogdan and La Cava, William G. and de Fran{\c{c}}a, Fabr{\'i}cio Olivetti},
  title = {Call for Action: Towards the Next Generation of Symbolic Regression Benchmark},
  booktitle = {Proceedings of the Genetic and Evolutionary Computation Conference Companion},
  pages = {2529--2538},
  year = {2025},
  publisher = {Association for Computing Machinery},
  doi = {10.1145/3712255.3734309}
}

@inproceedings{lacava2021contemporary,
  title = {Contemporary Symbolic Regression Methods and their Relative Performance},
  author = {La Cava, William and Orzechowski, Patryk and Burlacu, Bogdan and de Fran{\c{c}}a, Fabr{\'i}cio Olivetti and Virgolin, Marco and Jin, Ying and Kommenda, Michael and Moore, Jason H.},
  booktitle = {Proceedings of the Neural Information Processing Systems Track on Datasets and Benchmarks},
  year = {2021},
  url = {https://datasets-benchmarks-proceedings.neurips.cc/paper/2021/hash/c0c7c76d30bd3dcaefc96f40275bdc0a-Abstract-round1.html},
  volume = {1}
}

@article{matsubara2024rethinking,
  title = {Rethinking Symbolic Regression Datasets and Benchmarks for Scientific Discovery},
  author = {Matsubara, Yoshitomo and Chiba, Naoya and Igarashi, Ryo and Ushiku, Yoshitaka},
  journal = {Journal of Data-centric Machine Learning Research},
  year = {2024},
  volume = {1},
  number = {3},
  url = {https://openreview.net/forum?id=qrUdrXsiXX},
  pages = {1--38}
}

@misc{shojaee2025llmsrbench,
  title = {{LLM-SRBench}: A New Benchmark for Scientific Equation Discovery with Large Language Models},
  author = {Shojaee, Parshin and Meidani, Kazem and Gupta, Shashank and Farimani, Amir Barati and Reddy, Chandan K.},
  year = {2025},
  url = {https://arxiv.org/abs/2504.10415}
}

@inproceedings{shojaee2025llmsr,
  title = {{LLM-SR}: Scientific Equation Discovery via Programming with Large Language Models},
  author = {Shojaee, Parshin and Meidani, Kazem and Gupta, Shashank and Farimani, Amir Barati and Reddy, Chandan K.},
  booktitle = {International Conference on Learning Representations},
  year = {2025},
  url = {https://openreview.net/forum?id=m2nmp8P5in}
}

@inproceedings{shojaee2023tpsr,
  title = {Transformer-Based Planning for Symbolic Regression},
  author = {Shojaee, Parshin and Meidani, Kazem and Farimani, Amir Barati and Reddy, Chandan K.},
  booktitle = {Advances in Neural Information Processing Systems},
  volume = {36},
  year = {2023},
  pages = {45907--45919},
  doi = {10.52202/075280-1990}
}

@misc{cranmer2023pysr,
  title = {Interpretable Machine Learning for Science with {PySR} and {SymbolicRegression}.jl},
  author = {Cranmer, Miles},
  year = {2023},
  url = {https://arxiv.org/abs/2305.01582}
}

@inproceedings{burlacu2020operon,
  title = {{Operon C++}: An Efficient Genetic Programming Framework for Symbolic Regression},
  author = {Burlacu, Bogdan and Kronberger, Gabriel and Kommenda, Michael},
  booktitle = {Proceedings of the 2020 Genetic and Evolutionary Computation Conference Companion},
  pages = {1562--1570},
  year = {2020},
  publisher = {Association for Computing Machinery},
  doi = {10.1145/3377929.3398099}
}

@misc{brolos2021feyn,
	title = {An Approach to Symbolic Regression Using {Feyn}},
	author = {Brol{\o}s, Kevin Ren{\'e} and Machado, Meera Vieira and Cave, Chris and Kasak, Jaan and Stentoft-Hansen, Valdemar and Batanero, Victor Galindo and Jelen, Tom and Wilstrup, Casper},
	year = {2021},
	url = {https://arxiv.org/abs/2104.05417}
}

@inproceedings{petersen2021dsr,
	title = {Deep Symbolic Regression: Recovering Mathematical Expressions from Data via Risk-Seeking Policy Gradients},
	author = {Petersen, Brenden K. and Landajuela, Mikel and Mundhenk, T. Nathan and Santiago, Claudio P. and Kim, Sookyung and Kim, Joanne T.},
	booktitle = {International Conference on Learning Representations},
	year = {2021},
	url = {https://openreview.net/forum?id=m5Qsh0kBQG}
}

@inproceedings{mundhenk2021dsoseeding,
  title = {Symbolic Regression via Neural-Guided Genetic Programming Population Seeding},
  author = {Mundhenk, T. Nathan and Landajuela, Mikel and Glatt, Ruben and Santiago, Claudio P. and Faissol, Daniel M. and Petersen, Brenden K.},
  booktitle = {Advances in Neural Information Processing Systems},
  volume = {34},
  year = {2021},
  url = {https://proceedings.neurips.cc/paper/2021/hash/d073bb8d0c47f317dd39de9c9f004e9d-Abstract.html},
  pages = {24912--24923}
}

@inproceedings{landajuela2022unified,
  title = {A Unified Framework for Deep Symbolic Regression},
  author = {Landajuela, Mikel and Lee, Chak and Yang, Jiachen and Glatt, Ruben and Santiago, Claudio P. and Aravena, Ignacio and Mundhenk, Terrell N. and Mulcahy, Garrett and Petersen, Brenden K.},
  booktitle = {Advances in Neural Information Processing Systems},
  volume = {35},
  year = {2022},
  pages = {33985--33998},
  doi = {10.52202/068431-2463}
}

@inproceedings{kamienny2022end,
  title = {End-to-End Symbolic Regression with Transformers},
  author = {Kamienny, Pierre-Alexandre and d'Ascoli, St{\'e}phane and Lample, Guillaume and Charton, Fran{\c{c}}ois},
  booktitle = {Advances in Neural Information Processing Systems},
  volume = {35},
  year = {2022},
  pages = {10269--10281},
  doi = {10.52202/068431-0746}
}

@misc{huang2025imcts,
  title = {Improving {Monte Carlo} Tree Search for Symbolic Regression},
  author = {Huang, Zhengyao and Huang, Daniel Zhengyu and Xiao, Tiannan and Ma, Dina and Ming, Zhenyu and Shi, Hao and Wen, Yuanhui},
  year = {2025},
  url = {https://arxiv.org/abs/2509.15929}
}

@inproceedings{zhang2025ragsr,
	title = {{RAG-SR}: Retrieval-Augmented Generation for Neural Symbolic Regression},
	author = {Zhang, Hengzhe and Chen, Chao and Xue, Bing and Banzhaf, Wolfgang and Zhang, Mengjie},
	booktitle = {International Conference on Learning Representations},
	year = {2025},
	url = {https://openreview.net/forum?id=NdHka08uWn}
}

@misc{wang2025drsr,
  title = {{DrSR}: {LLM}-Based Scientific Equation Discovery with Dual Reasoning from Data and Experience},
  author = {Wang, Runxiang and Wang, Boxiao and Li, Kai and Zhang, Yifan and Cheng, Jian},
  year = {2025},
  url = {https://arxiv.org/abs/2506.04282}
}

@inproceedings{biggio2021nesymres,
  title = {Neural symbolic regression that scales},
  author = {Biggio, Luca and Bendinelli, Tommaso and Neitz, Alexander and Lucchi, Aurelien and Parascandolo, Giambattista},
  booktitle = {International Conference on Machine Learning},
  pages = {936--945},
  year = {2021}
}

@inproceedings{kiela2021dynabench,
	title = {{Dynabench}: Rethinking Benchmarking in {NLP}},
	author = {Kiela, Douwe and Bartolo, Max and Nie, Yixin and Kaushik, Divyansh and Geiger, Atticus and Wu, Zhengxuan and Vidgen, Bertie and Prasad, Grusha and Singh, Amanpreet and Ringshia, Pratik and Ma, Zhiyi and Thrush, Tristan and Riedel, Sebastian and Waseem, Zeerak and Stenetorp, Pontus and Jia, Robin and Bansal, Mohit and Potts, Christopher and Williams, Adina},
	booktitle = {Proceedings of the 2021 Conference of the North American Chapter of the Association for Computational Linguistics: Human Language Technologies},
	pages = {4110--4124},
	year = {2021},
	url = {https://aclanthology.org/2021.naacl-main.324/}
}

@inproceedings{ma2021dynaboard,
  title = {{Dynaboard}: An Evaluation-As-A-Service Platform for Holistic Next-Generation Benchmarking},
  author = {Ma, Zhiyi and Ethayarajh, Kawin and Thrush, Tristan and Jain, Somya and Wu, Ledell and Jia, Robin and Potts, Christopher and Williams, Adina and Kiela, Douwe},
  booktitle = {Advances in Neural Information Processing Systems},
  volume = {34},
  year = {2021},
  url = {https://proceedings.neurips.cc/paper/2021/hash/55b1927fdafef39c48e5b73b5d61ea60-Abstract.html},
  pages = {10351--10367}
}

@inproceedings{mazumder2023dataperf,
  title = {{DataPerf}: Benchmarks for Data-Centric {AI} Development},
  author = {Mazumder, Mark and Banbury, Colby and Yao, Xiaozhe and Karla{\v{s}}, Bojan and Gaviria Rojas, William and Diamos, Sudnya and Diamos, Greg and He, Lynn and Parrish, Alicia and Kirk, Hannah Rose and Quaye, Jessica and Rastogi, Charvi and Kiela, Douwe and Jurado, David and Kanter, David and Mosquera, Rafael and Cukierski, Will and Ciro, Juan and Aroyo, Lora and Acun, Bilge and Chen, Lingjiao and Raje, Mehul and Bartolo, Max and Eyuboglu, Evan Sabri and Ghorbani, Amirata and Goodman, Emmett and Howard, Addison and Inel, Oana and Kane, Tariq and Kirkpatrick, Christine R. and Sculley, D. and Kuo, Tzu-Sheng and Mueller, Jonas W. and Thrush, Tristan and Vanschoren, Joaquin and Warren, Margaret and Williams, Adina and Yeung, Serena and Ardalani, Newsha and Paritosh, Praveen and Zhang, Ce and Zou, James and Wu, Carole-Jean and Coleman, Cody and Ng, Andrew and Mattson, Peter and Janapa Reddi, Vijay},
  booktitle = {Advances in Neural Information Processing Systems},
  volume = {36},
  year = {2023},
  pages = {5320--5347},
  doi = {10.52202/075280-0235}
}

@inproceedings{perlitz2024efficient,
  title = {Efficient Benchmarking (of Language Models)},
  author = {Perlitz, Yotam and Bandel, Elron and Gera, Ariel and Arviv, Ofir and Ein-Dor, Liat and Shnarch, Eyal and Slonim, Noam and Shmueli-Scheuer, Michal and Choshen, Leshem},
  booktitle = {Proceedings of the 2024 Conference of the North American Chapter of the Association for Computational Linguistics: Human Language Technologies (Volume 1: Long Papers)},
  year = {2024},
  pages = {2519--2536},
  doi = {10.18653/v1/2024.naacl-long.139}
}

@inproceedings{polo2024tinybenchmarks,
	title = {{tinyBenchmarks}: Evaluating {LLMs} with Fewer Examples},
	author = {Polo, Felipe Maia and Weber, Lucas and Choshen, Leshem and Sun, Yuekai and Xu, Gongjun and Yurochkin, Mikhail},
	booktitle = {Proceedings of the 41st International Conference on Machine Learning},
	pages = {34303--34326},
	year = {2024},
	url = {https://proceedings.mlr.press/v235/polo24a.html}
}

@inproceedings{vivek2024anchorpoints,
  title = {Anchor Points: Benchmarking Models with Much Fewer Examples},
  author = {Vivek, Rajan and Ethayarajh, Kawin and Yang, Diyi and Kiela, Douwe},
  booktitle = {Proceedings of the 18th Conference of the European Chapter of the Association for Computational Linguistics (Volume 1: Long Papers)},
  year = {2024},
  pages = {1576--1601},
  doi = {10.18653/v1/2024.eacl-long.95}
}

@inproceedings{saranathan2025sublime,
	title = {{SubLIME}: Subset Selection via Rank Correlation Prediction for Data-Efficient {LLM} Evaluation},
	author = {Saranathan, Gayathri and Xu, Cong and Alam, Mahammad Parwez and Kumar, Tarun and Foltin, Martin and Wong, Soon Yee and Bhattacharya, Suparna},
	booktitle = {Proceedings of the 63rd Annual Meeting of the Association for Computational Linguistics (Volume 1: Long Papers)},
	pages = {30572--30593},
	year = {2025},
	doi = {10.18653/v1/2025.acl-long.1477},
}

@misc{yu2026fepysr,
  title = {{FePySR}: A Neural Feature Extraction Framework for Efficient and Scalable Symbolic Regression},
  author = {Yu, Zhiming and Lu, Wangtao and Lai, Xin},
  year = {2026},
  url = {https://arxiv.org/abs/2605.12704}
}

@article{tsoi2025symbolfit,
  title = {{SymbolFit}: Automatic Parametric Modeling with Symbolic Regression},
  author = {Tsoi, Ho Fung and Rankin, Dylan and Caillol, Cecile and Cranmer, Miles and Dasu, Sridhara and Duarte, Javier and Harris, Philip and Lipeles, Elliot and Loncar, Vladimir},
  journal = {Computing and Software for Big Science},
  volume = {9},
  pages = {12},
  year = {2025},
  doi = {10.1007/s41781-025-00140-9}
}

@misc{kitchin2024jaxsr,
  title = {{JAXSR}: {JAX}-based Symbolic Regression},
  author = {Kitchin, John},
  year = {2024},
  url = {https://github.com/jkitchin/jaxsr},
  howpublished = {Software repository}
}

@misc{stephens2016gplearn,
  author = {Trevor Stephens},
  title = {gplearn: Genetic Programming in {Python}, with a scikit-learn Inspired {API}},
  year = {2016},
  url = {https://github.com/trevorstephens/gplearn},
  howpublished = {Software repository}
}

@article{meurer2017sympy,
  title   = {SymPy: symbolic computing in Python},
  author  = {Meurer, Aaron and Smith, Christopher P. and Paprocki, Mateusz
             and {\v C}ert{\'i}k, Ond{\v r}ej and Kirpichev, Sergey B.
             and Rocklin, Matthew and Kumar, AMiT and Ivanov, Sergiu
             and Moore, Jason K. and Singh, Sartaj and Rathnayake, Thilina
             and Vig, Sean and Granger, Brian E. and Muller, Richard P.
             and Bonazzi, Francesco and Gupta, Harsh and Vats, Shivam
             and Johansson, Fredrik and Pedregosa, Fabian and Curry, Matthew J.
             and Terrel, Andy R. and Rou{\v c}ka, {\v S}t{\v e}p{\'a}n
             and Saboo, Ashutosh and Fernando, Isuru and Kulal, Sumith
             and Cimrman, Robert and Scopatz, Anthony},
  journal = {PeerJ Computer Science},
  volume  = {3},
  pages   = {e103},
  year    = {2017},
  doi     = {10.7717/peerj-cs.103}
}

@article{grattafiori2024llama3,
  title   = {The Llama 3 Herd of Models},
  author  = {Grattafiori, Aaron and others},
  journal = {arXiv preprint arXiv:2407.21783},
  year    = {2024}
}

@misc{anthropic2026opus5,
  author       = {{Anthropic}},
  title        = {System Card: Claude Opus 5},
  year         = {2026},
  month        = jul,
  howpublished = {\url{https://www.anthropic.com/claude-opus-5-system-card}}
}
